\documentclass[conference]{IEEEtran}
\IEEEoverridecommandlockouts
\usepackage{cite}
\usepackage{amsmath,amssymb,amsfonts}
\usepackage{algorithm}
\usepackage{graphicx}
\usepackage{textcomp}
\usepackage{xcolor}
\usepackage{hyperref}
\usepackage{cleveref}
\usepackage{tikz}

\def\BibTeX{{\rm B\kern-.05em{\sc i\kern-.025em b}\kern-.08em
    T\kern-.1667em\lower.7ex\hbox{E}\kern-.125emX}}

\usepackage{tikz}
\usepackage{amsmath}
\usepackage{amssymb}
\usepackage{cleveref}
\usepackage{subcaption}
\usepackage{algpseudocode}

\newcommand{\org}{\textit{org}}
\newcommand{\pert}{\textit{pert}}

\newcommand{\pred}{\textit{pred}}
\newcommand{\refe}{\textit{ref}}

\newcommand{\diff}{\textit{diff}}
\newcommand{\real}{\textit{real}}
\newcommand{\iou}{\textit{iou}}
\newcommand{\spawn}{\textit{spawn}}
\newcommand{\steps}{\textit{steps}}
\newcommand{\new}{\textit{new}}
\newcommand{\numb}{\textit{numb}}
\newcommand{\start}{\textit{start}}
\newcommand{\s}{\textit{s}}
\newcommand{\cent}{\textit{c}}
\newcommand{\gt}{\textit{gt}}
\newcommand{\lerr}{\textit{le}}
\newcommand{\legt}{\textit{legt}}
\newcommand{\rgb}{\textit{rgb}}
\newcommand{\inp}{\textit{in}}
\newcommand{\gG}{\mathcal{G}}

\begin{document}

\makeatletter

\newcommand{\linebreakand}{%
  \end{@IEEEauthorhalign}
  \hfill\mbox{}\par
  \mbox{}\hfill\begin{@IEEEauthorhalign}
}

\title{Learning to Detect Label Errors by Making Them: A Method for Segmentation and Object Detection Datasets 
}

\author{
  \IEEEauthorblockN{1\textsuperscript{st} Sarina Penquitt}
  \IEEEauthorblockA{\textit{Department of Mathematics} \\
    \textit{University of Wuppertal}\\
    Wuppertal, Germany \\
    penquitt@uni-wuppertal.de}
  \and
  \IEEEauthorblockN{2\textsuperscript{nd} Tobias Riedlinger}
  \IEEEauthorblockA{\textit{Department of Mathematics} \\
    \textit{Technical University of Berlin}\\
    Berlin, Germany \\
    riedlinger@math.tu-berlin.de}
  \and
    \IEEEauthorblockN{3\textsuperscript{rd} Timo Heller}
  \IEEEauthorblockA{\textit{Department of Mathematics} \\
    \textit{University of Wuppertal}\\
    Wuppertal, Germany \\
    timo.heller@uni-wuppertal.de}
  \linebreakand
  \IEEEauthorblockN{4\textsuperscript{th} Markus Reischl}
  \IEEEauthorblockA{\textit{Institute for Automation and Applied Informatics} \\
    \textit{Karlsruhe Institute of Technology (KIT)}\\
    Karlsruhe, Germany  \\
    markus.reischl@kit.edu}
  \and
    \IEEEauthorblockN{5\textsuperscript{th} Matthias Rottmann}
  \IEEEauthorblockA{\textit{Institute of Computer Science} \\
    \textit{Osnabrück University}\\
    Osnabrück, Germany \\
    matthias.rottmann@uos.de}
}

\maketitle

\begin{abstract}
Recently, detection of label errors and improvement of label quality in datasets for supervised learning tasks has become an increasingly important goal in both research and industry. The consequences of incorrectly annotated data include reduced model performance, biased benchmark results, and lower overall accuracy. Current state-of-the-art label error detection methods often focus on a single computer vision task and, consequently, a specific type of dataset, containing, for example, either bounding boxes or pixel-wise annotations. Furthermore, previous methods are not learning-based.
In this work, we overcome this research gap. We present a unified method for detecting label errors in object detection, semantic segmentation, and instance segmentation datasets. In a nutshell, our approach -- learning to detect label errors by making them -- works as follows: we inject different kinds of label errors into the ground truth. Then, the detection of label errors, across all mentioned primary tasks, is framed as an instance segmentation problem based on a composite input.
In our experiments, we compare the label error detection performance of our method with various baselines and state-of-the-art approaches of each task's domain on simulated label errors across multiple tasks, datasets, and base models. This is complemented by a generalization study on real-world label errors. Additionally, we release 459 real label errors identified in the Cityscapes dataset and provide a benchmark for real label error detection in Cityscapes. 
\end{abstract}

\section{Introduction}

\begin{figure}[ht]
    \centering
    \includegraphics[scale=0.5,width=\linewidth]{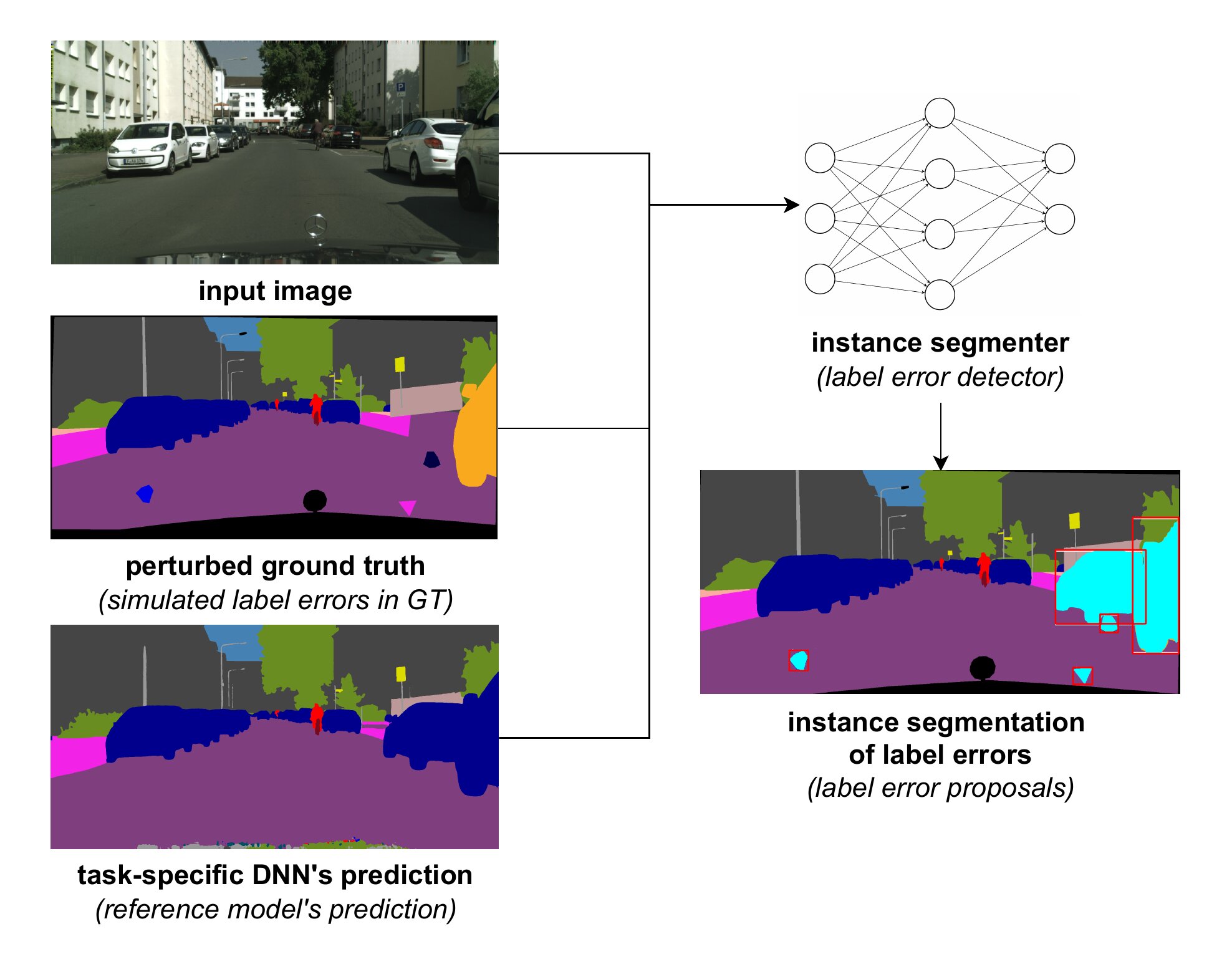}
    \caption{Learning task of our label error method exemplary for Cityscapes in semantic segmentation.}
    \label{fig:teaser}
\end{figure}

Deep learning thrives on data: the more complex the task, the more data is required. In computer vision, larger training datasets consistently improve model performance~\cite{Sun2017Effect}, driving demand for large-scale, high-quality annotations. However, creating such datasets, especially for tasks like object detection or semantic segmentation, is expensive and labor-intensive~\cite{Lundervold2019MRI, Feng2021Multi, Muller2019Mislabeled, Haakman2021AI, Schmidt2019Crowdproduktion, Heller2018ImperfectSL}. The more tedious the annotation process, the higher likelier is human error~\cite{Feng2021Multi, jakubik2024improvinglabelerrordetection, Muller2019Mislabeled, Cleanlab2023Segmentation, Pleiss2020Mislab}.
Despite this, large amounts of high-quality labeled data remain the gold standard in critical domains like autonomous driving or medical imaging~\cite{Kuutti2021Survey, Lundervold2019MRI, Feng2021Multi}. Accordingly, recent research has increasingly focused on label error detection in image classification~\cite{Northcutt2021Confi, jakubik2024improvinglabelerrordetection, Muller2019Mislabeled}, object detection~\cite{Schubert2024Label, Baer2023Localization, Cleanlab2023Object}, and semantic segmentation~\cite{Rottmann2023Label, Cleanlab2023Segmentation, Kim2024Correction}.
Label errors can substantially degrade model performance~\cite{Northcutt2021Confi, Agnew2023Effects, He2019Regr, Brodley2011Mislabeled, Schilling2022Variability}, causing poor generalization~\cite{Baer2023Localization, Pleiss2020Mislab}, biased predictions~\cite{Agnew2024AnnotQual}, and unstable benchmarks~\cite{northcutt2021pervasive}. While minor localization noise may be tolerable, missing or wrongly labeled objects can seriously disrupt learning~\cite{Agnew2024AnnotQual}. For instance, larger models are more likely to memorize noisy data, leading to worse predictions~\cite{northcutt2021pervasive}. In~\cite{Baer2023Localization} a network for localization label refinement was introduced, showing a performance drop in state-of-the-art object detectors trained with noisy labels. A semi-automated label error detection system could thus dramatically reduce the manual review burden and associated costs.
Designing such systems is challenging due to diverse annotation formats, the absence of ground truth for error locations, and intrinsic ambiguity caused by aleatoric uncertainty. 
To our knowledge, no existing method detects label errors across object detection, semantic segmentation, and instance segmentation. In this paper, we propose a unified framework (see fig.~\ref{fig:teaser}) that detects three types of label errors, i.e., missing annotations, wrong classes, and spurious (invented) labels, across multiple datasets and dataset types: Pascal VOC~\cite{PascalVOC}, COCO~\cite{COCO}, Cityscapes~\cite{Cityscapes}, ADE20K~\cite{ADE20K}, and LIVECell~\cite{LiveCell}. We simulate label noise using a perturbation algorithm and train an instance segmentation model to detect these errors. As input, this instance segmenter receives an original image, the perturbed ground truth (GT) and for simplification more reference information to be specified later on. The predictions of our instance segmenter serve as potential label error proposals.

We complement our methodological contribution by a benchmark for label error detection using simulated and real label errors and show that our approach outperforms existing state-of-the-art methods, also in detecting real label errors. Finally, we release a revised version of the Cityscapes ground truth, informed by our method. We contribute:
\begin{itemize}
\item We propose the first unified method to detect label errors across object detection, semantic segmentation, and instance segmentation. Furthermore, we close the research gap of \emph{learning} the detection of label errors.
\item Inspired by typical real label errors, we simulate three types of label errors by randomly perturbing the ground truth of a given dataset and learn their detection.
In a broad numerical study, we evaluate our method in terms of simulated errors in PascalVOC, COCO, Cityscapes, ADE20K and LIVECell, as well as real errors in Cityscapes.
We provide multiple ablation studies, e.g., on the composition of the input to our label error detection method and perturbation frequency and rate.
\item Finally, we introduce a benchmark for label error detection across diverse vision tasks and complement it with a revised version of the Cityscapes ground truth, where we found 459 label errors.
\end{itemize}

The code is available under \href{https://github.com/spenquitt/Learning_to_detect_label_errors_in_segmentation_and_object_detection_datasets}{GitHub}.

\section{Related Work}\label{related_work}

In this section, we provide an overview of existing label error detection methods for the computer vision tasks image classification, object detection, semantic segmentation, and instance segmentation. These methods differ in terms of their objectives and approach.
The effects of label errors should not be underestimated as pointed out by~\cite{Grad2024BenchInst}. They provide an annotation tool that simulates label noise in instance segmentation datasets and demonstrate that the performance of instance segmentation models suffers when label noise is present in the training data. To reduce the influence of label noise in training, \cite{Yang2020Inst} introduced a method for training an instance segmentation model, using different losses such that the model becomes robust under noisy labels.
Some works jointly consider the cost of labeling and reviewing, aiming at reducing overall annotation cost. In~\cite{Schubert2024DeepActive} a basic label error detection method is combined with active learning and a trade-off policy between querying new labels and reviewing already available ones. This approach maximizes model performance while minimizing annotation costs. The same goal is pursued by \cite{Chen2022Inst}, designing a method that suppresses mislabeling between foreground and background, as well as noise from mislabeled instances. In~\cite{Baer2023Localization} a localization label refinement network is developed to reduce label localization noise in object detection datasets and thus improve the performance of the object detector that suffers from this type of noise. The reduction in labeling costs is closely linked to the correction of label errors. In~\cite{Kim2024Correction} a method is proposed for detecting label errors and actively correcting them in semantic segmentation datasets.
The authors of~\cite{Schilling2023AI2Seg} propose a method for the automated inspection of biomedical instance segmentation datasets with respect to noisy annotations. Domain experts can use the tool to review their annotated dataset and subsequently improve the quality of the annotations using the integrated annotation tool.
There are various methods for label error detection, which take differing methodological approaches. One line of work focuses on assigning label quality scores to dataset annotations. These scores indicate the quality of the annotation of an image and induces a priority ranking for a potential review process. In the field of label error detection scoring methods, an empirical study was published by \cite{kuan2022labelquality} that focuses on classification datasets. In~\cite{Cleanlab2023Object} the framework \textit{ObjectLab} estimates which images in object detection datasets are correctly annotated by assigning a label quality score. In~\cite{Cleanlab2023Segmentation}, different quality scores for semantic segmentation datasets are studied, such as the proportion of correctly classified pixels, using the model's predicted probabilities or the softmin score.
Another approach is to consider the loss of a neural network in the decision about mislabeled data. In~\cite{Yue2024CTRL} a label error detection algorithm was developed that applies clustering algorithms to the training loss distribution for the distinction of clean and noisy labels in image classification datasets.
For object detection datasets, in \cite{Schubert2024Label} label errors are detected by examination of instance-wise object detection losses.
\begin{figure*}[!h]
    \centering
    \includegraphics[scale=0.33]{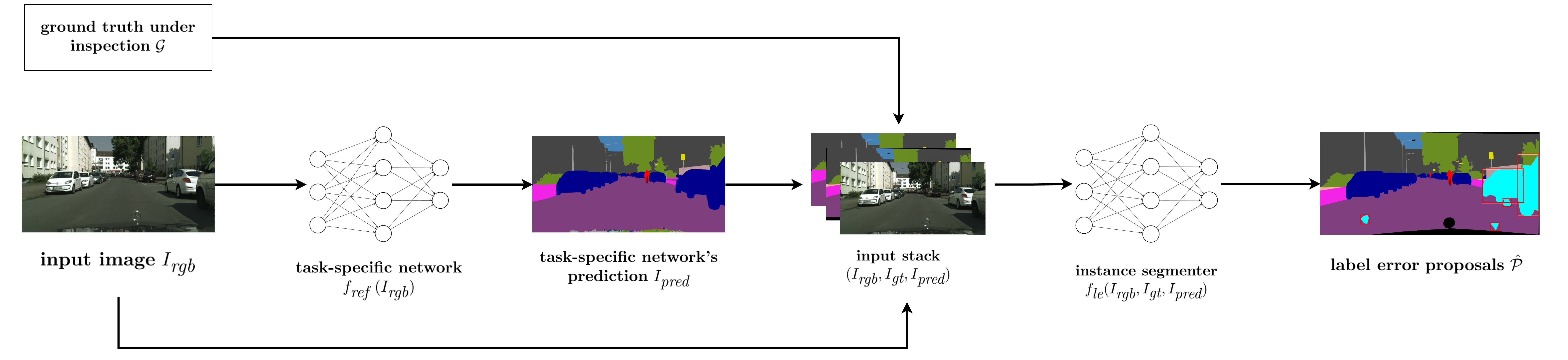}
    \caption{Visualization of our label error detection method.}
    \label{fig:lab_err_method}
\end{figure*}
The use of uncertainty quantification is also widespread in image classification, object detection, as well as in semantic and instance segmentation. For image classification datasets, so-called confident learning for label noise estimation and label error detection was introduced in~\cite{Northcutt2021Confi}. \cite{jakubik2024improvinglabelerrordetection} developed more sophisticated model uncertainty measures to detect label errors and to reflect model's predictive uncertainty accurately. 
The authors of~\cite{Lmd2023GCPR, Lmd2025IJCV} presents a post-processing method that uses uncertainty quantification to detect label errors in LiDAR object detection datasets, by estimating the prediction quality of a LiDAR object detector. In~\cite{Rottmann2023Label} a method for semantic segmentation datasets was developed that detects label errors based on connected components and uncertainty quantification. In~\cite{Cloud2023Segmentation} the authors built on \cite{Rottmann2023Label} and integrate additionally trustworthiness and hardness indices for the detection of incorrect labels in large Earth observation semantic segmentation datasets.

As opposed to all previously mentioned works, our contribution is a learning based method. Instead of focusing on a single type of annotation, we apply our method to object detection, semantic and instance segmentation annotations.

\section{Label Error Detection Method}\label{method}

In this section, we introduce our method for detecting label errors in object detection, semantic segmentation and instance segmentation datasets. We proceed as follows: First, we present the label error detection model, followed by its training procedure. Thereafter, we describe our label error simulation and perturbation algorithm, before finally explaining how we evaluate label error detection methods.

\textbf{Label Error Detection Model.} Our label error detection model receives an input image $I_\rgb$ as well as a corresponding ``ground truth'' $\gG = \{ p_1, \ldots, p_S \}$, $S \in \mathbb{N}$. 
Each $p_i$ is comprised of a polygon representing an object (potentially contained in $I_\rgb$) and a class label $c_i$ for all $i=1,\ldots,S$. 
With $I_\gt$ we denote an image or mask representation of $\gG$. Suppose that we have a label error ground truth (LEGT) $\gG_\legt = \{p_1^\legt,\ldots,p_{M}^\legt\}$, $M \in \mathbb{N}$, with $p_i^\legt$ defined analogously to $p_i$, where $\gG_\legt$ describes the set of all incorrectly annotated objects in $\gG$. The goal is that the label error detector (here an instance segmentation model) identifies label errors in $\gG$. As output our model provides an instance segmentation $\hat{\mathcal{P}}$ of possible label errors. Figure \ref{fig:lab_err_method} provides an overview of our model.
For learning label error detection in a given ground truth $\gG$, the knowledge acquired by deep neural networks (DNNs) that learned the original task (associated with $\gG$) of the given dataset are utilized.
We call such a DNN a \emph{reference DNN} $f_\refe$, that provides a \emph{reference prediction} $I_{\pred} = f_\refe(I_\rgb)$. If a given dataset's annotation consists of semantic segmentation masks, the reference DNN predicts such masks as well, analogously for object detection and instance segmentation. 
The input image $I_\rgb$, the ground truth mask $I_\gt$ and the reference prediction $I_\pred$ stacked together serve as input for the downstream instance segmenter, that learns the label error detection. That is, $f_\lerr: ( I_\rgb, I_\gt, I_{\pred} ) \mapsto \hat{\mathcal{P}} $, where $\hat{\mathcal{P}}$ is supposed to estimate $\gG_\legt$.
Depending on the dataset, we introduce a fourth input which is an additional difference mask $I_\diff$ explicitly containing differences between 
$I_\gt$ and $I_{\pred}$. 

\begin{figure*}[htbp]
    \centering
    \includegraphics[scale=0.35]{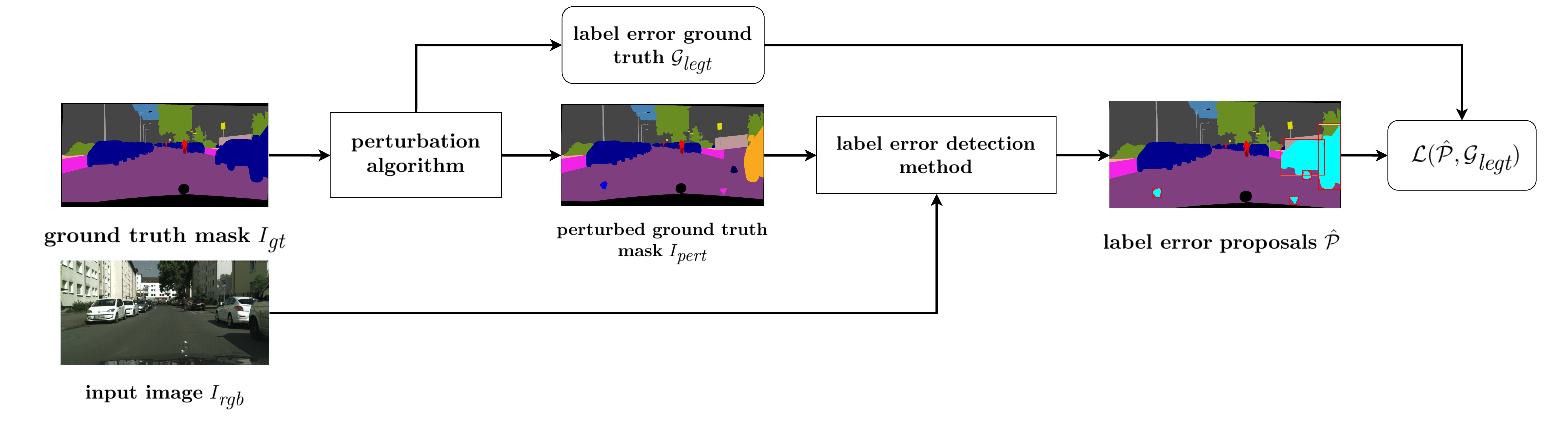}
    \caption{Visualization of the learning process in our label error detection method.}
    \label{fig:learning_err}
\end{figure*}
\textbf{Learning to Detect Label Errors.} We now specify the learning process, that is summarized in Figure \ref{fig:learning_err}.
We start with a given input image $I_\rgb$ with its corresponding ground truth $\gG_\org$ and ground truth mask $I_\org$. Then we apply a label perturbation algorithm that randomly perturbs labels of $\gG_\org$ at a chosen rate $q$. Therefrom, we obtain a perturbed ground truth $\gG_\pert$ (taking the role of the ground truth under inspection $\gG$), a perturbed ground truth mask $I_\pert$ and a corresponding label error ground truth $\gG_\legt$. As we algorithmically perturb labels, we have the knowledge which of our labels deviate from the original ones in $\gG_\org$ and this information is stored in $\gG_\legt$. The label perturbation algorithm is introduced in the next paragraph. 
Afterwards, a triplet $(I_\rgb, \gG_\pert, \gG_\legt)$ is given for each sample in the chosen dataset and this is used to train $f_\lerr$ using typical off-the-shelf instance segmentors. Losses $\mathcal{L}$ are then computed as functions of $\hat{\mathcal{P}}$ and $\gG_\legt$.

\textbf{Label Perturbation Algorithm.} We developed a perturbation algorithm for the label error simulation.
For an arbitrary dataset, let $I_\rgb$ be an RGB image with ground truth annotations $\gG_\org=\{p_1^\org,\ldots,p_N^\org\}$, $N\in\mathbb{N}$. We assume that these annotations are clean and accurate.
Therein, each $p_i^\org$ is comprised of a polygon covering a ground truth object and a class label $c_i^\org$ for all $i=1,\ldots,N$. 
Such a polygon can be a bounding box or a segment.
Drawing all $p_i^\org$ onto the image plane, e.g., via a coloring scheme, yields a ground truth mask which we refer to as $I_\org$.
In our perturbation algorithm, we randomly generate a perturbed ground truth by modifying $\gG_\org$.
The algorithm iterates over each ground truth polygon $p_i^\org$, for $i=1,\ldots,N$, generating a perturbed ground truth $\gG_\pert=\{p_1^\pert,\ldots,p_{N'}^\pert\}$, $N' \in \mathbb{N}$, as well as a label error ground truth (LEGT) $\gG_\legt=\{p_1^\legt,\ldots,p_{M}^\legt\}$, $M \in \mathbb{N}$.
Within these two sets, $p_i^\pert$ and $p_i^\legt$ are defined analogously to $p_i^\org$. 
Corresponding to the perturbed ground truth $\gG_\pert$ (the original ground truth consisting of simulated label errors) we define the perturbed ground truth mask as $I_\pert$.
The label error ground truth $\gG_\legt$ includes all simulated label error objects. To perform data augmentation, we apply the label perturbation algorithm multiple times to a given dataset, obtaining a number of copies of the dataset with different label errors. We denote the number of copies by $t$.
The algorithms initially starts with $\gG_\pert = \gG_\legt = \emptyset$ . In each iteration $i$ we perform a Bernoulli experiment such that with probability $q$ we add a perturbation to the dataset and with probability $1-q$ we do not.
If we add a perturbation, then we uniformly choose a random perturbation from three different types of label errors.
\emph{Drop:} $p_i^\org$ is deleted and is not part of the perturbed ground truth $\gG_\pert$, but therefore added to $\gG_\legt$.
\emph{Flip:} the class affiliation of $p_i^\org$ is altered (drawing uniformly a label from the remaining ones), resulting in a $p_i^\pert$ with altered $c_i^\pert \neq c_i^\org$, which is added to $\gG_\pert$ and $\gG_\legt$.
\emph{Spawn:} $p_i^\org$ is also part of the perturbed ground truth $\gG_\pert$, i.e., $p_i^\org = p_i^\pert$. However, we add another spawned ground truth polygon to $\gG_\pert$ and $\gG_\legt$, which does not represent any object in $I_\rgb$ and is generated randomly. 
The above description is for a single image-ground-truth pair. Performing the iteration for all image-ground-truth pairs of a given dataset yields an perturbed ground truth for the whole dataset. 

\textbf{Evaluation.} 
We evaluate the predictions of our label error detector by means of label error ground truth.
We have two cases: evaluation on 1) simulated label errors $\hat{\mathcal{P}}_\pert = f_\lerr( I_\rgb, I_\pert, I_\pred)$ and 2) real label errors $\hat{\mathcal{P}}_\real = f_\lerr( I_\rgb, I_\gt, I_\pred )$. 
In the first case, we compare $\hat{\mathcal{P}}_\pert$ with $\gG_\legt^\pert$ on test data not used for training $f_\lerr$, where the label error ground truth $\gG_\legt^\pert$ is obtained by the perturbation algorithm. 
In the second case, $\gG_\legt^\real$ is provided by human review of the given dataset's annotations and we compare predictions $\hat{\mathcal{P}}_\real$ of our method (wherein the labels under inspection are the original ones from a given dataset) with the dataset's annotations $\gG_\legt^\real$.
Following the usual approach to image detection and segmentation methods, our aim is to identify as many label errors as possible from the label error ground truth, while minimizing false predictions of label errors.
For each prediction of a given label error detection method, we decide whether it is a label error (true positive, TP$_\lerr$) or no label error (false positive, FP$_\lerr$).
A prediction is a TP$_\lerr$, if the Intersection over Union (IoU) between the prediction under consideration and a label error (from the label error ground truth) is greater than or equal to a chosen threshold $\tau_\iou\in(0,1]$. 
Otherwise it is an FP$_\lerr$.
Label errors from the label error ground truth that do not receive a match during this procedure are referred to as false negatives (FN$_\lerr$).
In both of the above cases (inspecting methods in simulated and real label errors), the evaluation of $\hat{\mathcal{P}}$ via $\gG_\legt$ is performed using evaluation protocol introduced by the COCO benchmark~\cite{COCO}. 
We consider the metrics Average Precision (AP), Precision, Recall and F1-Score. Here the AP is defined as the precision averaged over 10 Intersection over Union (IoU) values from 0.5 to 0.95~\cite{COCO}.


\section{Numerical Experiments}\label{numerical}

In this section, we apply our label error detection method to the semantic segmentation datasets Cityscapes and ADE20K, the object detection datasets PascalVOC and COCO, and the instance segmentation dataset LIVECell. 
We compare our results with existing state-of-the-art label error detection methods. After that, we present the results of ablation studies on the Cityscapes dataset and examine the detection of real label errors.

\textbf{Implementation Details.}
We implemented our method and benchmark in Detectron2 library ~\cite{detectron2}.
We used the instance segmentation model Cascade Mask-RCNN~\cite{cascade} as a label error detector, training it on only one class. This means that, even though we simulated three different types of label errors (\emph{drop}, \emph{flip} and \emph{spawn}), all of these were included in the training as class \emph{error}. 
In the semantic segmentation datasets, we limited the perturbation by setting minimum and maximum sizes for the objects being perturbed. In Cityscapes, objects required an area of at least 800 pixels, and in LIVECell and ADE20K, 200 pixels. The maximum size was set at 10,000 pixels. We introduced these constraints as there are many small objects in Cityscapes, however, our aim is to create a training dataset that contains visible but not unrealistically large label errors for training. For LIVECell and ADE20K, we opted for a smaller segment size due to the smaller image sizes. 
The creation of ground truth and perturbed ground truth masks differs in the respective computer vision task. In the semantic segmentation datasets, we use existing color coding to assign a color to each class. For instance segmentation, we randomly generate a color for each object to distinguish the individual segments. For object detection, we generate an HSV color model to reflect overlapping regions between two bounding boxes in the color map.
This enables us to display each object in full without losing any information about it. An overview of the different types of masks can be found in figure~\ref{fig:gtmasks}.
\begin{figure}
    \centering
    \raisebox{0.7cm}{(a)}\quad
    \includegraphics[width=0.4\linewidth]{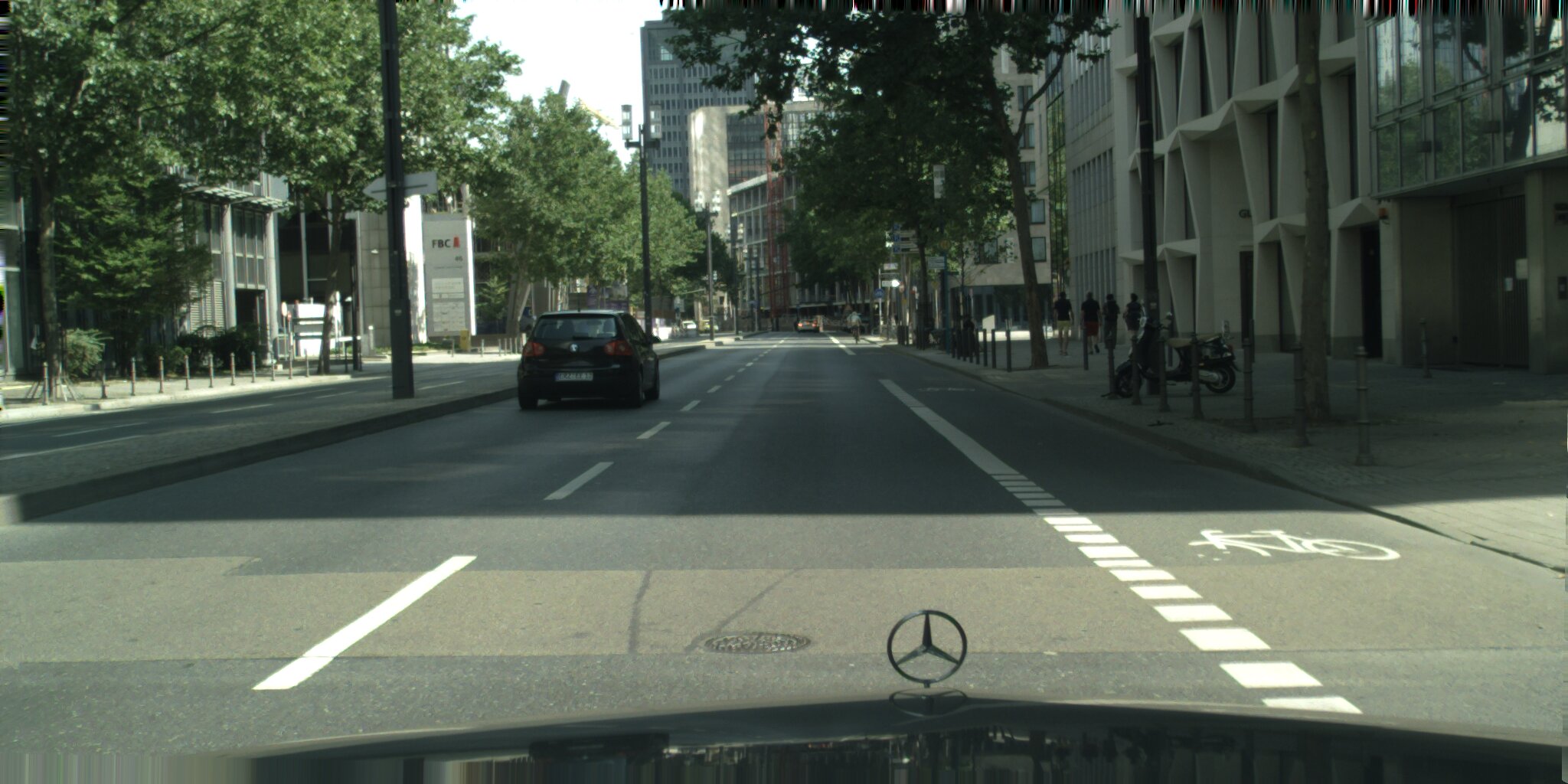}\quad
    \includegraphics[width=0.4\linewidth]{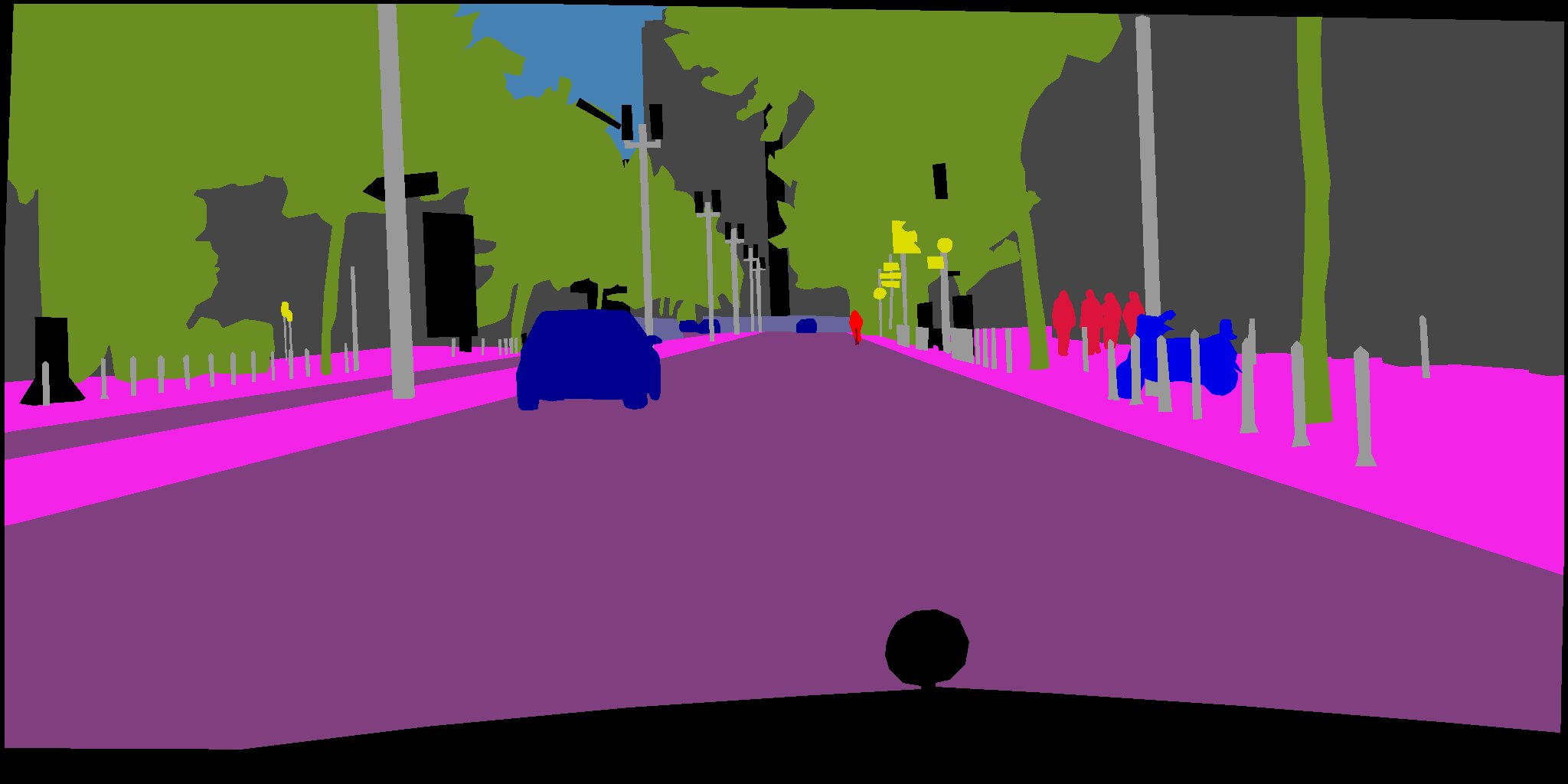}
    \\
    \raisebox{1cm}{(b)}\quad
    \includegraphics[width=0.4\linewidth,clip,trim=50 100 0 0]{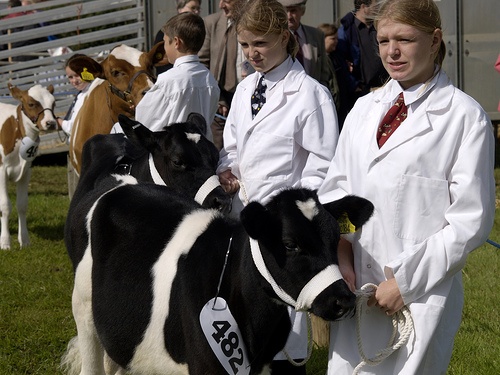}\quad
    \includegraphics[width=0.4\linewidth,clip,trim=50 100 0 0]{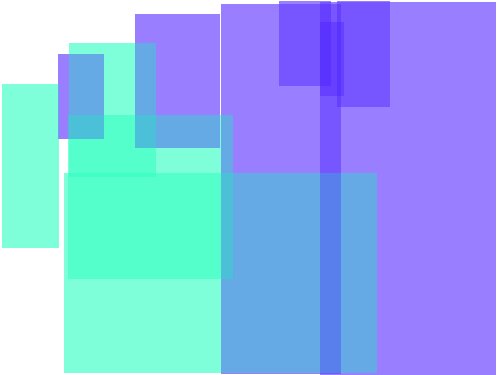}
    \\
    \raisebox{1cm}{(c)}\quad
    \includegraphics[width=0.4\linewidth,clip,trim=50 100 0 0]{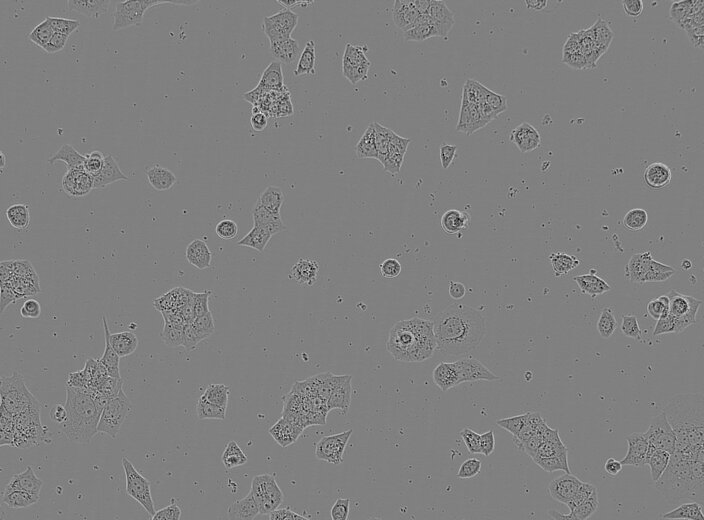}\quad
    \includegraphics[width=0.4\linewidth,clip,trim=50 100 0 0]{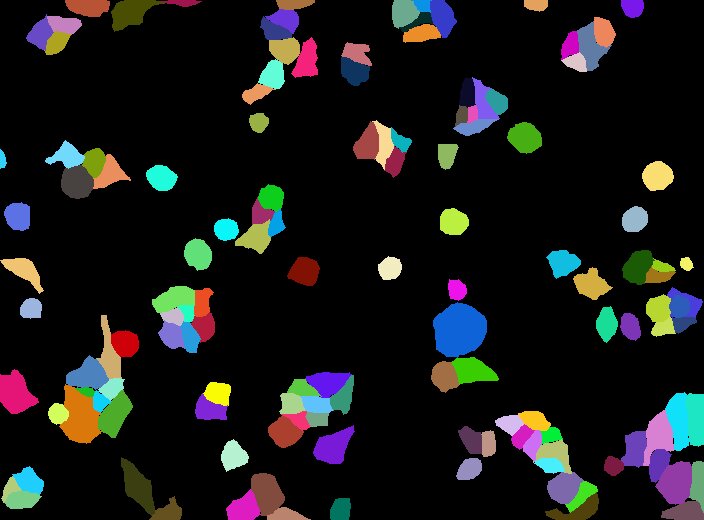}
     \caption{Overview of ground truth masks for semantic segmentation (a), object detection (b) and instance segmentation (c) datasets.}
    \label{fig:gtmasks}
\end{figure}
As our label error detector is an instance segmenter, label errors in object detection datasets are learned and predicted in rectangular bounding box shape. 

\textbf{Baselines.}
The \textit{naive baseline} investigates the predictions of the reference DNN. Those predictions that are false positive w.r.t.\ to ground truth $\gG$ under inspection serve as label error proposals, since they represent differences between the ground truth under inspection and the reference prediction. A prediction is false positive if it overlaps with an object of the same class in the ground truth by less than 0.3 IoU.
For object detection and instance segmentation we also consider a \textit{score baseline}, which differs from the naive baseline by taking into account the predicted score of the reference DNN, which is used for thresholding in the evaluation.
For object detection, we also consider the \textit{Loss inspection} method by \cite{Schubert2024Label}. This method is based on object detection loss at the instance level. The method depends on the architecture in which it was implemented, with the specific architecture determining the way in which it functions. Thus, we only apply it to Cascade R-CNN predictions as in the original work.
The final baseline for semantic segmentation datasets is the DNN + uncertainty quantification \textit{(DNN+UQ)} method developed by~\cite{Rottmann2023Label}. This method detects label errors based on connected components from a predicted semantic segmentation and UQ on component-level. A minimum object size of 100 pixels has been set for this method. In the following comparisons, where we used this baseline, we made sure that all predictions included were also at least 100 pixels in size in order to make the methods more comparable.

\begin{table*}[htbp]
\centering
\caption{Performance of the trained label error detector on perturbed datasets on Cityscapes ($q=0.2$, $t=1$0), ADE20K ($q=0.2$, $t=2$) and LIVECell ($q=0.1$, $t=1$). For evaluation we chose a score threshold of $0.5$ and an IoU threshold of $0.1$.}
\scalebox{0.9}{
\begin{tabular}{cccccccc}
\hline
\text{} & \textbf{AP} & \textbf{TPs} & \textbf{FPs} & \textbf{FNs} & \textbf{Precision} & \textbf{Recall} & \textbf{F1-Score} \\
\hline
\hline
& \multicolumn{7}{c}{\textit{Cityscapes - Semantic Segmentation}} \\
\hline
\hline
naive baseline & - & 1336 & 9554 & 1770 & 12.27 & 43.01 & 10.09\\
DNN+UQ & 4.47 & 1029 & 4640 & 2077 & 18.15 & 33.13 & 23.45\\
$f_\lerr(I_\rgb,I_\pert,I_\pred,I_\diff)$ & 61.74 & 2975 & 4423 & 131 & 40.21 & 95.78 & 56.65 \\
\hline
\hline
& \multicolumn{7}{c}{\textit{ADE20K - Semantic Segmentation}} \\
\hline
\hline
naive baseline & - & 1967 & 13,019 & 2169 & 13.13 & 47.56 & 20.57\\
DNN+UQ & 3.42 & 1292 & 4029 & 2844 & 24.28 & 31.24 & 27.32\\
$f_\lerr(I_\rgb,I_\pert,I_\pred,I_\diff)$ & 46.58 & 3405 & 6223 & 731 & 35.37 & 82.33 & 49.48 \\
\hline
\hline
& \multicolumn{7}{c}{\textit{LIVECell - Instance Segmentation}} \\
\hline
\hline
naive baseline & - & 21,888 & 91,254 & 22,229 & 19.35 & 49.61 & 27.84\\
score baseline & 11.79 & 15,138 & 16,382 & 28,979 & 48.03 & 34.31 & 40.03\\
$f_\lerr(I_\rgb,I_\pert,I_\pred)$ & 52.19 & 32,196 & 4657 & 11,921 & 87.36 & 72.98 & 79.53\\
\hline
\end{tabular}
}
\label{tab:segm_pert_results}
\end{table*}

\begin{table*}[!ht]
\centering
\caption{Performance of the trained label error detector on perturbed datasets on PascalVOC ($q=0.2$, $t=10$) and COCO ($q=0.2$, $t=2$). For evaluation we chose a score threshold of $0.5$ and an IoU threshold of $0.1$.}
\scalebox{0.9}{
\begin{tabular}{cccccccc}
\hline
\text{} & \textbf{AP} & \textbf{TPs} & \textbf{FPs} & \textbf{FNs} & \textbf{Precision} & \textbf{Recall} & \textbf{F1-Score} \\
\hline
\hline
& \multicolumn{7}{c}{\textit{PascalVOC - Object Detection}} \\
\hline
\hline
naive baseline & - & 1998 & 8251 & 968 & 19.49 & 67.36 & 30.24\\
score baseline & 24.23 & 1545 & 2547 & 1421 & 37.76 & 52.09 & 43.78\\
loss inspection & 1.35 & 720 & 2707 & 2246 & 21.01 & 24.28 & 22.52\\
$f_\lerr(I_\rgb,I_\pert,I_\pred,I_\diff)$ & 59.56 & 2533 & 3439 & 433 & 42.41 & 85.40 & 56.68 \\
\hline
\hline
& \multicolumn{7}{c}{\textit{COCO - Object Detection}} \\
\hline
\hline
naive baseline & - & 4744 & 15,252 & 2763 & 23.72 & 63.19 & 34.50\\
score baseline & 29.13 & 3272 & 1962 & 4235 & 62.51 & 43.59 & 51.36\\
loss inspection & 1.39 & 2667 & 10,531 & 4840 & 20.21 & 35.53 & 25.76 \\
$f_\lerr(I_\rgb,I_\pert,I_\pred,I_\diff)$ & 55.82 & 5722 & 5465 & 1785 & 51.15 & 76.22 & 61.22\\
\hline
\end{tabular}
}
\label{tab:od_pert_results}
\end{table*}
\textbf{Results for Simulated Label Errors.}
In this section, we present results for five datasets across the tasks semantic and instance segmentation (\cref{tab:segm_pert_results}) and object detection (\cref{tab:od_pert_results}). With one exception, our method outperforms state-of-the-art baselines across all datasets and evaluation metrics. The only exception is the precision value on the COCO dataset, where the score baseline achieves slightly superior results. However, in that case our method is clearly superior in terms of recall, F1 and AP. Taking a closer look, we see that our method sometimes overproduces false positives. Precision gives an indication of how much reviewing work needs to be invested to find a label error. A precision of 50\% means that every second proposal of our method is indeed a label error. The more important metric is probably recall, which is tightly related to the achievable annotation quality by any of the considered methods. A recall of 90\% means that 90\% of the label errors that we injected by our label perturbation algorithm can be detected. Note that TP, FP and FN metrics can still be traded. E.g.\ to achieve a higher recall, we can lower the score threshold, which of course increases the amount of reviewing work or in other words decreases the precision.
In some cases we outperform the previous state-of-the-art by enormous margins. E.g.\ in all segmentation tasks, the recall of our method exceeds the one of the DNN+UQ baseline by 38--62 percent points. Note, however, that the DNN+UQ baseline is not able to detect spawns, which means a default offset of 33 percent points between DNN+UQ and our method. Even when taking this offset into account, we outperform the DNN+UQ baseline.
We note that learning on a distribution that underwent the same perturbation algorithm as the test distribution might give our distribution a conceptual edge over the competition.
That is why we evaluate on real label errors in the Cityscapes dataset in the next section.

\textbf{Ablation Studies.}
In this section, we examine the performance results of trained label error detectors on simulated data and conduct ablation experiments on the Cityscapes dataset. To this end, we selected various training parameters, including altering the perturbation rate and the reference DNNs for the predicted masks, and varying the input, as broken down in table~\ref{tab:ablationstudies}. We performed evaluations of each model and baseline on the same perturbed validation dataset, created with a perturbation rate of $q=0.2$, inducing 3,106 artifical label errors.
\begin{table*}[htbp]
    \centering
    \caption{Ablationstudies and performance comparison on different training parameters of label error detector on perturbed label errors. Evaluation of all experiments on the same perturbed validation dataset with perturbation rate $q=0.2$.}
    \scalebox{0.9}{
        \begin{tabular}{cccccccc}
            \hline
            \text{} & \textbf{AP} & \textbf{TPs} & \textbf{FPs} & \textbf{FNs} & \textbf{Precision} & \textbf{Recall} & \textbf{F1-Score}\\
            \hline
            \hline
            naive baseline & - & 1336 & 9554 & 1770 & 12.27 & 43.01 & 10.09 \\
            DNN+UQ & 4.47 & 1029 & 4640 & 2077 & 18.15 & 33.13 & 23.45\\
            \hline
            \hline 
            $t$ & \multicolumn{7}{c}{\textit{perturbation frequency}}\\
            \hline             
            \hline
            1 & 56.61 & 2869 & 3638 & 237 & 44.09 & 92.37 & 59.69 \\
            5 & 61.28 & 2969 & 4326 & 137 & 40.70 & 95.59 & 57.09 \\
            10 & 61.74 & 2975 & 4423 & 131 & 40.21 & 95.78 & 56.65 \\          
            \hline
            \hline
             $q$ & \multicolumn{7}{c}{\textit{perturbation rate}}\\
            \hline
            \hline
            0.05 & 54.80 & 2534 & 4018 & 572 & 38.68 & 81.58 & 52.47 \\
            0.1 & 58.82 & 2920 & 4451 & 186 & 39.61 & 94.01 & 55.74 \\
            0.2 & 61.28 & 2969 & 4326 & 137 & 40.70 & 95.59 & 57.09 \\
            \hline
            \hline
            $f_\refe$ & \multicolumn{7}{c}{\textit{semantic segmentation reference network}}\\
            \hline
            \hline
            segformer & 61.74 & 2975 & 4423 & 131 & 40.21 & 95.78 & 56.65 \\
            DeepLabV3+ & 64.20 & 2950 & 3000 & 156 & 49.58 & 94.98 & 65.15 \\
            vltseg & 62.65 & 2973 & 3397 & 133 & 46.67 & 95.72 & 62.75 \\
            \hline
            \hline
            $f_\lerr$ & \multicolumn{7}{c}{\textit{instance segmentation network (label error predictor)}}\\
            \hline
            \hline
            Cascade Mask-RCNN & 61.74 & 2975 & 4423 & 131 & 40.21 & 95.78 & 56.65 \\
            Mask-RCNN & 61.64 & 3019 & 5911 & 87 & 33.81 & 97.20 & 50.17 \\
            \hline
            \hline
             & \multicolumn{7}{c}{\textit{training inputs}}\\
            \hline
            \hline
            $f_\lerr(I_\rgb,I_\pert,I_\pred,I_\diff)$ & 61.74 & 2975 & 4423 & 131 & 40.21 & 95.78 & 56.65 \\
            $f_\lerr(I_\rgb,I_\pert,I_\pred)$ & 61.60 & 2972 & 4302 & 134 & 40.86 & 95.69 & 57.26 \\
            $f_\lerr(I_\pert,I_\pred,I_\diff)$ & 62.89 & 2991 & 4285 & 115 & 41.11 & 96.30 & 57.62 \\
            \hline
        \end{tabular}
        }
        \label{tab:ablationstudies}
\end{table*}
Using our method, we outperformed the baselines in every training configuration and metric, except when using Mask-RCNN~\cite{maskrcnn} as label error detector. We argue that Cascade Mask-RCNN achieves higher precision, whereas the Mask-RCNN prioritizes higher recall. This is also reflected in the higher number of true positives (TPs) of the Mask-RCNN compared to the Cascade Mask-RCNN.
The results also demonstrate that the choice of the reference DNN can influence the performance of the label error detector. The DeepLabV3+~\cite{deeplab} network is often characterized by its high precision, which is shown in the results. Label error detection with DeepLabV3+ predictions achieves almost 10\% higher precision than segformer~\cite{segformer} when used as a reference network. Segformer on the other hand, has superior recall in direct comparison to the other two reference networks, achieving highest recall marginally in front of VLTSeg~\cite{vltseg}.
When the number of copies $t$ of the perturbed dataset in the training is varied, two things become apparent. Firstly, the number of false positives increases, resulting in lower precision and F1-Score. However, by perturbing the dataset several times and thus increasing diversity of the training dataset (at fixed perturbation rate $q=0.2$), we can achieve a higher AP and recall. An increasing perturbation rate $q$ with constant $t=5$ has a positive effect on all metrics. Although the number of false positives grows, the simultaneous increase in TPs means that precision, recall and the F1-Score are still improved.
The performance results are surprising when we change the composition of the input tensor used to train the label error detector. In fact, we achieve the best results when we do not use the RGB input image. Directly comparing training $f_\lerr(I_\rgb,I_\pert,I_\pred)$ and $f_\lerr(I_\pert,I_\pred,I_\diff)$ shows that the label error detector performs better when trained with a difference mask than with an RGB image, even if the differences are marginal.
In summary, there are essentially no significant differences in performance, and all trained label detectors perform well regardless of training parameters. However, the influence of varying input and choice of reference network offer opportunities for finetuning to a specific use case.

\begin{table*}[!ht]
\centering
\caption{Evaluation on real label errors in Cityscapes. Our label error detector was trained on $t=10$ copies of Cityscapes with perturbation rate $q=0.2$.}
\scalebox{0.9}{
\begin{tabular}{cccccccc}
\hline
\text{} & \textbf{AP} & \textbf{TPs} & \textbf{FPs} & \textbf{FNs} & \textbf{Precision} & \textbf{Recall} & \textbf{F1-Score} \\
\hline
\hline
naive baseline & - & 294 & 8462 & 165 & 3.36 & 64.05 & 6.38\\
DNN+UQ & 4.13 & 274 & 4038 & 185 & 6.35 & 59.69 & 11.49\\
$f_\lerr(I_\rgb,I_\org,I_\pred,I_\diff)$ & 5.46 & 323 & 4568 & 136 & 6.60 & 70.37 & 12.07\\
\hline
\end{tabular}
}
\label{tab:cityscapes_real_results}
\end{table*}

\begin{figure}[htbp]
    \centering
    \includegraphics[scale=0.21,clip,trim=1600 700 0 0]{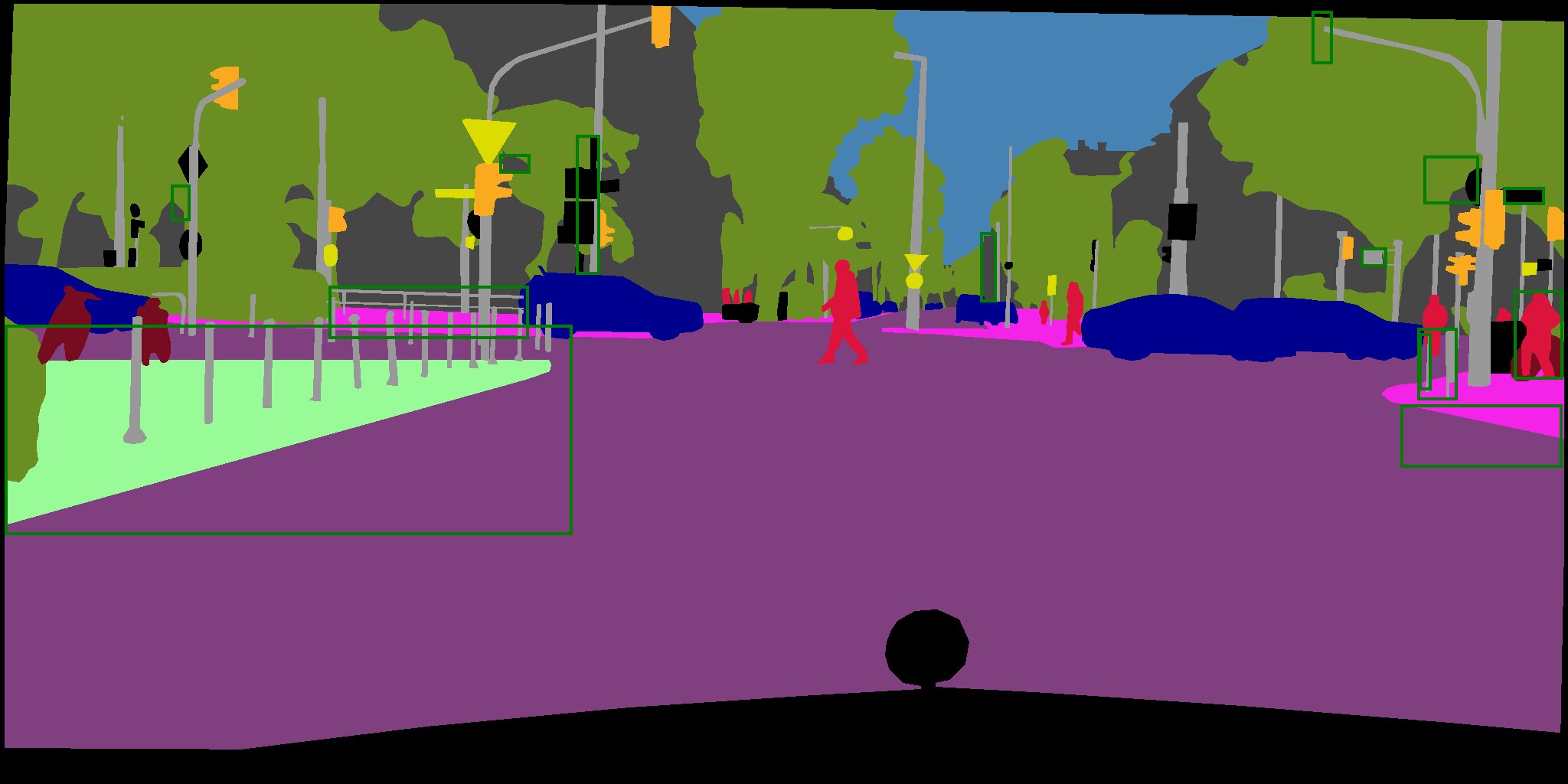}
    \includegraphics[scale=0.21,clip,trim=1600 700 0 0]{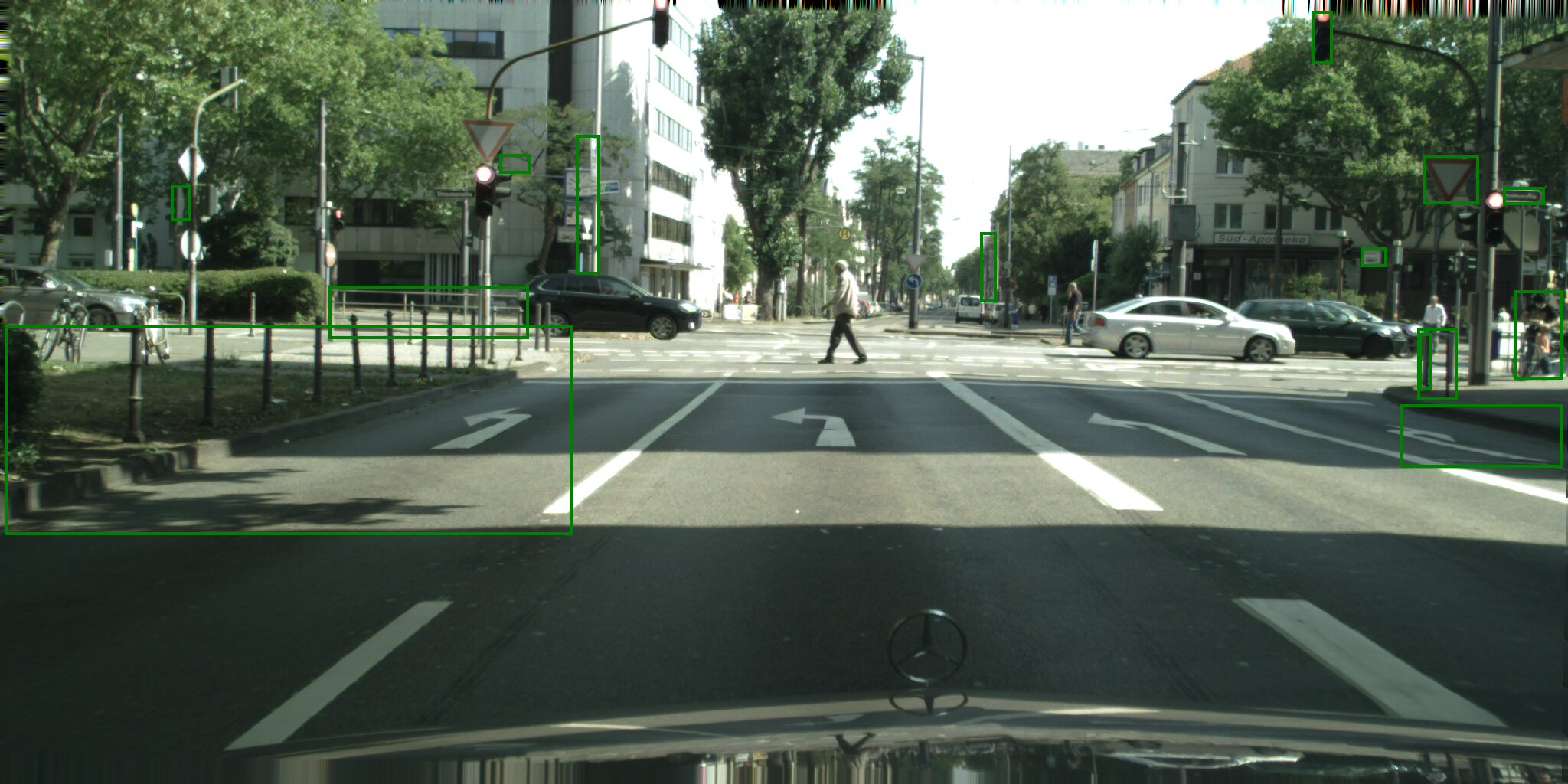}
    \includegraphics[scale=0.25,clip,trim=100 160 40 0]{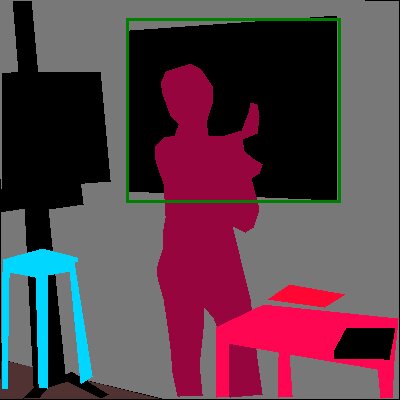}
    \includegraphics[scale=0.25,clip,trim=100 160 40 0]{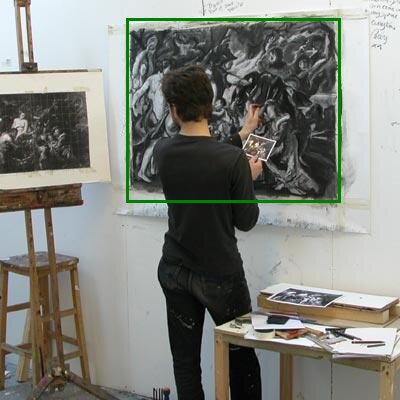}
    \includegraphics[scale=0.21,clip,trim=300 250 180 0]{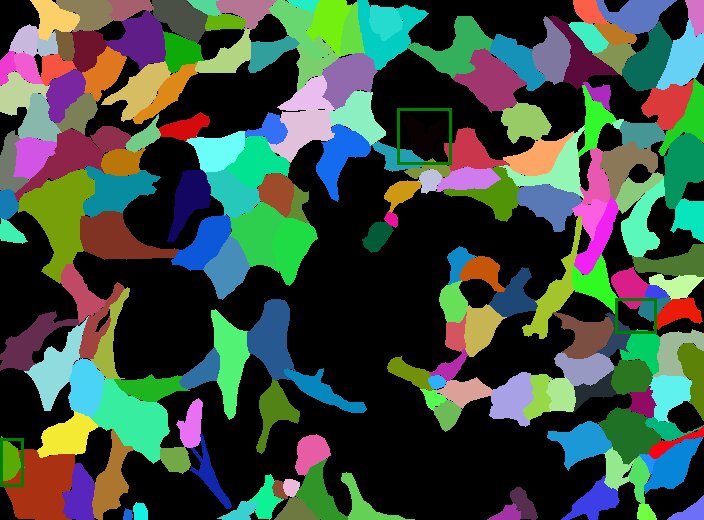}
    \includegraphics[scale=0.21,clip,trim=300 250 180 0]{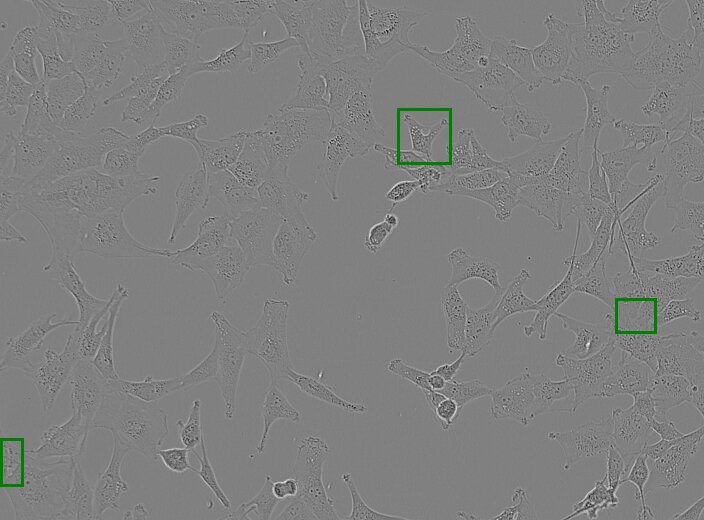}
    \includegraphics[scale=0.45,clip,trim=280 100 0 250]{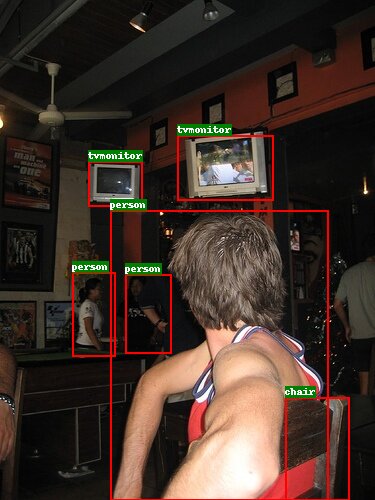}
    \includegraphics[scale=0.45,clip,trim=280 100 0 250]{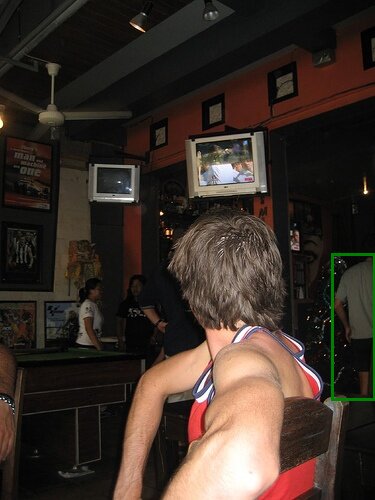}
    \includegraphics[scale=0.25, clip,trim=0 100 350 100]{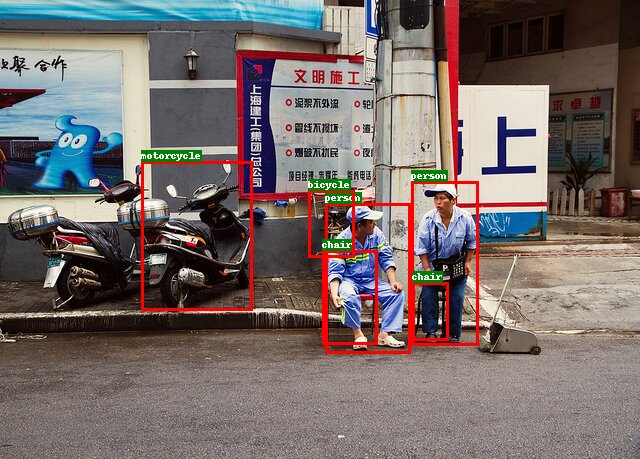}
    \includegraphics[scale=0.25,clip,trim=0 100 350 100]{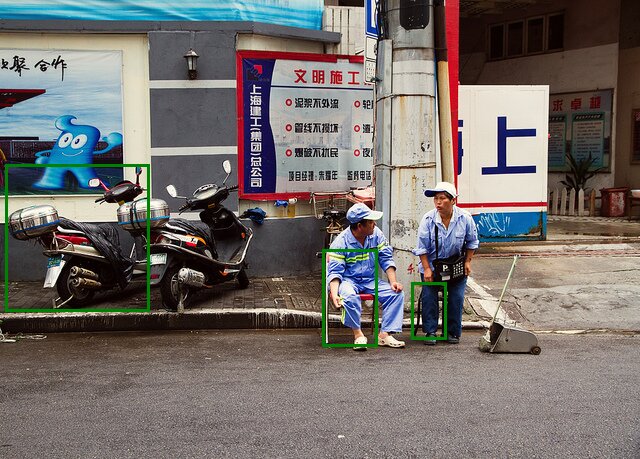}

    \caption{Examples of found real label errors in Cityscapes, ADE20K, LIVECell, PascalVOC und COCO (left to right). The ground truth and RGB images are shown alternately. The segmentation masks and the RGB images with red boxes show the GT annotations. In green are the label error proposals of our trained label error detector.}
    \label{fig:real_label_errs}
\end{figure}

\begin{table}[!ht]
\centering
\caption{Found real label errors predicted by our label error detector with a score threshold of $0.75$ of 200 randomly chosen images of validation dataset.}
\resizebox{\linewidth}{!}{
\begin{tabular}{cccccccc}
\hline
\textbf{Dataset} & \textbf{\#Label Errors} & \textbf{Precision} & \textbf{\#Predictions} \\
\hline
ADE20K & 21 & 5.57 & 377\\
LIVECell & 95 & 29.78 & 319\\
PascalVOC & 13 & 18.57 & 70\\
COCO & 35 & 41.12 & 85\\
\hline 
\end{tabular}
}
\label{tab:count_real}
\end{table}

\textbf{Results for Real Label Errors.}
The evaluation of our label error method in detecting real label errors in Cityscapes, ADE20K, PascalVOC, COCO and LiveCell is the focus of the final section on numerical experiments.
First, we created a label error ground truth of real label errors in the Cityscapes validation dataset by manually reviewing label error proposals from the DNN+UQ method~\cite{Rottmann2023Label} and our own method. We detected 228 label errors using DNN+UQ, and additional 231 using our method. At this point, we would like to emphasize that this is only a selection of real label errors in the Cityscapes validation dataset and is intended only as a basis for evaluation. We do not claim that all existing label errors have been detected. It is possible that more real label errors may be found when manually reviewing additional label error proposals.
To get proposals for possible real label errors, we change the input from the trained label error detector from $f_\lerr(I_\rgb,I_\pert,I_\pred,I_\diff)$ to $f_\lerr(I_\rgb,I_\gt,I_\pred,I_\diff)$, where the difference mask $I_\diff$ in combination with $I_\gt$ shows now the differences between the ground truth mask $I_\gt$ and the prediction mask $I_\pred$. In table~\ref{tab:cityscapes_real_results} the results of our label error method in comparison to the semantic segmentation baseline \emph{naive baseline} and \emph{DNN+UQ} are presented. Despite the higher number of false positives in our method, we can still outperform the baselines in particular in precision and recall.
Real label errors were also found in the ADE20K, LIVECell, PascalVOC and COCO datasets. Examples are presented in figure~\ref{fig:real_label_errs}. As a study, we randomly selected 200 images from validation set of each dataset. We then used our label error detector to predict label error proposals with a score threshold of $0.75$, and counted how many of these were real ones. The results are shown in table~\ref{tab:count_real}. Most of the label errors were found in the COCO and LIVECell datasets, whereas the PascalVOC dataset was well labeled from the very beginning. The ADE20K dataset was difficult to review due to its 150 different classes. However, at least 21 real label errors were found, although there could possibly be more.

\section{Conclusion}
In this work, we proposed the first method to detect label errors in object detection, semantic and instance segmentation datasets by learning detection from simulated label errors. 
For the label error simulation, we developed a perturbation algorithm to simulate three different label error types. In numerical experiments on simulated and real label errors, our label error detector method outperforms common baselines and existing state-of-the-art label error detection methods. Moreover, we introduce a benchmark for label error detection with an updated version of the Cityscapes ground truth.

\section*{Acknowledgements}
We thank Miriam Ackermann and Anna Hövermann for the label error review and we thank Marcel Schilling for discussion and useful advice.
S.P.\ and M.R.\ acknowledge support by the German Federal Ministry of Research, Technology and Space (BMFTR) within the junior research group project ``UnrEAL'' (grant no.\ 01IS22069).


\clearpage
\appendix

\subsection{Details on Perturbation Algorithm}

\begin{algorithm}[t]
\caption{Perturbation Algorithm Pseudocode}
\label{alg:pertalg}
\begin{algorithmic}
\State \textbf{Input:} ground truth annotations $\gG_\org=\{p_1^\org,\ldots,p_N^\org\}$, perturbation rate $q$, polygon type $k$, $I_\org$ size $(w,h)$, radius range $\bar{r}=(r_1,r_2)$ with $r_1<r_2$, spikeness $s\in(0,1)$, irregularity $i\in(0,1)$, maximal number vertices $v_{\textit{m}}$
\State \textbf{Output:} perturbed ground truth annotations $\gG_\pert=\{p_1^\pert,\ldots,p_{N'}^\pert\}$, label error ground truth  $\gG_\legt\{p_1^\legt,\ldots,p_{M}^\legt\}$
\State $\gG_\pert=\gG_\legt = \emptyset$
\For{$i=1$ to $N$}
    \State \# Bernoulli experiment
    \State Sample $X\sim\textit{Bernoulli}(q)$
    \If{$X= \mathrm{true}$}
        \State \# uniformly choose label error type
        \State Sample $T\sim\mathcal{U}(\{\textit{drop}, \textit{flip}, \textit{spawn}\})$
        \If{$T=\textit{drop}$}
            \State $\gG_\legt = \gG_\legt \cup \{ p_i^\org \}$
        \ElsIf{$T=\textit{flip}$}
           \State $p_i^\pert=\textit{ChangeClass}(p_i^\org)$
           \State $\gG_\pert=\gG_\pert\cup\{p_i^\pert\}$
           \State $\gG_\legt=\gG_\legt\cup\{p_i^\pert\}$
        \ElsIf{$T=\textit{spawn}$}
            \If{$k=\textit{rectangle}$}
            \State $p_{\textit{spawn}}=\textit{GenerateSpawnRectangle}(w,h)$
            \Else
                \State $p_{\textit{spawn}}=\textit{GenerateSpawn}(w,h,\bar{r},s,i,v_{\textit{m}})$
            \EndIf
            \State $\gG_\pert=\gG_\pert\cup \{p_i^\org, p_{\textit{spawn}}\}$
            \State $\gG_\legt=\gG_\legt\cup \{p_{\textit{spawn}}\}$
        \Else
            \State $\gG_\pert=\gG_\pert\cup\{p_i^\org\}$ 
        \EndIf
    \EndIf
\EndFor
\end{algorithmic}
\end{algorithm}
\begin{algorithm}[!t]
    \caption*{\textbf{Function} \textit{GeneratePolygon()}}
    \begin{algorithmic}
        \Function{\textit{GeneratePolygon}}{$((x_\cent,y_\cent),r,i,s,v_\numb)$}
    \State $i=(i\cdot2\cdot\pi) / v_\numb$
    \State $s=s\cdot r$
    \State \# generate list of random angles
    \State $\steps = \{\}$
    \State $l=(2\cdot\pi/v_\numb)-i$
    \State $u=(2\cdot\pi/v_\numb)+i$
    \State $c=0$
    \For{$m=0$ in $v_\numb$}
        \State \# uniformly choose angles
        \State $a\sim\mathcal{U}([l,u])$
        \State $\steps=\steps\cup\{a\}$
        \State $c=c+a$
    \EndFor
    \State $c=c/2\pi$
    \For{$a$ in $\steps$}
        \State $a=a/c$
    \EndFor
    \State \# uniformly choose start angle
    \State $a_\start\sim\mathcal{U}([0,2\pi])$
    \State \# collect edges of polygon
    \State $e = \{\}$
    \For{$j=0$ to $v_\numb$}
        \State $g\sim\mathcal{N}((r,s)$
        \State $r_\new=\min(2\cdot r, \max(g,0))$
        \State $e = e\cup\{[x_\cent+r_\new\cdot\cos(a),y_\cent+r_\new\cdot\sin(a)]\}$
        \State $a_\start = a_\start + \steps[j]$
    \EndFor
    \State \Return $e$
    \EndFunction
    \end{algorithmic}
\end{algorithm}
\begin{algorithm}[!t]
    \caption*{\textbf{Function} \textit{GenerateSpawn()}}
    \begin{algorithmic}
    \Function{\textit{GenerateSpawn}}{($w,h,(r_1,r_2),s,i,v_{\textit{m}}$)}
    \State \# generate random polygon with multiple vertices
    \State sample $r\sim\mathcal{U}(\{r_1,r_1+1,\ldots, r_2-1, r_2\})$
    \State \# uniformly choose center of polygon
    \State sample $x_\cent\sim\mathcal{U}(\{r,r+1,\ldots,w-r-1,w-r\})$
    \State sample $y_\cent\sim\mathcal{U}(\{r,r+1,\ldots,h-r-1,h-r\})$
    \State \# uniformly choose number of vertices
    \State \# polygon requires at least 4 coordinates
    \State $v_\numb\sim\mathcal{U}([4, 5,\ldots,v_{\textit{m}}])$
    \State $p_\spawn=\textit{GeneratePolygon}((x_\cent,y_\cent),r,i,s,v_\numb)$
    \State \# uniformly choose a class
    \State $c_\s\sim\mathcal{U}(\{0,\ldots,c-1\})$
    \State $p_\spawn\leftarrow c_\s$
    \State \Return $p_\spawn$
    \EndFunction
    \State
    \Function{\textit{GenerateSpawnRectangle}}{($w,h$)}
    \State \# generate quadratically spawn
    \State Sample $w_\s\sim\mathcal{U}([0,w])$
    \State Sample $h_\s\sim\mathcal{U}([0,h])$
    \State Sample $x_\s\sim\mathcal{U}([0,w-w_\s])$
    \State Sample $y_\s\sim\mathcal{U}([0,h-h_\s])$
    \State $e_1=[x_\s, y_\s]$
    \State $e_2=[x_\s+w_\s,y_\s]$
    \State $e_3=[x_\s+w_\s,y_\s+h_\s]$
    \State $e_4=[x_\s,y_\s+h_\s]$
    \State $p_\spawn = [e_1,e_2,e_3,e_4]$
    \State \# uniformly choose a class
    \State $c_\s\sim\mathcal{U}(\{0,\ldots,c-1\})$
    \State $p_\spawn\leftarrow c_\s$
    \State \Return $p_\spawn$
    \EndFunction
    \State
    \Function{\textit{ChangeClass}}{$p_i^\org$}
    \State \# $p_i^\org$ with class $c_i^\org$
    \State \# uniformly choose a new class
    \State Sample $\bar{c}\sim\mathcal{U}(\{0,\ldots,c-1\})\setminus\{c_i^\org\}$
    \State $p_i^\org\leftarrow\bar{c}$
    \State $p_i^\pert\leftarrow p_i^\org$
    \State \Return $p_i^\pert$
    \EndFunction
    \end{algorithmic}
\end{algorithm}

In this section, we discuss the label perturbation algorithm in greater detail. The algorithm has been outlined in the main part of the manuscript. The pseudo-code in~\cref{alg:pertalg} of the label perturbation algorithm is provided for the sake of completeness and contains additional sub-routines (\texttt{ChangeClass} performing label flips, \texttt{GenerateSpawnRectangle}, as well as \texttt{GenerateSpawn} performing label spawns) which have not been discussed in the main part of the paper. 

\textbf{Flip.} If the label error type \textit{flip} is chosen, then the class of the respective polygon is changed. To do this, we randomly select a new class from the set of all classes, excluding the current class of this polygon. After assigning a new class, we obtain our perturbed polygon.

\textbf{Spawn.} If the label error type sampled in the label perturbation algorithm is of type \textit{spawn}, we follow two different approaches to generate spawns. If the respective dataset is from object detection, the generated spawns has to be rectangular. Therefore, we uniformly sample the width and height of the rectangle, and depending on these values, we uniformly select the coordinates of the bottom left edge of the spawn. Using these coordinates, we can create a randomly generated polygon for object detection datasets. 

If the dataset is from a segmentation dataset, the form of this spawn has to have more vertices. The generation is fully random here as well. The code for spawn generation in segmentation datasets was copied in its entirety from~\cite{code}. For this polygon, we select fixed parameters for the radius range, spikiness and irregularity, as well as the maximum number of vertices. Irregularity describes the variance in the spacing of the angles between consecutive vertices. Spikeness refers to the variance in the distance of each vertex from the centre of the circumference. We uniformly sample a radius from the radius range, and depending on this parameter, as well as the image width and height, we uniformly sample the coordinates of the spawn's centre. Based on these parameters, we can generate the polygon. Starting from the centre of the polygon, we create it by sampling points on a circle around the centre. Random noise is added by varying the angular spacing between sequential points and the radial distance of each point from the centre. The resulting edges form the spawned polygon.

\begin{figure}[htbp]
    \centering
    \begin{subfigure}{0.48\columnwidth}
        \centering
        \includegraphics[width=\linewidth]{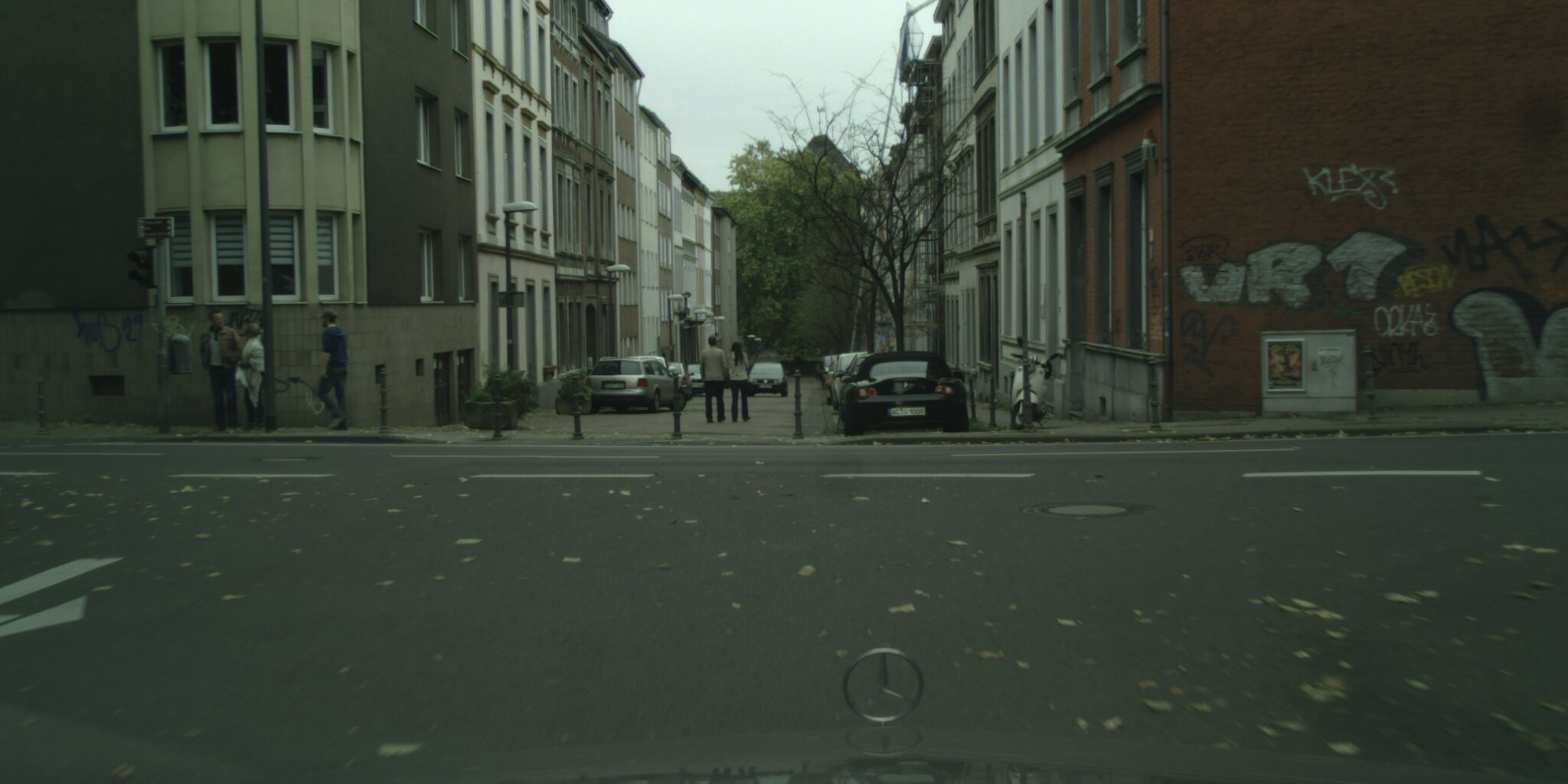}
        \caption*{input image}
        \label{fig:rgb}
    \end{subfigure}
    \hfill
    \begin{subfigure}{0.48\columnwidth}
        \centering
        \includegraphics[width=\linewidth]{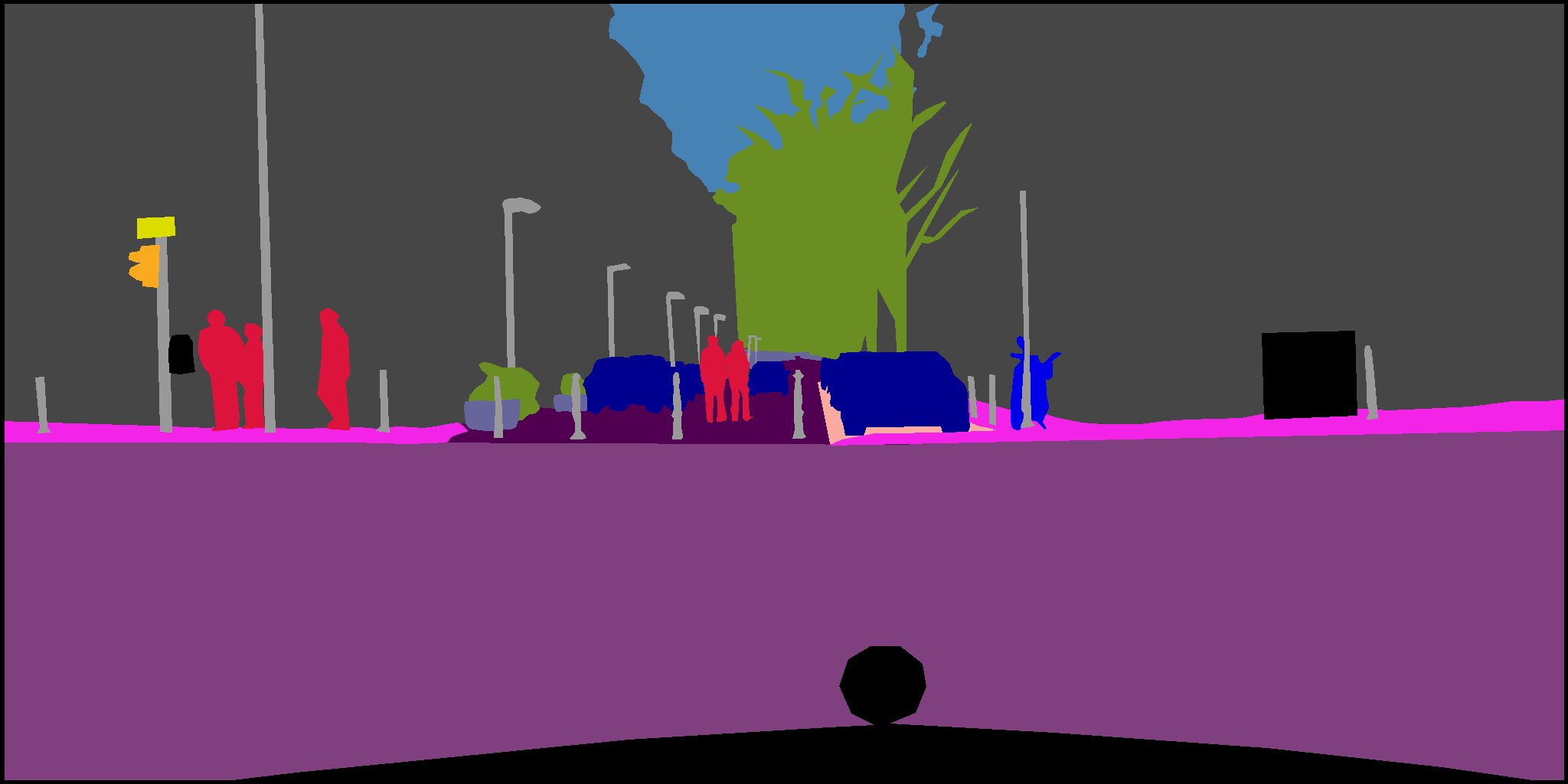}
        \caption*{ground truth mask}
        \label{fig:gtmask}
    \end{subfigure}

    \vspace{0.3cm}

    \begin{subfigure}{0.48\columnwidth}
        \centering
        \includegraphics[width=\linewidth]{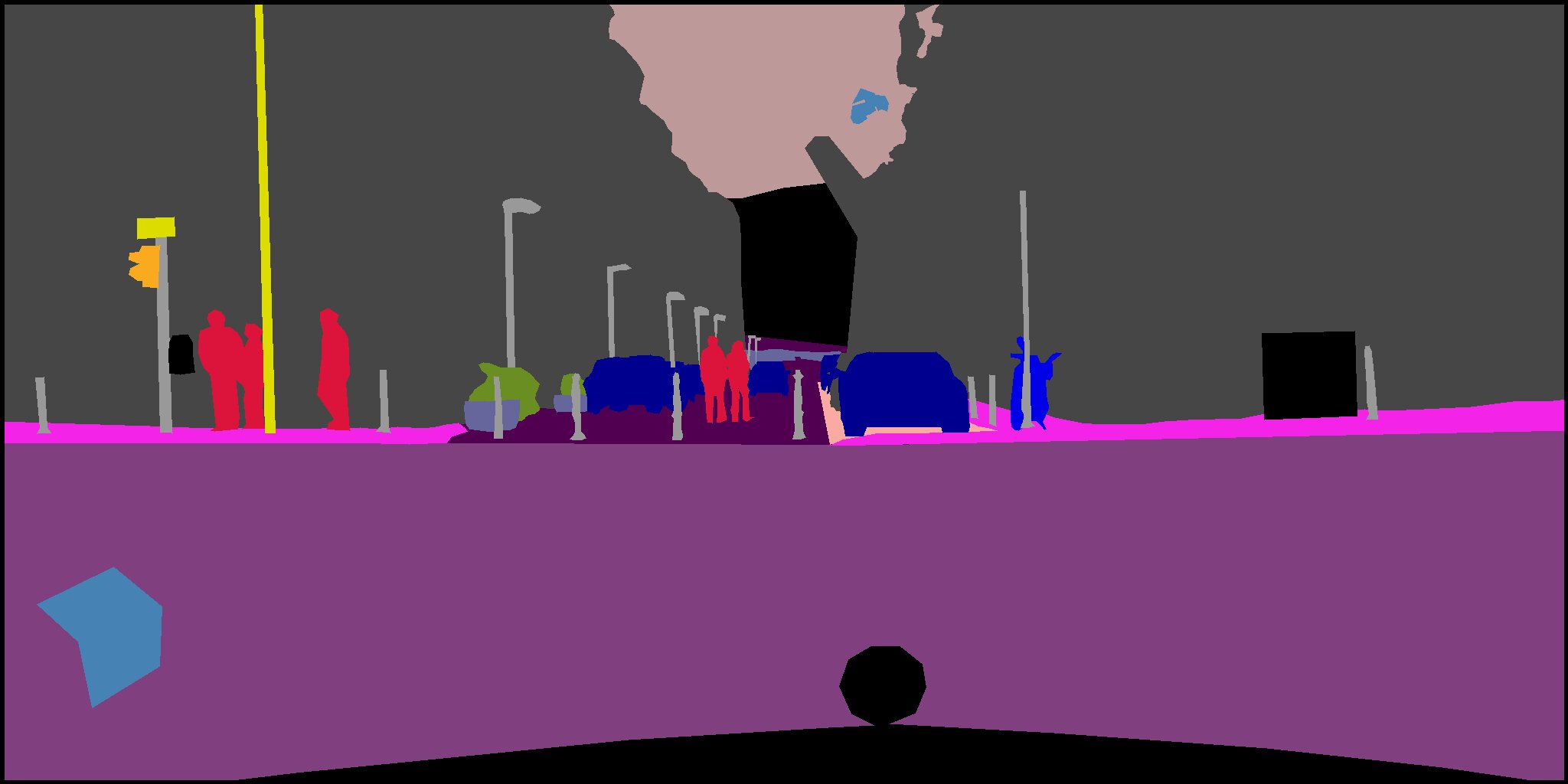}
        \caption*{1st perturbed mask}
        \label{fig:pertmask1}
    \end{subfigure}
    \hfill
    \begin{subfigure}{0.48\columnwidth}
        \centering
        \includegraphics[width=\linewidth]{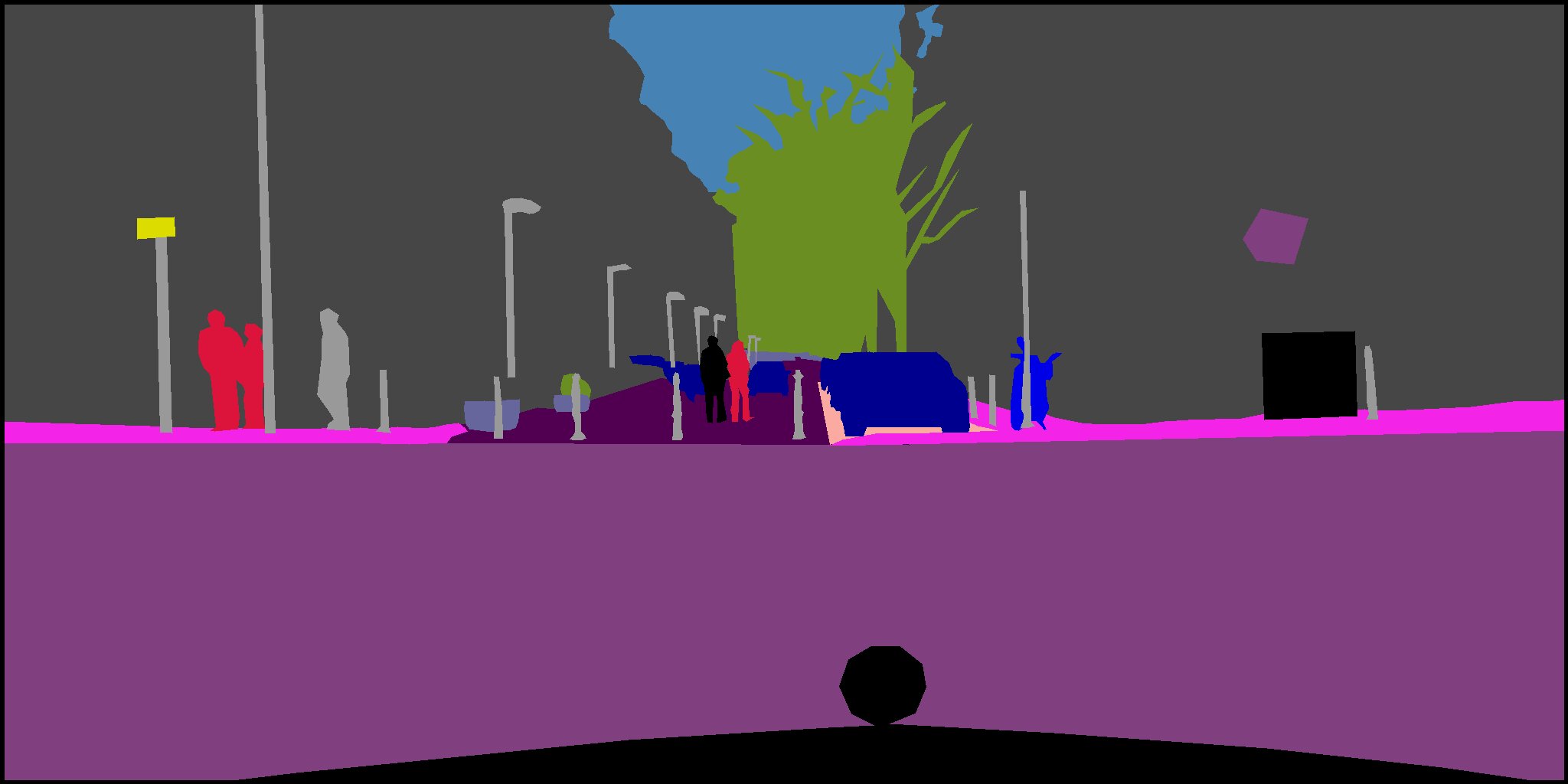}
        \caption*{2nd perturbed mask}
        \label{fig:pertmask2}
    \end{subfigure}

    \vspace{0.3cm}

    \begin{subfigure}{0.48\columnwidth}
        \centering
        \includegraphics[width=\linewidth]{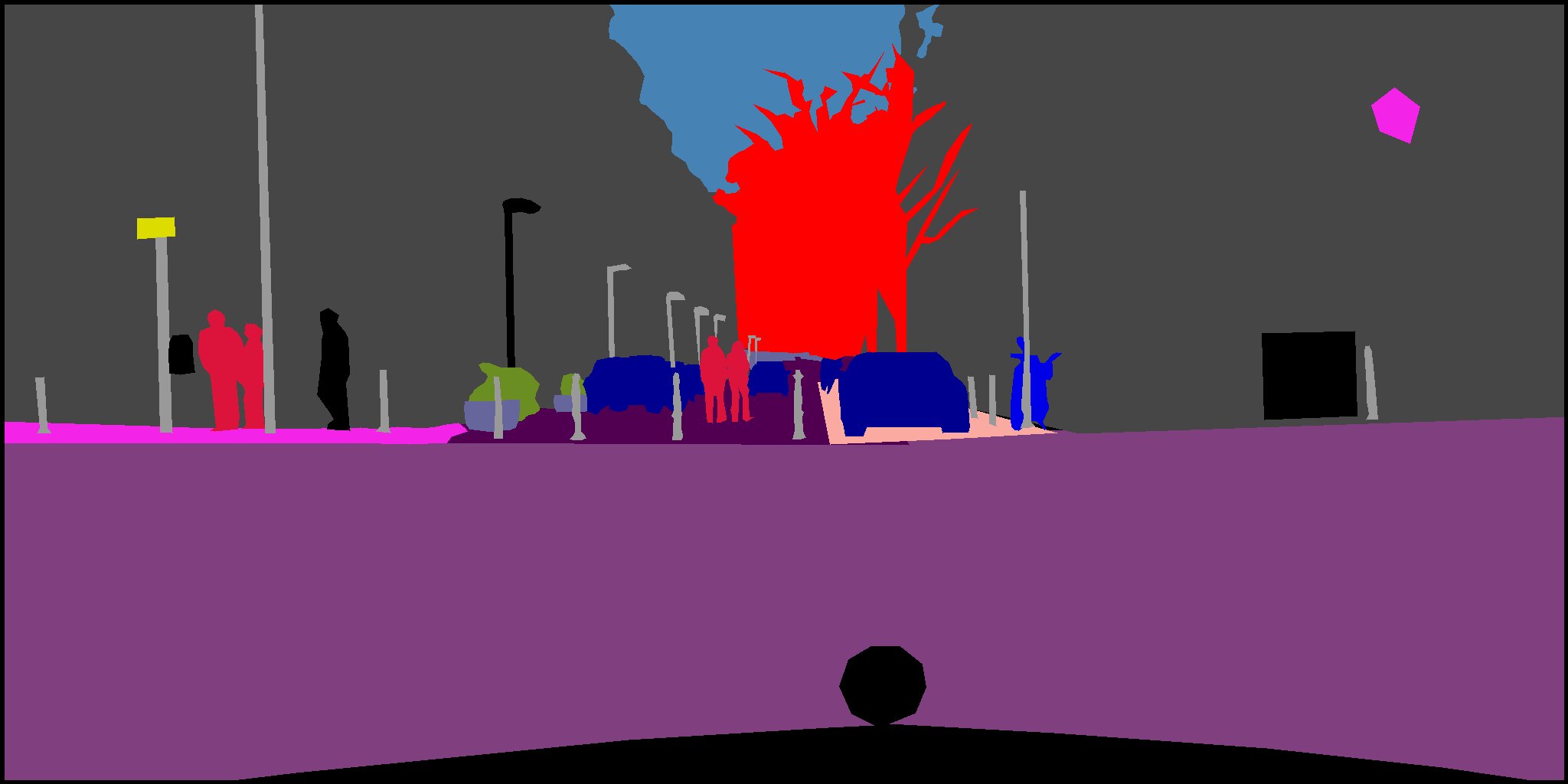}
        \caption*{3rd perturbed mask}
        \label{fig:pertmask3}
    \end{subfigure}
    \hfill
    \begin{subfigure}{0.48\columnwidth}
        \centering
        \includegraphics[width=\linewidth]{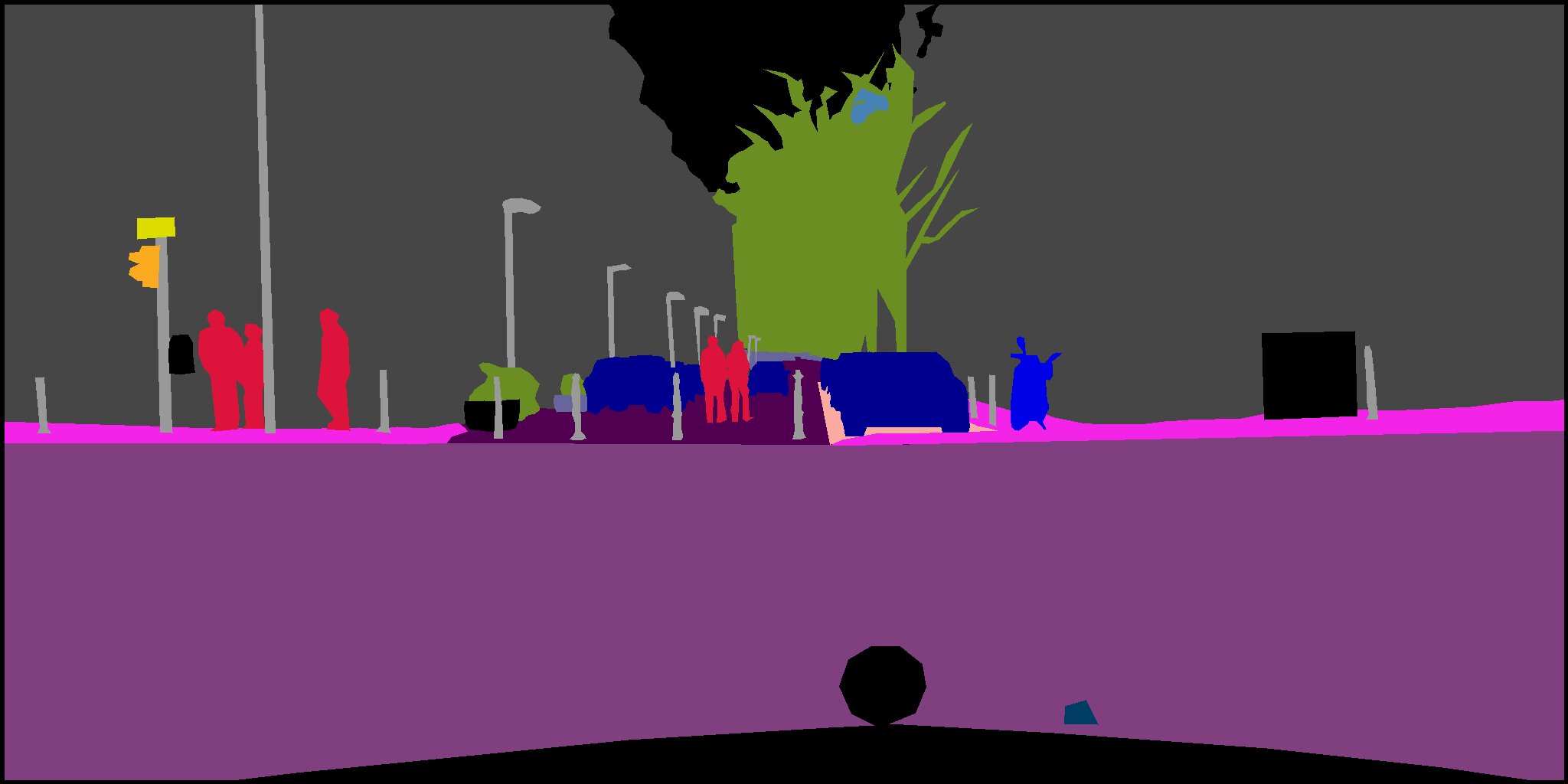}
        \caption*{4th perturbed mask}
        \label{fig:pertmask4}
    \end{subfigure}

    \caption{Results of multiple application of perturbation algorithm}
    \label{fig:pertalg_frequ}
\end{figure}

For label error detection, we apply the label perturbation algorithm multiple times to a given dataset as additional data augmentation for training data. This results in a number of copies of the dataset with different label errors. An overview of possible results are shown in \cref{fig:pertalg_frequ}.

\newpage
\subsection{Implementation Details}

\begin{table}[!ht]
\centering
\caption{An overview of the dataset implementations in relation to the perturbed dataset and the reference networks used.}
\resizebox{\linewidth}{!}{%
\begin{tabular}{cccccc}
\hline
& & &\multicolumn{2}{c}{no.\ of label errors} &\\
\hline 
\text{Dataset} & $q$ & $t$ & \textbf{train} & \textbf{val} & \textbf{reference DNNs}\\
\hline
Cityscapes & 0.2 & 10 & 194,635 & 3558 & segformer\\
ADE20K & 0.2 & 2 & 79,077 & 4136 & mask2former\\
PascalVOC & 0.2 & 10 & 94,202 & 2911 & GroundingDino\\
COCO & 0.2 & 2 & 344,709 & 7507 & Co-DETR\\
LIVECell & 0.1 & 1 & 117,799 & 44,331 & ResNeSt\\
\hline
\end{tabular}%
}
\label{tab:implementation}
\end{table}

\subsubsection{Datasets}

\paragraph{PascalVOC.} The PascalVOC dataset contains images of everyday scenes with annotations for 20 classes and is used among other as object detection dataset. We used the \textit{train+val2007} and \textit{train+val2012} datasets, comprising a total of 16,551 images, for training, and the \textit{test2007} dataset, comprising 4,952 images, for validation. We chose a perturbation rate of $q=0.2$ and created $t=10$ copies of the training datasets, resulting in approximately 97,000 simulated label errors for training and validation. Moreover we did not set any size restriction for objects due to the partly limited number of boxes per image. As reference network we opted for GroundingDino with a box threshold of $0.35$ and a text threshold of $0.3$. These parameters were chosen to prevent predictions with duplicate labels. Additionally, we applied Non-Maximum Suppression (NMS) to delete predicted boxes of the same class, as well as boxes of different classes that overlap significantly with an intersection over union (IoU) of at least $0.7$. Of the overlapping boxes of different classes, the box with the higher score was chosen.
\paragraph{COCO.} The object detection dataset COCO is characterized by its large number of images representing everyday scenes and categories. There are 80 different classes annotated on 118,287 images in the training set and 5,000 images in the validation set. The perturbed dataset was created using $t=2$ copies and a perturbation rate of $q=0.2$. In contrast to PascalVOC, we applied the perturbation algorithm only twice, given the already high number of images. The Co-DETR object detector was used for reference COCO predictions. It is based on a CoDETR backbone, a CoDINOHead as query head and a CoATSSHead as bounding box head. The pre-trained checkpoint we used achieves a box AP of $66.0$ and we collected all predictions with a score of at least $0.3$.
\paragraph{Cityscapes.} The Cityscapes semantic segmentation dataset contains high-resolution images of street scenes with 19 annotated classes. For training and validation, we use the 2,975 training images and 500 validation images, as well as generating different perturbed datasets for the ablation studies. The comparisons with baselines were performed with a perturbation rate of $q=0.2$ and $t=10$ copies of the training set. This dataset consists of images containing many annotated objects. The more objects in an image, the more likely label errors are to appear. Thus, we opted for a perturbation rate of 20\%, which is a realistic number of possible label errors. For the spawn generation, we decided for a radius range of $\bar{r}=(10,80)$, an irregularity $i=0.35$ and a spikeness of $s=0.2$. For the ablation studies, we selected segformer, DeeplabV3+ and VLTSeg as reference networks. The pre-trained segformer checkpoint achieves an mIoU of 82.25\%, and the DeepLabV3+ checkpoint achieves a value of 79.88\%. The VLTSeg checkpoint achieves an mIoU of 86.30\% on Cityscapes.
\paragraph{ADE20K.} The ADE20K semantic segmentation dataset of everyday scenes contains annotated objects from 150 different classes. We simulated label errors in 20,210 training and 2,000 validation images by applying the perturbation algorithm with a perturbation rate of $q=0.2$, a number of copies of $t=2$, a radius range of $\bar{r}=(10,30)$, irregularity of $i=0.35$ and spikeness $s=0.05$. The reduced number of copies compared to Cityscapes is again due to the large number of training images already available. All 150 classes were incorporated into the label perturbation algorithm. The pre-trained Mask R-CNN checkpoint, which we used for reference predictions, consists of a D2SwinTransformer backbone model and achieves an mIoU of 53.9\%.
\paragraph{LIVECell.} We chose the LIVECell dataset, consisting of images with eight different cell types that are all categorized to class \textit{cell}, as the instance segmentation dataset for our experiments. For training, we combined the training and validation data to create a single set of 3,727 images, and for validation, we used the 1,512 test images. We applied the label perturbation algorithm with perturbation rate of $q=0.1$ and applying the algorithm only once $t=1$ with the same parameters for spawn generation as in ADE20K. The reasons for this are twofold: firstly, due to the high number of objects, applying the algorithm more frequently would take a long time, and secondly, due to the similarity of the images, where more copies do not introduce any variance. Due to the dataset only containing a the single class, we only simulated \textit{drops} and \textit{spawns}. The reference predictions are based on the anchor-based model from the Detectron2-ResNeSt architecture. It achieves a mask mAP of 47.89\%. We collected all predictions with a score threshold of at least $0.1$.

\subsubsection{Training Details}
We opted for the Cascade Mask-RCNN for training on simulated label errors. Training all the datasets took 150,000 iterations, with a batch size of 10, and the solver made steps after 80,000 and 120,000 iterations reducing the learning rate by a factor of $0.1$.  
In addition to creating multiple perturbed copies of the original dataset by the label perturbation algorithm, we applied further data augmentations such as random flip and rotation, random crop and resizing of the shortest edge. Training of all datasets was performed with a learning rate of $0.01$, except for Cityscapes where a learning rate of $0.001$ was chosen. LIVECell is the only dataset for which we trained without a difference mask. For all the other datasets, we opted to train with an input image $I_\rgb$, a perturbed mask $I_\pert$, a predicted mask $I_\pred$ and a difference mask $I_\diff$. This choice turned out most beneficial in our experiments.

Another configuration that we adjusted before training was the input image normalization. For training purposes, the input images' pixel values were normalized by subtracting the mean and dividing by the standard deviation of the images.
Since we have nine or twelve channels, respectively, 
we perform the normalization of each channel individually. For training our label error detector, we computed the mean and standard deviation of the input image $I_\rgb$ and the perturbed mask $I_\pert$. Then, we set the same values for the perturbed mask $I_\pert$, the prediction mask $I_\pred$ and, if used, the difference mask $I_\diff$ as well. The exact computations are shown below based on the case of four input images, i.e., when including $I_\diff$. The computations are analogously for input $(I_\rgb,I_\pert,I_\pred)$.

Given are the RGB image $I_\rgb$, the perturbed mask $I_\pert$, the prediction mask $I_\pred$ and the difference mask $I_\diff$ with $I_\rgb,I_\pert,I_\pred,I_\diff\in\{0,1,\ldots,255\}^{H\times W\times 3}$. The input tensor is a stack / concatenation of all four images

$$I_{\inp}=[I_\rgb,I_\pert,I_\pred,I_\diff]\in\{0,1,\ldots,255\}^{H\times W\times 12},$$

where $W$ is the width and $H$ the height in pixels of the images. For the normalization of each input tensor $I_\inp$, we compute for each channel in $I_\rgb$ the mean and standard deviation by

$$\mu_c^\rgb=\frac{1}{H\cdot W}\sum_{h=1}^H\sum_{w=1}^W i_{h,w,c}^\rgb$$
$$\sigma_c^\rgb=\sqrt{\frac{1}{H\cdot W}\sum_{h=1}^H\sum_{w=1}^W(i_{h,w,c}^\rgb - \mu_c^\rgb)^2}$$

for all channels $c=1,2,3$. For the perturbed mask the computations are performed analogously. The results for $I_\rgb$ are

$$\mu_\rgb=(\mu_1^\rgb, \mu_2^\rgb, \mu_3^\rgb)\quad\text{and}\quad \sigma_\rgb=(\sigma_1^\rgb,\sigma_2^\rgb,\sigma_3^\rgb)$$

and mean and standard deviation of $I_\pert$ are

$$\mu_\pert=(\mu_1^\pert,\mu_2^\pert,\mu_3^\pert) \quad\text{and}\quad \sigma_\pert=(\sigma_1^\pert,\sigma_2^\pert,\sigma_3^\pert).$$

The RGB image $I_\rgb$ is normalized using its computed mean and standard deviation. The perturbed mask $I_\pert$, predicted mask $I_\pred$ and difference mask $I_\diff$ are normalized by $\mu_\pert$ and $\sigma_\pert$. The result is

$$I_\inp^\textit{norm}=\frac{I_\inp-(\mu_\rgb,\mu_\pert,\mu_\pert,\mu_\pert)}{(\sigma_\rgb,\sigma_\pert,\sigma_\pert,\sigma_\pert)}\in \mathbb{R}^{H\times W\times 12}.$$

\begin{table*}[!t]
\centering
\caption{Performance results on validation data of a trained label error detector $f_\lerr(I_\rgb,I_\pert,I_\pred,I_\diff)$ on classes \emph{drop}, \emph{flip} and \emph{spawn} on a perturbed Cityscapes validation dataset with perturbation rate $q=0.2$ and $t=1$). For evaluation we chose a score threshold of $0.5$ and an IoU threshold of $0.1$.}
\begin{tabular}{cccccccc}
\hline
\textbf{Class} & \textbf{AP} & \textbf{TPs} & \textbf{FPs} & \textbf{FNs} & \textbf{Precision} & \textbf{Recall} & \textbf{F1-Score} \\
\hline
drop & 32.03 & 723 & 1463 & 159 & 33.07 & 81.97 & 47.13 \\
flip & 46.07 & 909 & 1741 & 124 & 34.30 & 88.00 & 49.36\\
spawn & 88.77 & 1153 & 199 & 38 & 85.28 & 96.81 & 90.68\\
drop, flip, spawn & 55.62 & 2785 & 3403 & 321 & 45.01 & 89.67 & 59.93\\
\hline 
\end{tabular}
\label{tab:segm_city_classes}
\end{table*}

\subsection{Details on Difference Mask Generation}

\begin{figure}[h]
    \centering
    \begin{subfigure}{0.48\columnwidth}
        \centering
        \includegraphics[width=\linewidth]{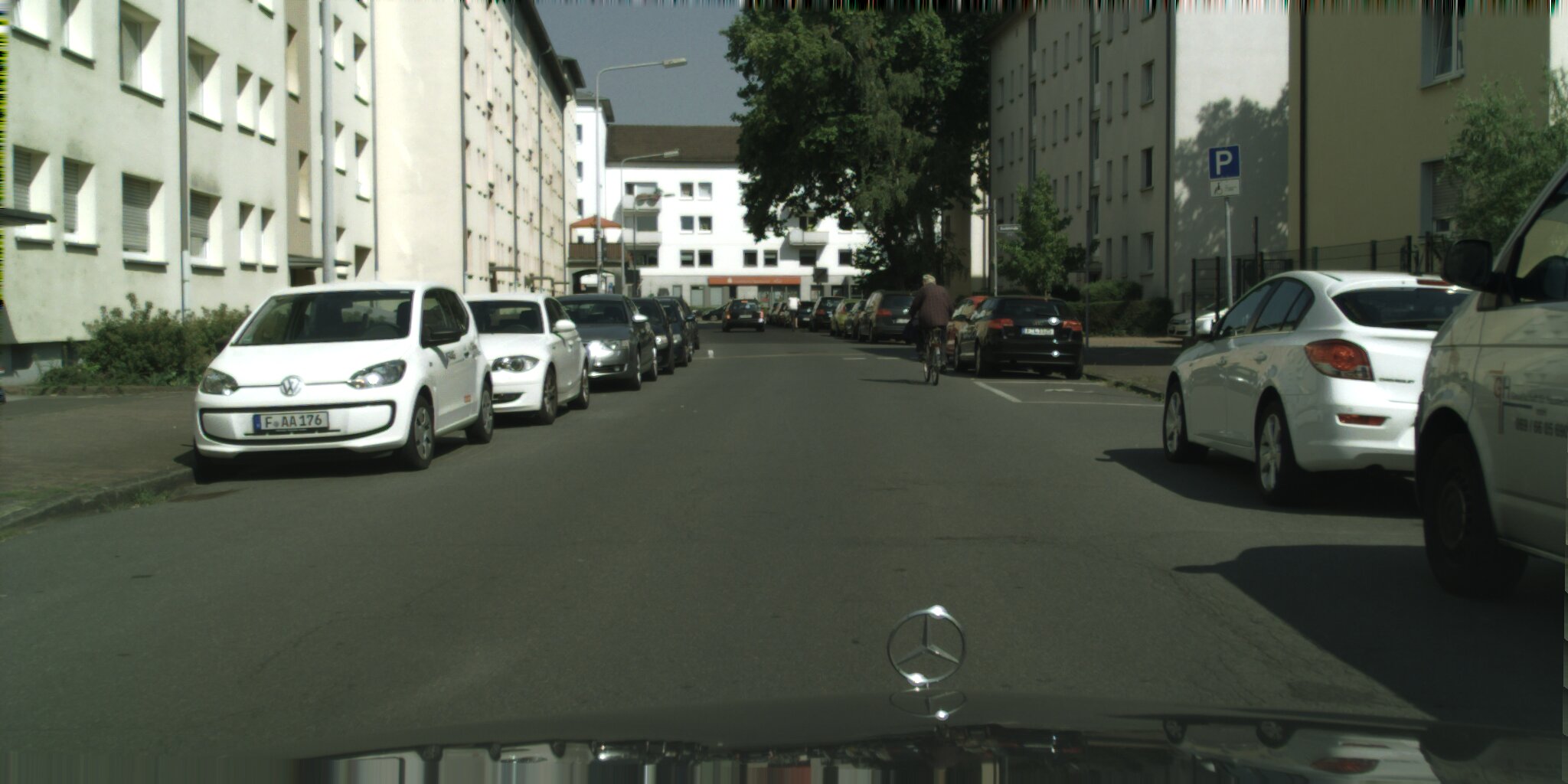}
        \caption*{input image}
    \end{subfigure}
    \hfill
    \begin{subfigure}{0.48\columnwidth}
        \centering
        \includegraphics[width=\linewidth]{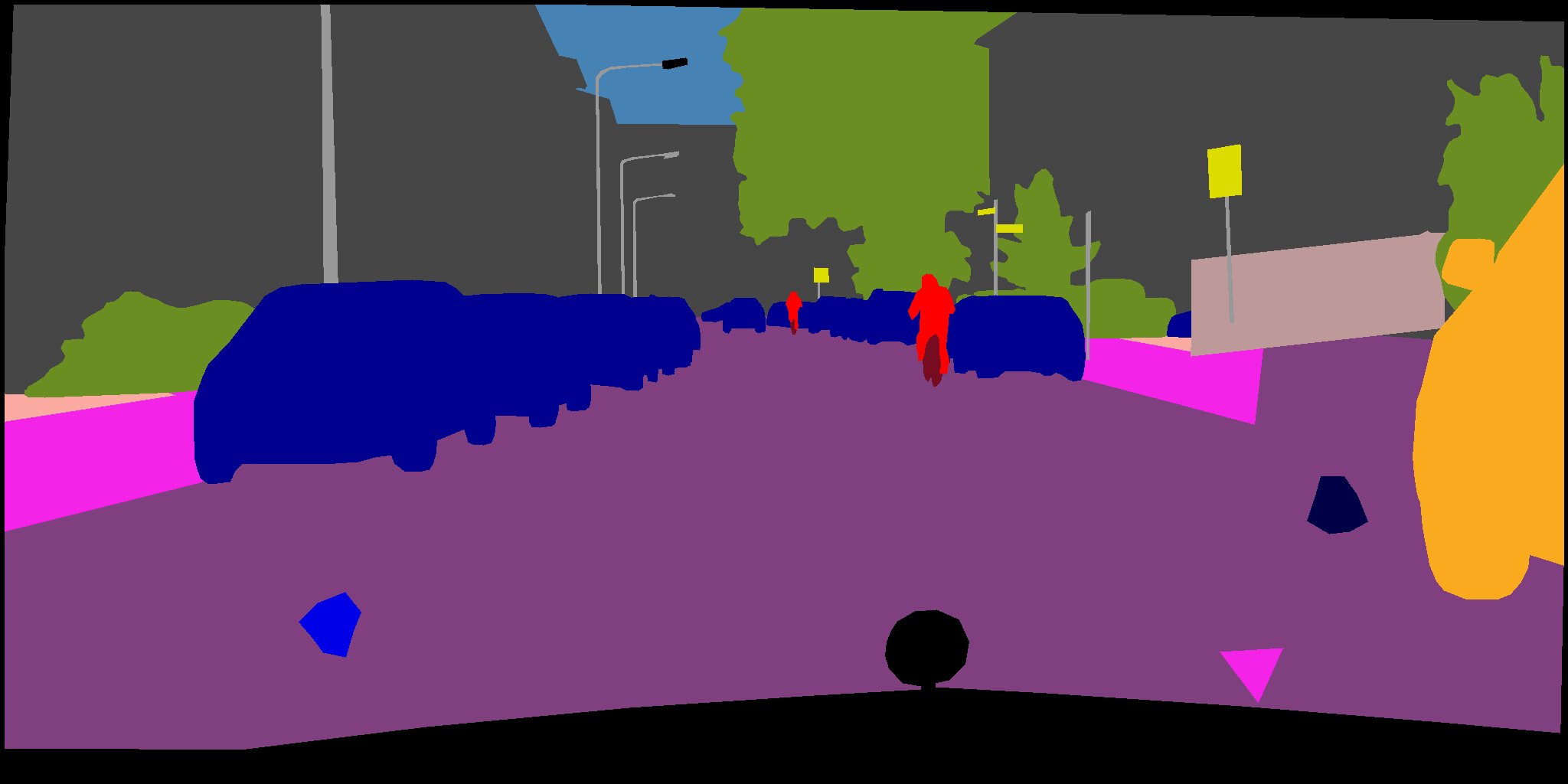}
        \caption*{perturbed mask}
    \end{subfigure}

    \vspace{0.3cm}

    \begin{subfigure}{0.48\columnwidth}
        \centering
        \includegraphics[width=\linewidth]{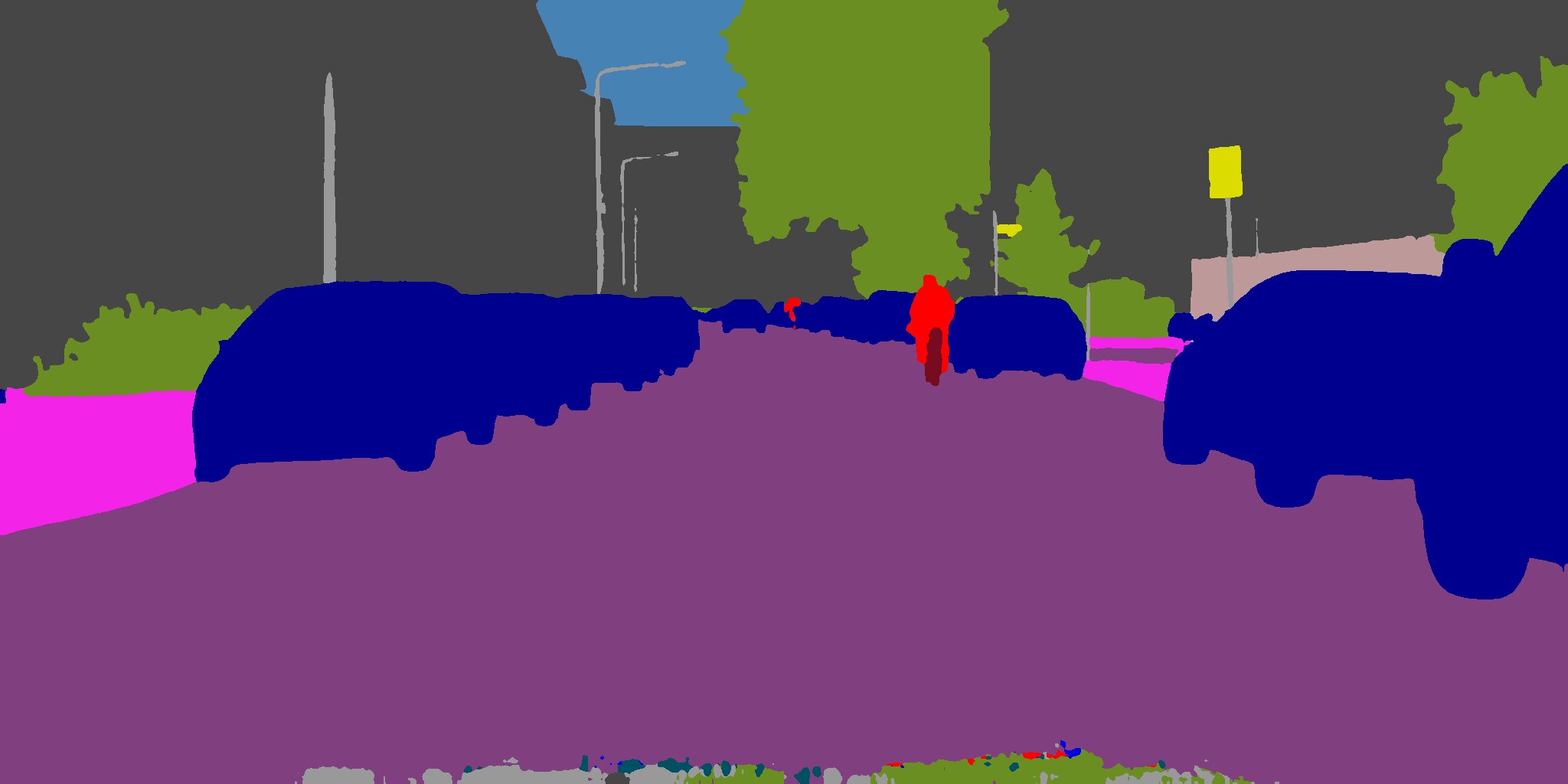}
        \caption*{predicted mask}
    \end{subfigure}
    \hfill
    \begin{subfigure}{0.48\columnwidth}
        \centering
        \includegraphics[width=\linewidth]{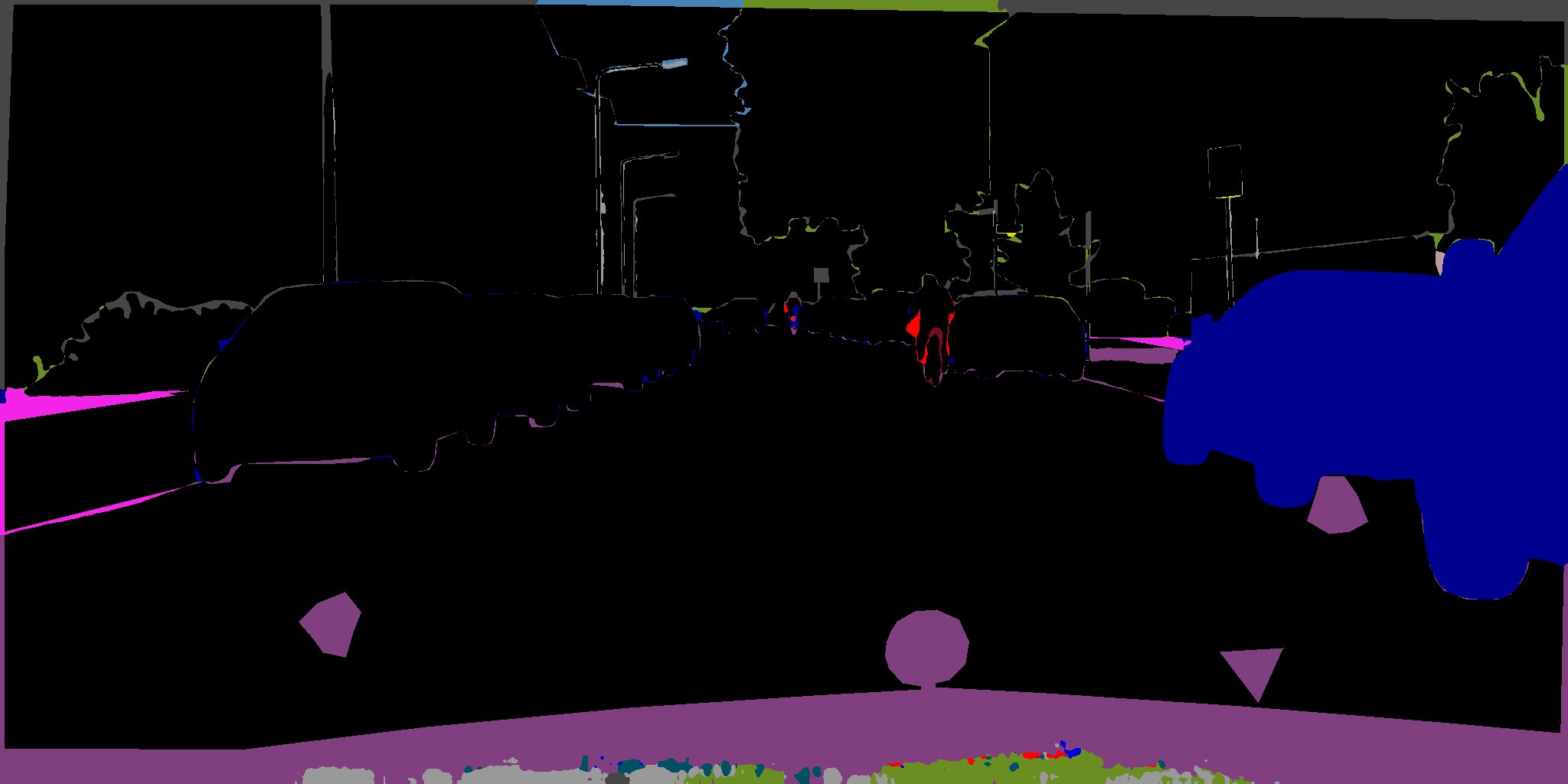}
        \caption*{difference mask}
    \end{subfigure}

    \vspace{0.5cm}

    \begin{subfigure}{0.48\columnwidth}
        \centering
        \includegraphics[width=\linewidth,clip,trim=50 100 0 0]{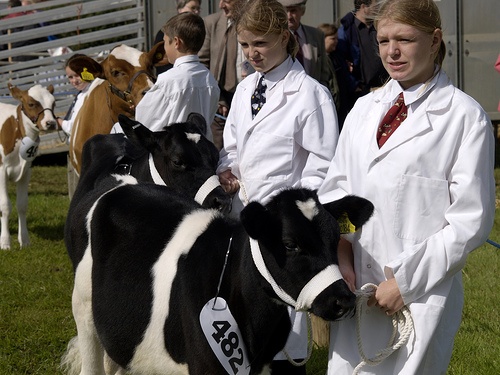}
        \caption*{input image}
    \end{subfigure}
    \hfill
    \begin{subfigure}{0.48\columnwidth}
        \centering
        \includegraphics[width=\linewidth,clip,trim=50 100 0 0]{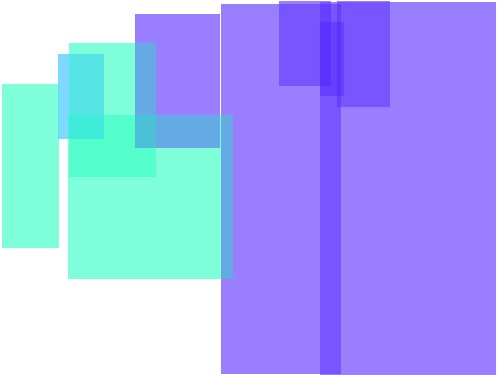}
        \caption*{perturbed mask}
    \end{subfigure}
    \hfill
    \begin{subfigure}{0.48\columnwidth}
        \centering
        \includegraphics[width=\linewidth,clip,trim=50 100 0 0]{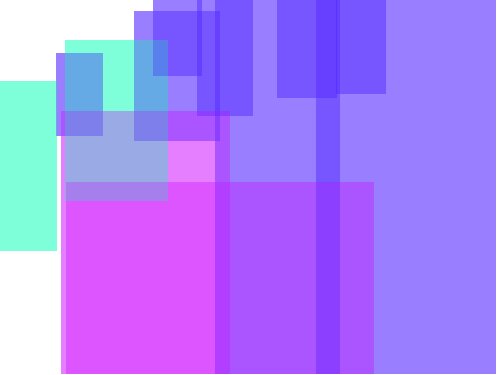}
        \caption*{predicted mask}
    \end{subfigure}
    \hfill
    \begin{subfigure}{0.48\columnwidth}
        \centering
        \includegraphics[width=\linewidth,clip,trim=50 100 0 0]{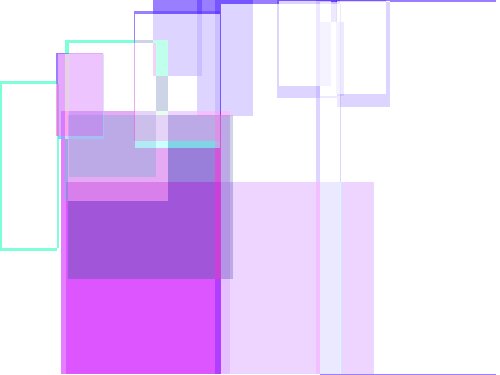}
        \caption*{difference mask}
    \end{subfigure}

    \caption{Overview of input image, perturbed mask, prediction mask and difference mask for semantic segmentation (first four images) and object detection datasets.}
    \label{fig:binary}
\end{figure}

The difference masks illustrate the differences between the perturbed ground truth and the prediction mask in the semantic segmentation and object detection datasets (see \cref{fig:binary}). For Cityscapes and ADE20K, the difference mask is created by comparing the perturbed and predicted masks pixel by pixel. Each pixel that differs is given the same color as the corresponding pixel in the predicted mask. For object detection datasets, only the differences between the HSV-colored perturbed and predicted masks are computed. For instance segmentation datasets, especially in the case of LIVECell, each object is assigned a random color, making it difficult to find corresponding instances between the perturbed and predicted masks. Since training the label error detector on perturbed LIVECell data was successful without a difference mask, we decided to proceed without one for instance segmentation.

\subsection{Ablation Studies}

In addition to the ablation studies in the main part of the manuscript, we performed a study training on three separate classes: \emph{drop}, \emph{flip} and \emph{spawn}. To train the label error detector, we used the Cascade Mask-RCNN as the instance segmentor and a perturbed dataset with a perturbation rate of $q=0.2$ and a number of copies of $t=10$ as training data. Instead of using one class \emph{error} for the instance segmentor's training feedback (as done in all previous experiments), we used the  classes \emph{drop}, \emph{flip} and \emph{spawn} for training the instance segmentor. Training was performed with a learning rate of $0.001$, using the difference mask $I_\diff$ as additional input alongside the input image $I_\rgb$, the perturbed mask $I_\pert$ and the predicted mask $I_\pred$ from the reference network Segformer.
The results are shown in \cref{tab:segm_city_classes}. We evaluated the trained label error detector on a validation dataset with a perturbation rate of $q=0.2$ and a number of copies of $t=1$. In the evaluation, we take all three classes into consideration. It can be seen that the label error objects known as \emph{spawns} are the easiest to learn and also easiest to find giving rise to the highest AP, precision, recall and F1-score out of all classes. By comparing the results of \emph{drop} and \emph{flip}, we observe that the latter performs even better in terms of 14\% higher AP and 8\% higher recall. As seen in the main part of this paper, training on class \emph{error} achieves a higher AP of 61.74\% and a higher recall of 95.78\%.

\subsection{A Collection of Real Label Errors}
Below, we present a collection of real label errors that we have found using our label error detector. Each collage contains 18 pairs of images, each of which represents one or more real label errors. These five figures show label errors from PascalVOC (\cref{fig:real_voc}), COCO (\cref{fig:real_coco}), ADE20K (\cref{fig:real_ade}), Cityscapes (\cref{fig:real_city}) and LIVECell (\cref{fig:real_cell}). 

\begin{figure}[h]
  \centering
  \begin{minipage}{\linewidth}
    \centering
    \makebox[0.3\linewidth]{%
  \shortstack[c]{Ground Truth}%
    }
    \makebox[0.3\linewidth]{%
  \shortstack[c]{Label Error\\ Proposal}%
    }\makebox[0.3\linewidth]{%
    \shortstack[c]{Reference\\ Prediction}%
    }\par
    \includegraphics[width=0.3\linewidth,clip,trim=50 30 80 30]{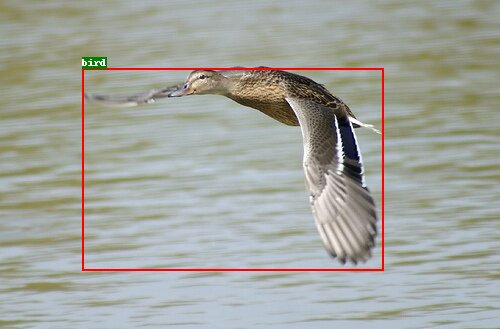}
    \includegraphics[width=0.3\linewidth,clip,trim=50 30 80 30]{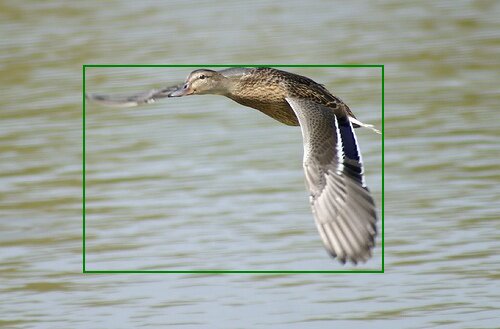}
    \includegraphics[width=0.3\linewidth,clip,trim=50 30 80 30]{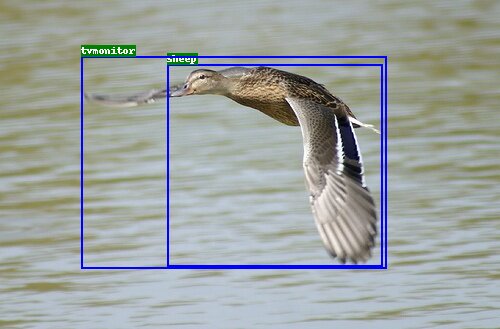}
    \includegraphics[width=0.3\linewidth,clip,trim=50 0 100 140]{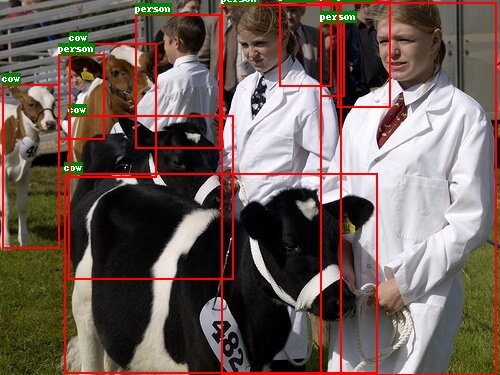}
    \includegraphics[width=0.3\linewidth,clip,trim=50 0 100 140]{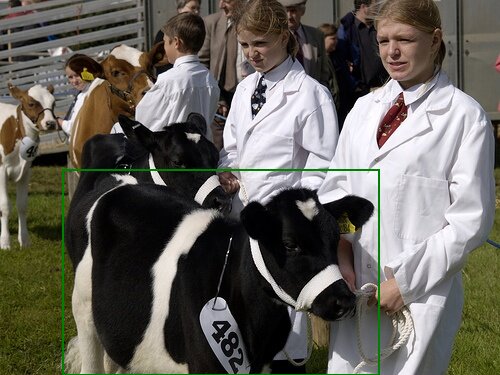}
    \includegraphics[width=0.3\linewidth,clip,trim=50 0 100 140]{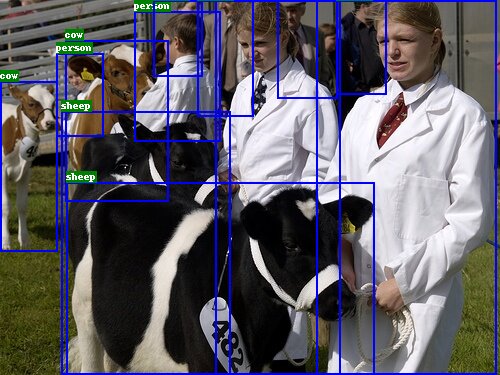}
  \end{minipage}
  \begin{minipage}{\linewidth}
    \centering
    \includegraphics[width=0.3\linewidth,clip,trim=240 80 0 80]{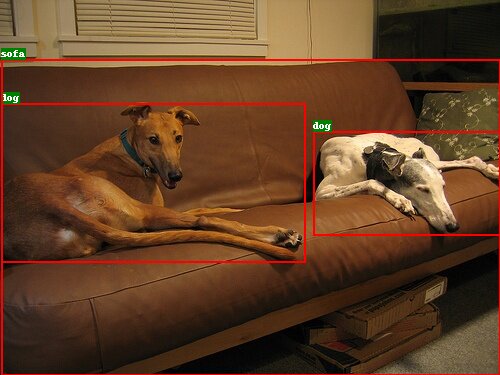}
    \includegraphics[width=0.3\linewidth,clip,trim=240 80 0 80]{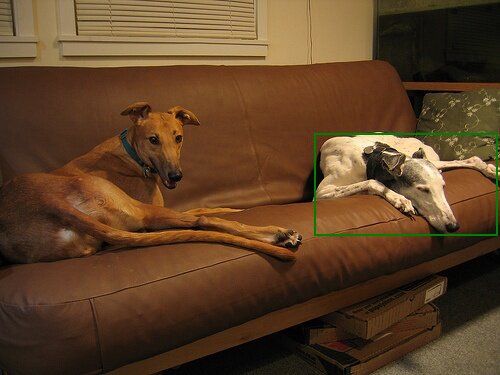}
    \includegraphics[width=0.3\linewidth,clip,trim=240 80 0 80]{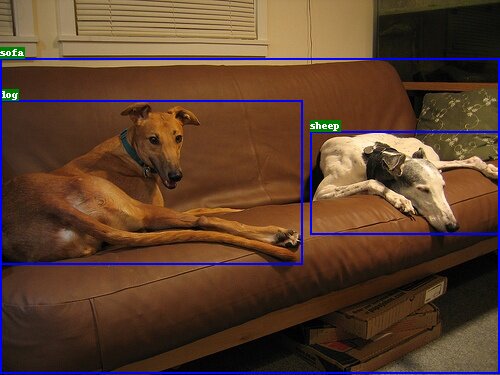}
    \includegraphics[width=0.3\linewidth,clip,trim=0 40 100 170]{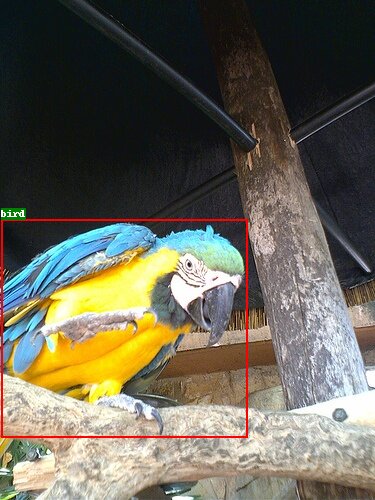}
    \includegraphics[width=0.3\linewidth,clip,trim=0 40 100 170]{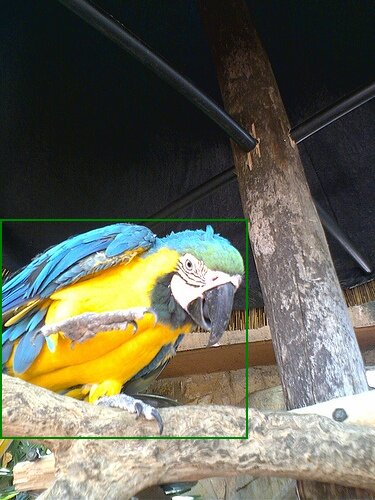}
    \includegraphics[width=0.3\linewidth,clip,trim=0 40 100 170]{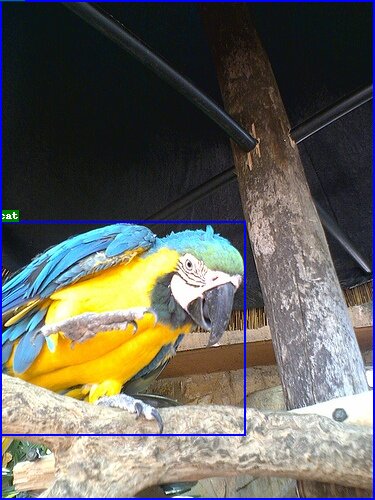}
  \end{minipage}

  \caption{Examples of false positive label error proposals of our label error detector on PascalVOC dataset. In red are the ground truth bounding boxes, in blue are the predicted bounding boxes from reference network GroundingDino and in green are the label error proposals of our label error detector for real label errors.}
  \label{fig:limitation}
\end{figure}

\subsection{Limitations}

We performed a study of the false positives of our label error detection method in \cref{fig:limitation}. In particular in the case of object detection annotations, we noticed that our method (as well as the baselines considered in our work) is influenced by false predictions of the reference model. This represent a gap that will be closed by stronger object detectors in future. 
Furthermore, although we demonstrated the generalization of our method to the detection of real label errors, it is to be expected that the label errors that we simulate via our label perturbation algorithm do not cover the entire distribution of potential label errors. 

\newpage
\onecolumn

\begin{figure}
\centering
  \newcommand{\imgsep}{1cm} 
\includegraphics[width=0.15\linewidth,clip,trim=80 10 250 100]{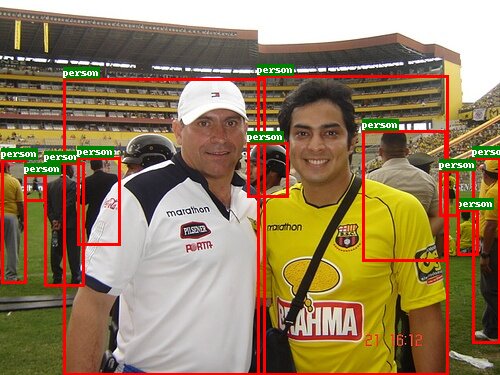}%
\includegraphics[width=0.15\linewidth,clip,trim=80 10 250 100]{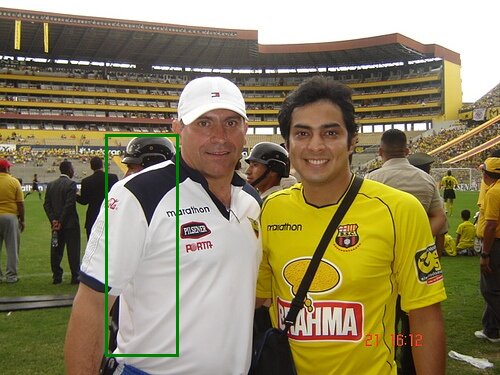}%
\includegraphics[width=0.15\linewidth,clip,trim=0 20 350 80]{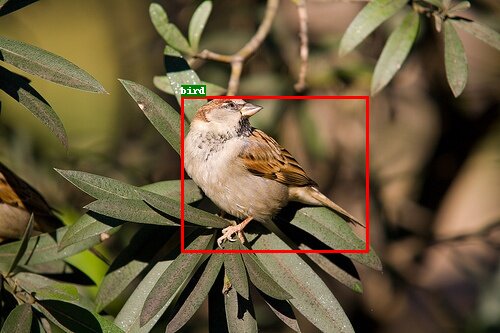}%
\includegraphics[width=0.15\linewidth,clip,trim=0 20 350 80]{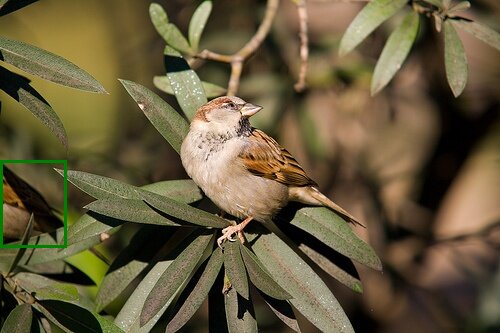}%
\includegraphics[width=0.15\linewidth,clip,trim=0 80 350 20]{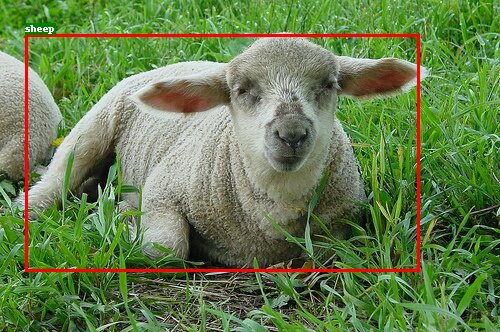}%
\includegraphics[width=0.15\linewidth,clip,trim=0 80 350 20]{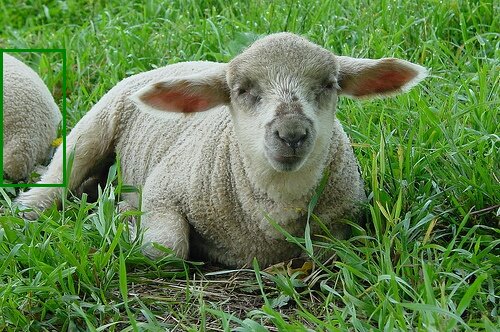}
\hspace{\imgsep}%
\includegraphics[width=0.15\linewidth,clip,trim=0 90 0 90]{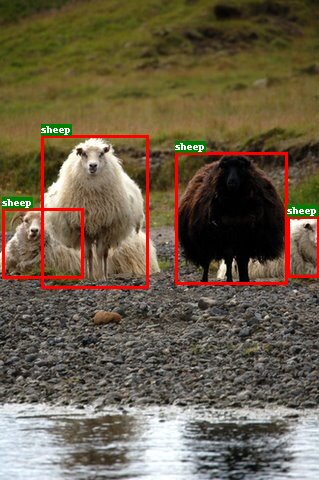}%
\includegraphics[width=0.15\linewidth,clip,trim=0 90 0 90]{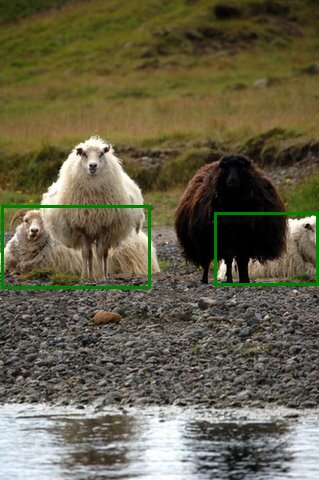}%
\includegraphics[width=0.15\linewidth,clip,trim=100 80 150 20]{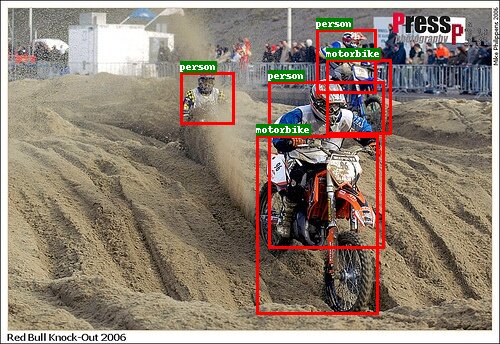}%
\includegraphics[width=0.15\linewidth,clip,trim=100 80 150 20]{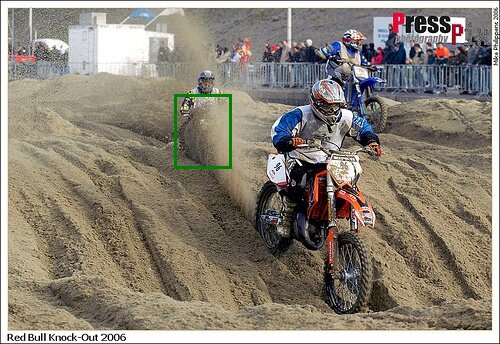}%
\includegraphics[width=0.15\linewidth,clip,trim=0 30 350 160]{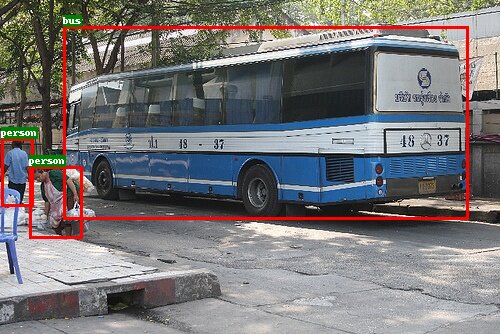}%
\includegraphics[width=0.15\linewidth,clip,trim=0 30 350 160]{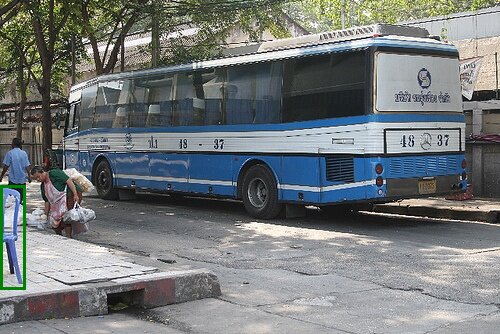}
\hspace{\imgsep}%
\includegraphics[width=0.15\linewidth,clip,trim=100 0 100 100]{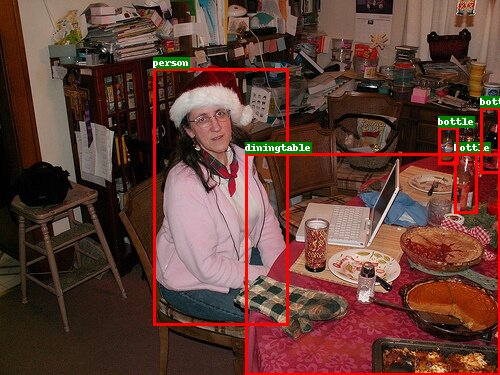}%
\includegraphics[width=0.15\linewidth,clip,trim=100 0 100 100]{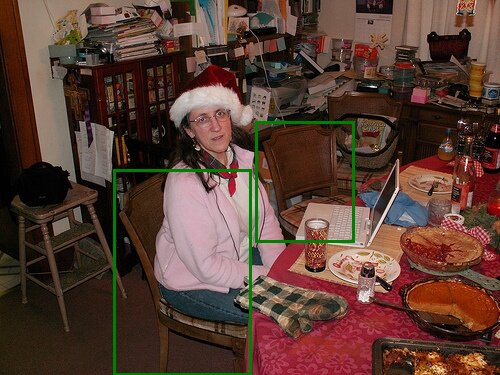}%
\includegraphics[width=0.15\linewidth,clip,trim=350 0 0 240]{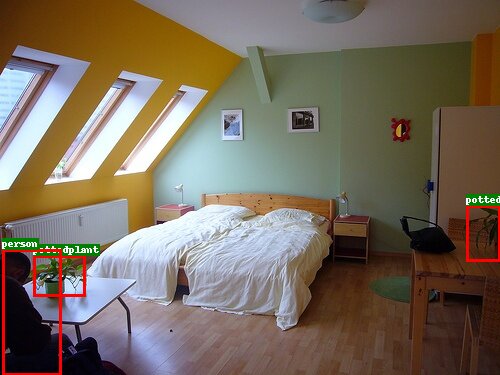}%
\includegraphics[width=0.15\linewidth,clip,trim=350 0 0 240]{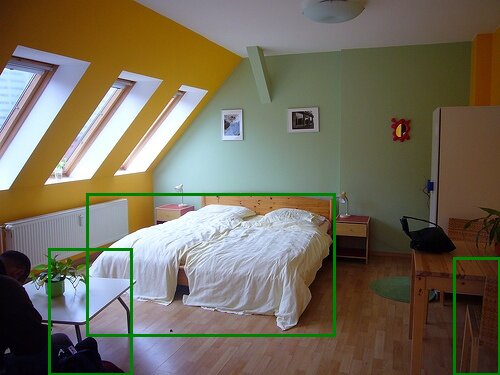}%
\includegraphics[width=0.15\linewidth,clip,trim=50 20 150 100]{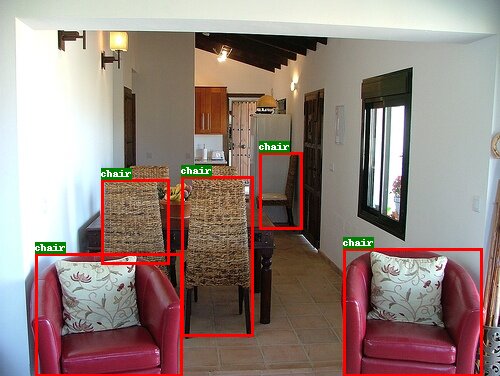}%
\includegraphics[width=0.15\linewidth,clip,trim=50 20 150 100]{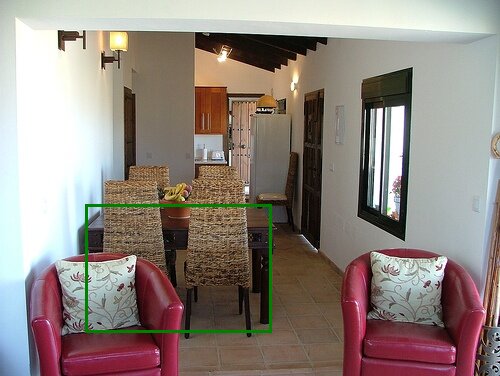}
\hspace{\imgsep}%
\includegraphics[width=0.15\linewidth,clip,trim=0 100 300 50]{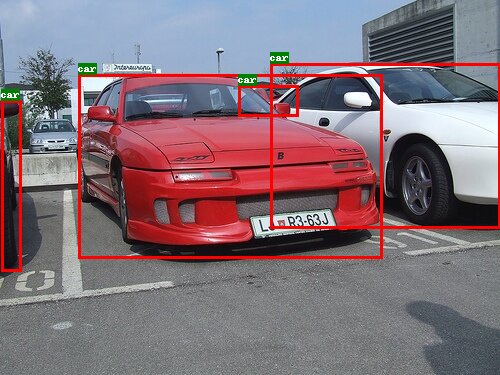}%
\includegraphics[width=0.15\linewidth,clip,trim=0 100 300 50]{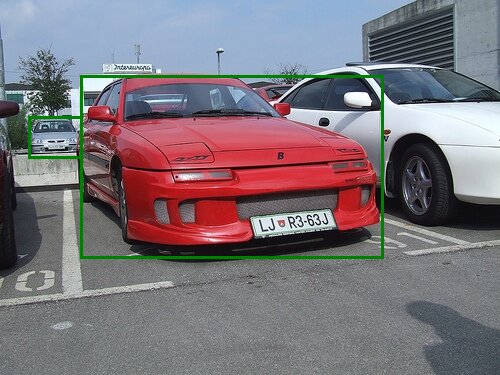}%
\includegraphics[width=0.15\linewidth,clip,trim=0 100 300 10]{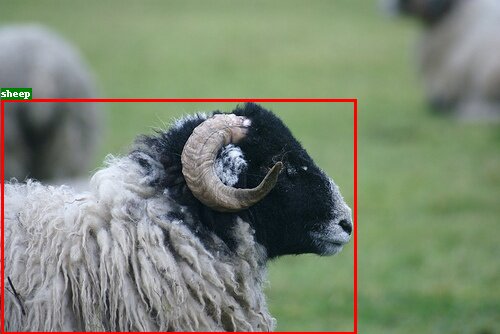}%
\includegraphics[width=0.15\linewidth,clip,trim=0 100 300 10]{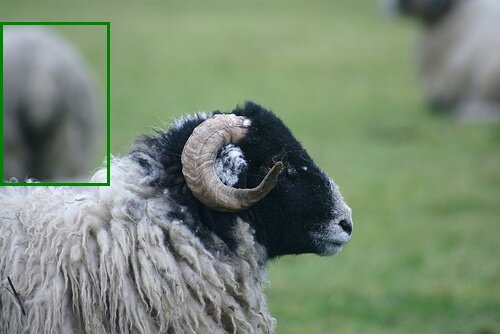}%
\includegraphics[width=0.15\linewidth,clip,trim=100 240 0 0]{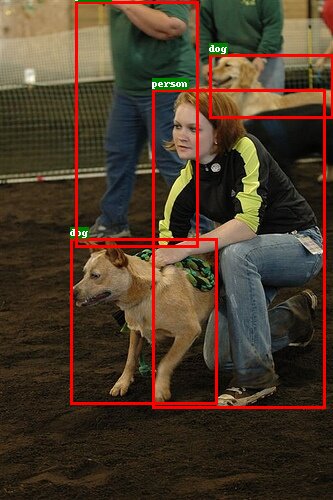}%
\includegraphics[width=0.15\linewidth,clip,trim=100 240 0 0]{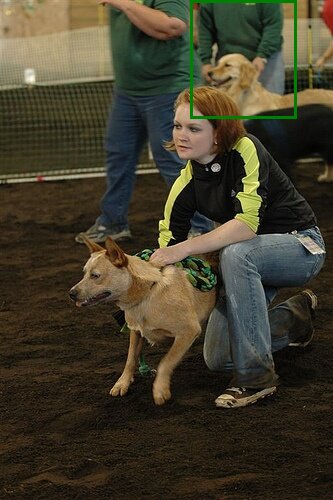}
\hspace{\imgsep}%
\includegraphics[width=0.15\linewidth,clip,trim=300 80 0 80]{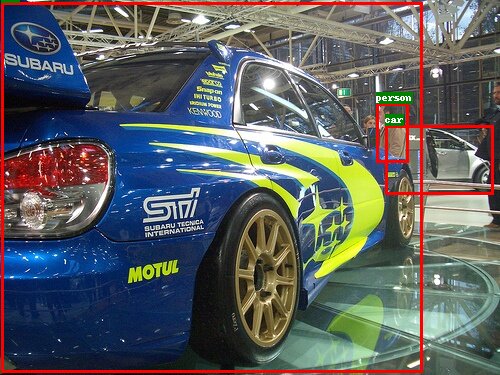}%
\includegraphics[width=0.15\linewidth,clip,trim=300 80 0 80]{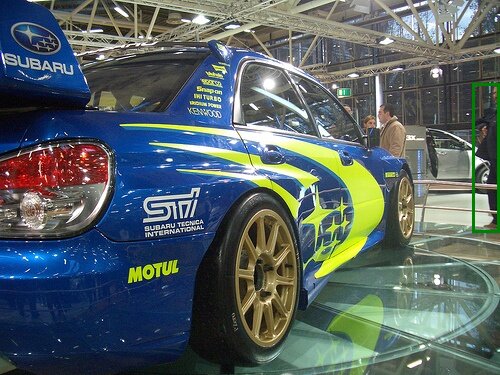}%
\includegraphics[width=0.15\linewidth,clip,trim=0 0 200 10]{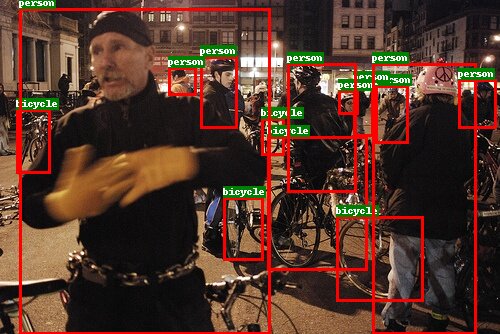}%
\includegraphics[width=0.15\linewidth,clip,trim=0 0 200 10]{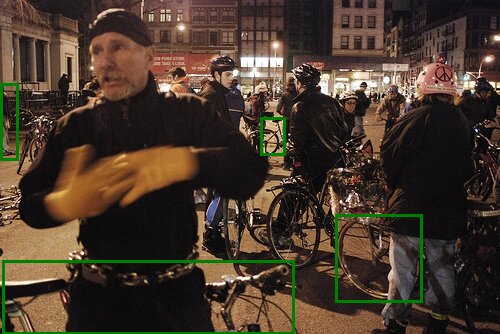}%
\includegraphics[width=0.15\linewidth,clip,trim=0 0 150 0]{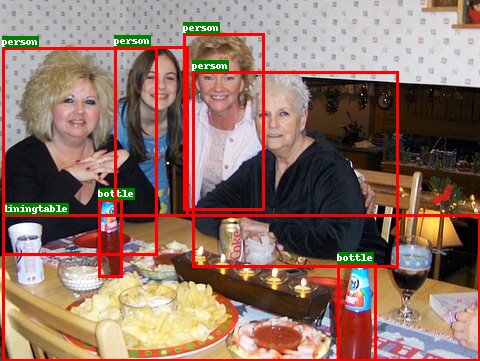}%
\includegraphics[width=0.15\linewidth,clip,trim=0 0 150 0]{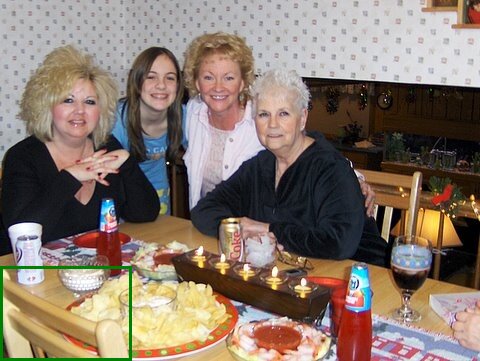}
\hspace{\imgsep}%
\includegraphics[width=0.15\linewidth,clip,trim=0 100 200 0]{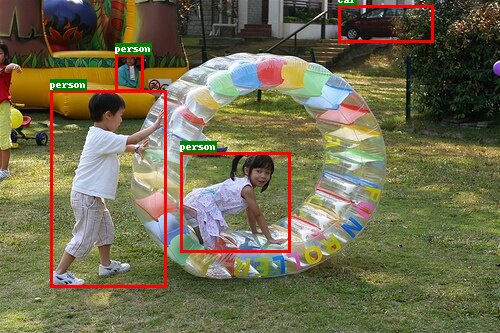}%
\includegraphics[width=0.15\linewidth,clip,trim=0 100 200 0]{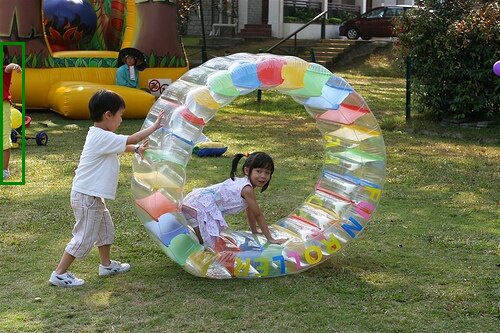}%
\includegraphics[width=0.15\linewidth,clip,trim=70 0 0 0]{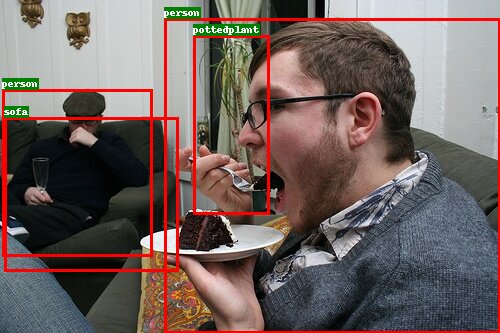}%
\includegraphics[width=0.15\linewidth,clip,trim=70 0 0 0]{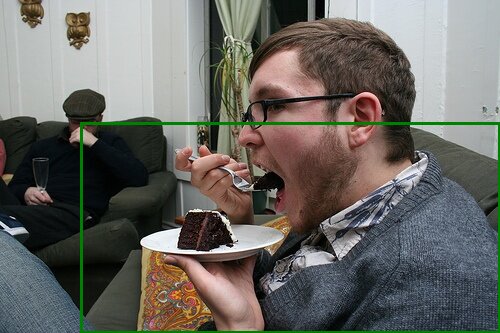}%
\includegraphics[width=0.15\linewidth,clip,trim=0 100 200 0]{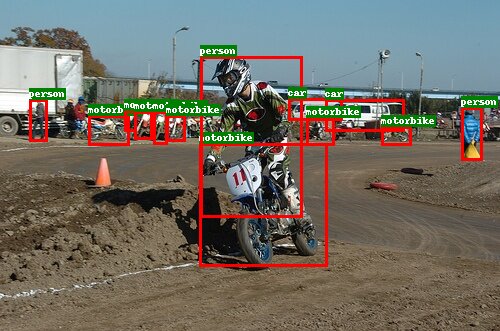}%
\includegraphics[width=0.15\linewidth,clip,trim=0 100 200 0]{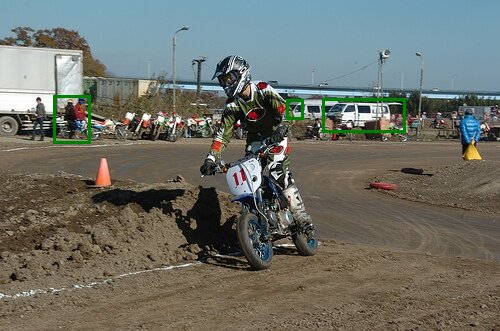}

\caption{A collection of label errors present in PascalVOC. In red are the ground truth boxes and in green the predictions of our label error detection method.}
\label{fig:real_voc}
\end{figure}

\begin{figure}
\centering
  \newcommand{\imgsep}{1cm} 
\includegraphics[width=0.15\linewidth,clip,trim=100 140 0 40]{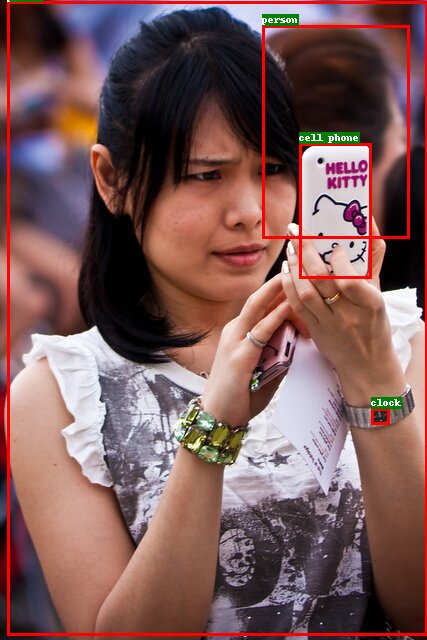}%
\includegraphics[width=0.15\linewidth,clip,trim=100 140 0 40]{images/real_errs_coco/000000001296_rgb.jpg}%
\includegraphics[width=0.15\linewidth,clip,trim=300 0 0 0]{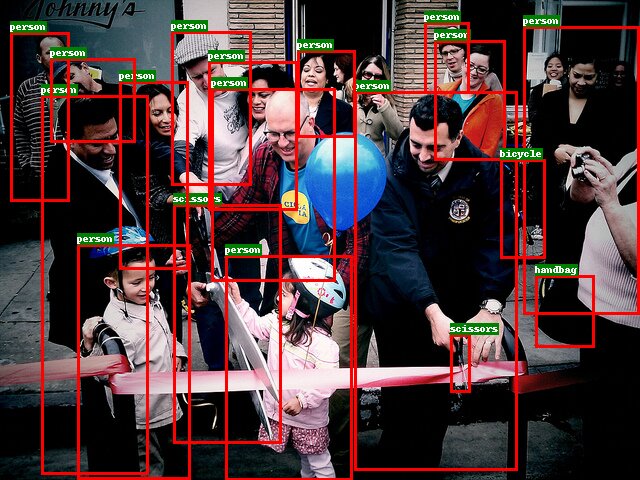}%
\includegraphics[width=0.15\linewidth,clip,trim=300 0 0 0]{images/real_errs_coco/000000005001_rgb.jpg}%
\includegraphics[width=0.15\linewidth,clip,trim=0 0 150 170]{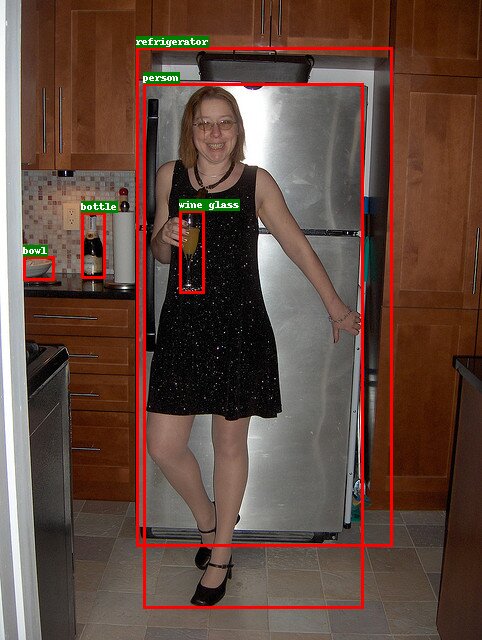}%
\includegraphics[width=0.15\linewidth,clip,trim=0 0 150 170]{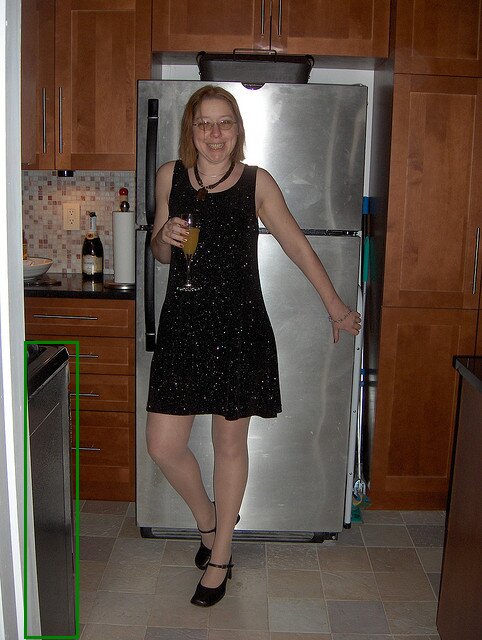}%
\hspace{\imgsep}%
\includegraphics[width=0.15\linewidth,clip,trim=150 0 100 0]{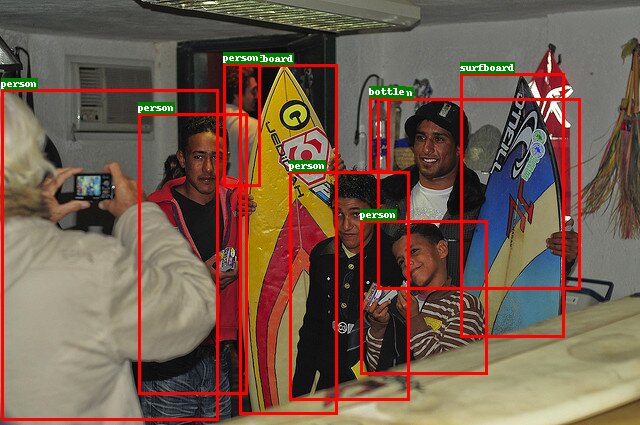}%
\includegraphics[width=0.15\linewidth,clip,trim=150 0 100 0]{images/real_errs_coco/000000005193_rgb.jpg}
\includegraphics[width=0.15\linewidth,clip,trim=0 100 300 10]{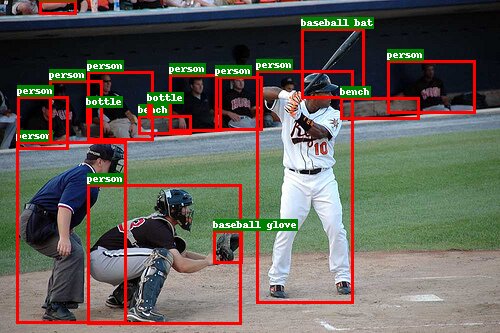}%
\includegraphics[width=0.15\linewidth,clip,trim=0 100 300 10]{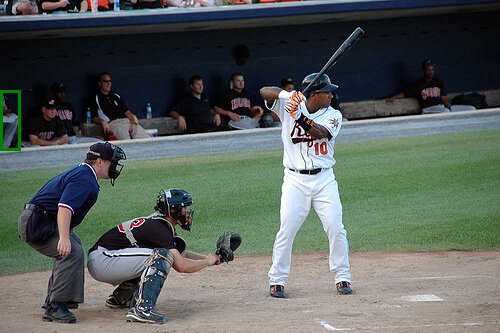}%
\includegraphics[width=0.15\linewidth,clip,trim=50 0 0 0]{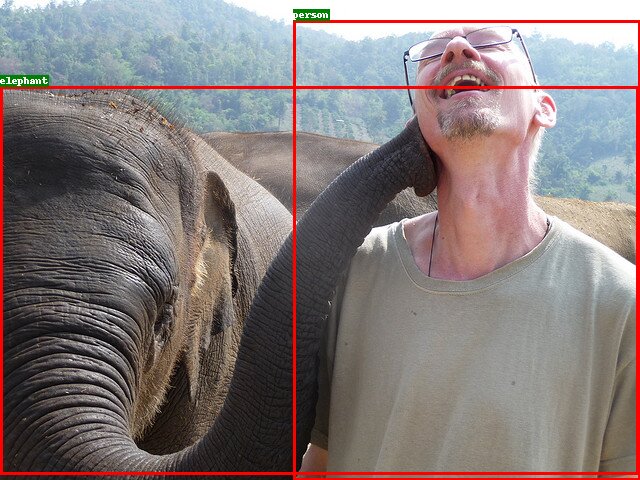}%
\includegraphics[width=0.15\linewidth,clip,trim=50 0 0 0]{images/real_errs_coco/000000006894_rgb.jpg}
\hspace{\imgsep}%
\includegraphics[width=0.15\linewidth,clip,trim=300 0 0 0]{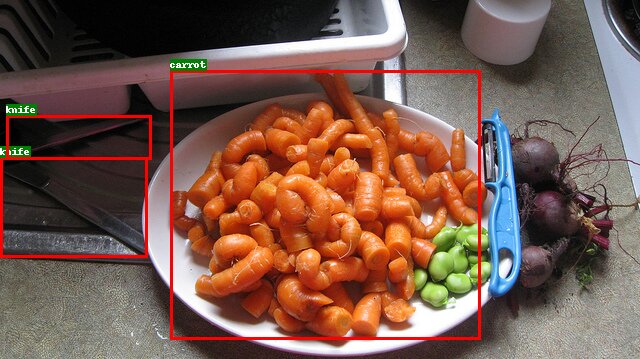}%
\includegraphics[width=0.15\linewidth,clip,trim=300 0 0 0]{images/real_errs_coco/000000007991_rgb.jpg}%
\includegraphics[width=0.15\linewidth,clip,trim=50 40 50 260]{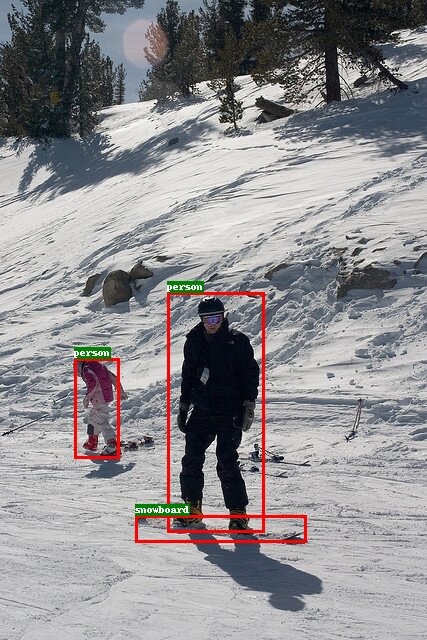}%
\includegraphics[width=0.15\linewidth,clip,trim=50 40 50 260]{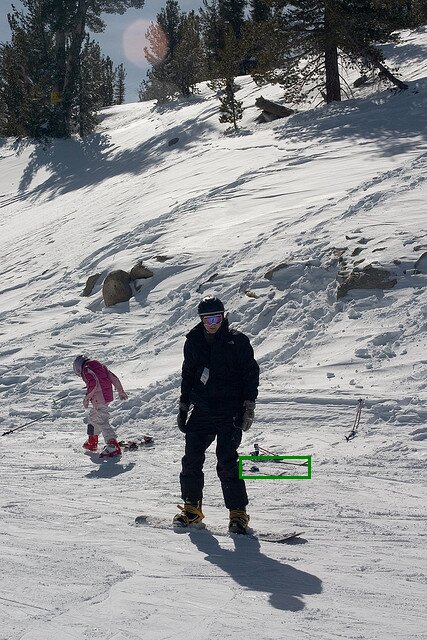}%
\includegraphics[width=0.15\linewidth,clip,trim=300 30 0 40]{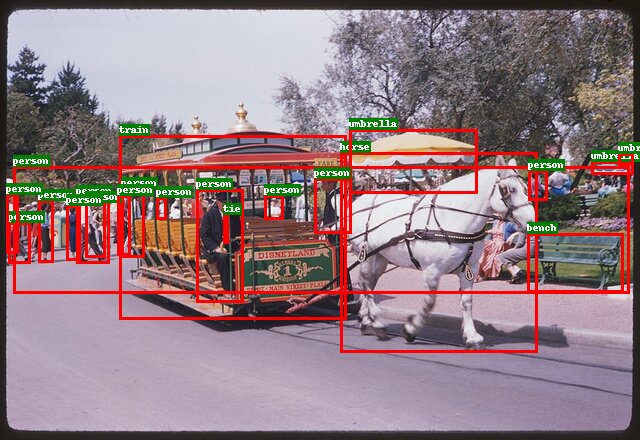}%
\includegraphics[width=0.15\linewidth,clip,trim=300 30 0 40]{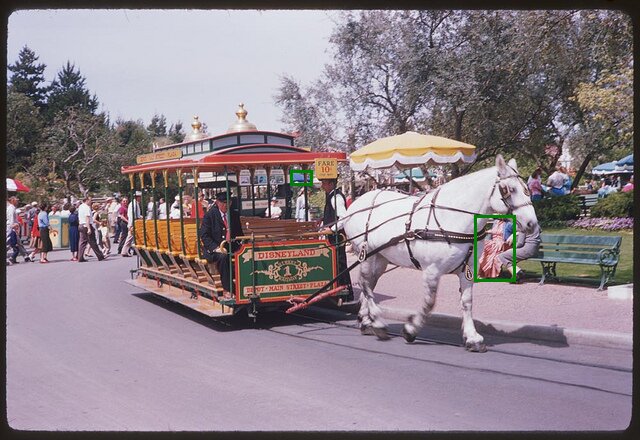}%
\hspace{\imgsep}%
\includegraphics[width=0.15\linewidth,clip,trim=300 0 0 80]{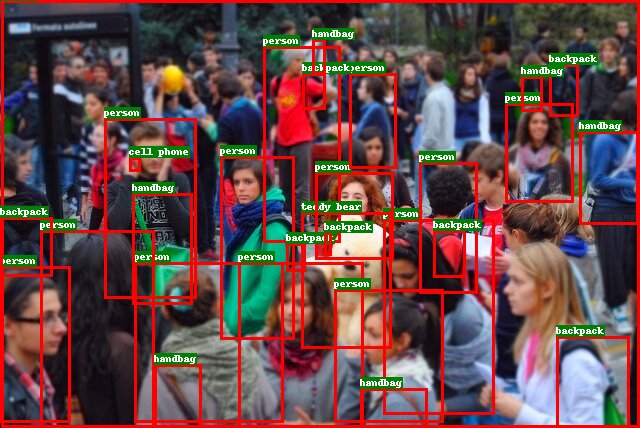}%
\includegraphics[width=0.15\linewidth,clip,trim=300 0 0 80]{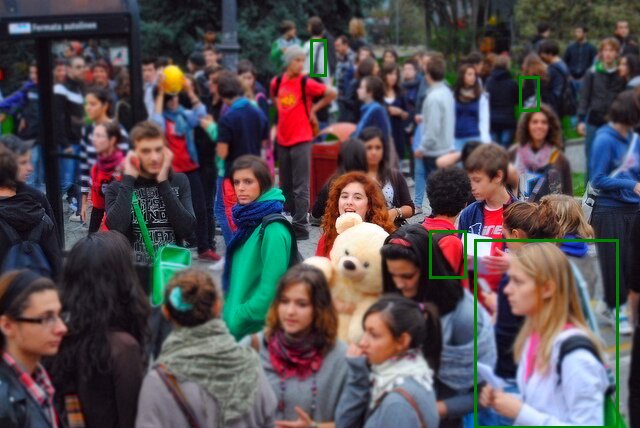}%
\includegraphics[width=0.15\linewidth,clip,trim=0 0 0 160]{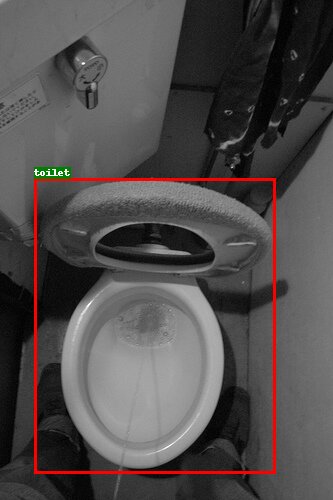}%
\includegraphics[width=0.15\linewidth,clip,trim=0 0 0 160]{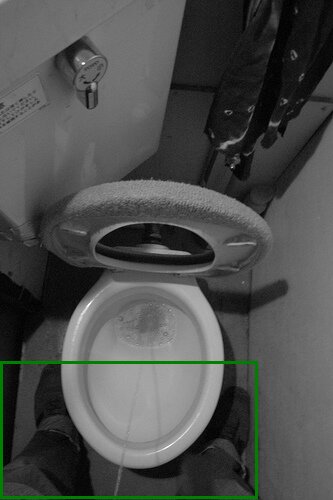}%
\includegraphics[width=0.15\linewidth,clip,trim=0 30 200 0]{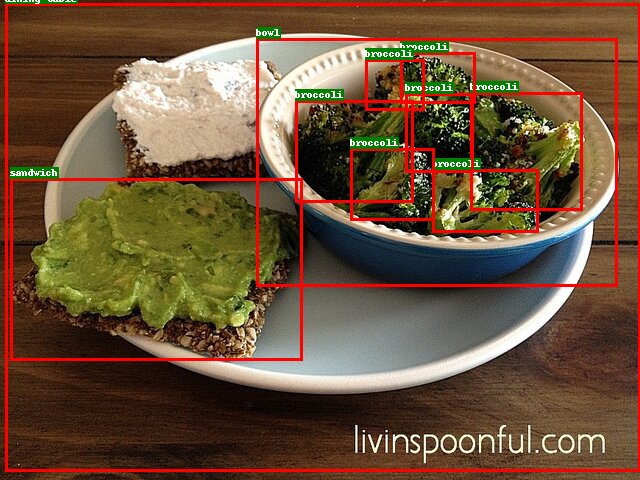}%
\includegraphics[width=0.15\linewidth,clip,trim=0 30 200 0]{images/real_errs_coco/000000030785_rgb.jpg}
\hspace{\imgsep}%
\includegraphics[width=0.15\linewidth,clip,trim=100 0 100 10]{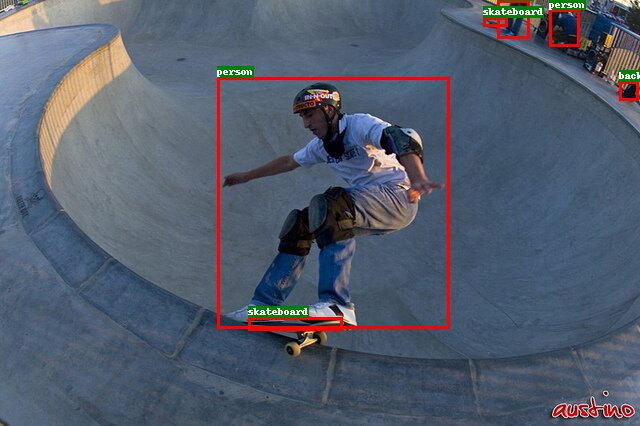}%
\includegraphics[width=0.15\linewidth,clip,trim=100 0 100 10]{images/real_errs_coco/000000031093_rgb.jpg}%
\includegraphics[width=0.15\linewidth,clip,trim=200 50 100 50]{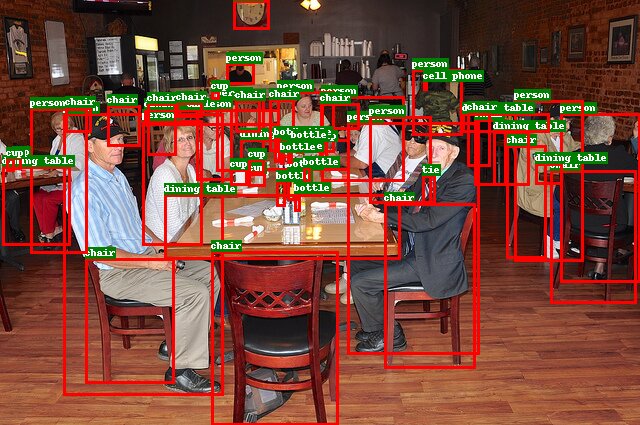}%
\includegraphics[width=0.15\linewidth,clip,trim=200 50 100 50]{images/real_errs_coco/000000031296_rgb.jpg}%
\includegraphics[width=0.15\linewidth,clip,trim=100 0 0 270]{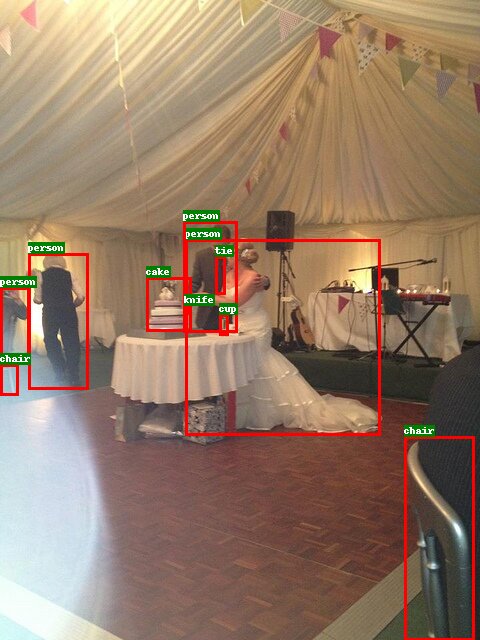}%
\includegraphics[width=0.15\linewidth,clip,trim=100 0 0 270]{images/real_errs_coco/000000031620_rgb.jpg}
\hspace{\imgsep}%
\includegraphics[width=0.15\linewidth,clip,trim=0 280 100 0]{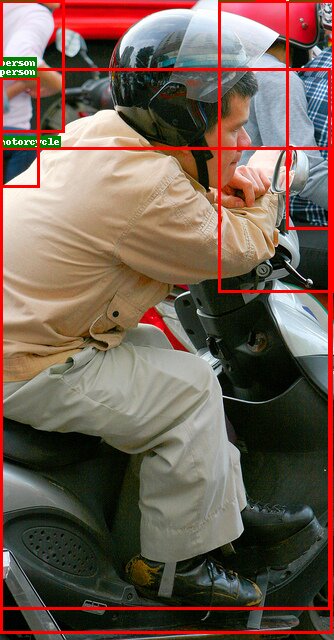}%
\includegraphics[width=0.15\linewidth,clip,trim=0 280 100 0]{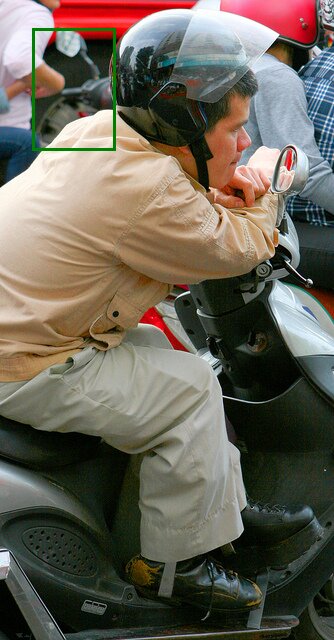}%
\includegraphics[width=0.15\linewidth,clip,trim=0 0 200 150]{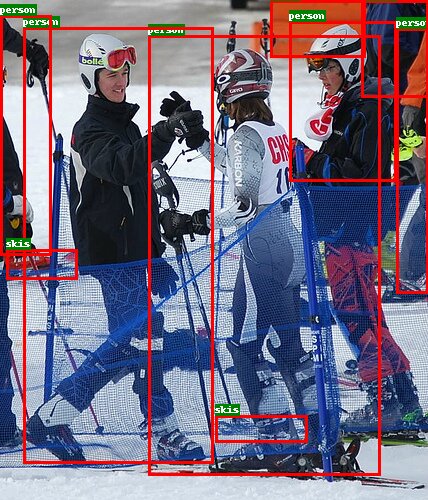}%
\includegraphics[width=0.15\linewidth,clip,trim=0 0 200 150]{images/real_errs_coco/000000033104_rgb.jpg}%
\includegraphics[width=0.15\linewidth,clip,trim=330 0 0 0]{images/real_errs_coco/000000037670_gt.jpg}%
\includegraphics[width=0.15\linewidth,clip,trim=330 0 0 0]{images/real_errs_coco/000000037670_rgb.jpg}%

\caption{A collection of label errors present in COCO. In red are the ground truth boxes and in green the predictions of our label error detection method.}
\label{fig:real_coco}
\end{figure}

\begin{figure}
\centering
\newcommand{\imgsep}{1cm} 
\includegraphics[width=0.165\linewidth,clip,trim=50 0 0 70]{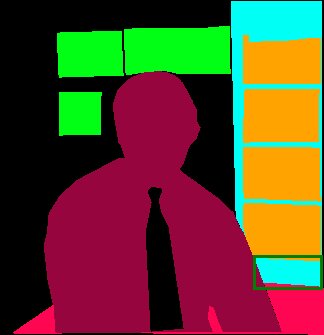}%
\includegraphics[width=0.165\linewidth,clip,trim=50 0 0 70]{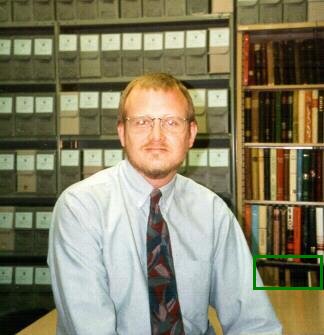}%
\includegraphics[width=0.165\linewidth,clip,trim=200 0 0 40]{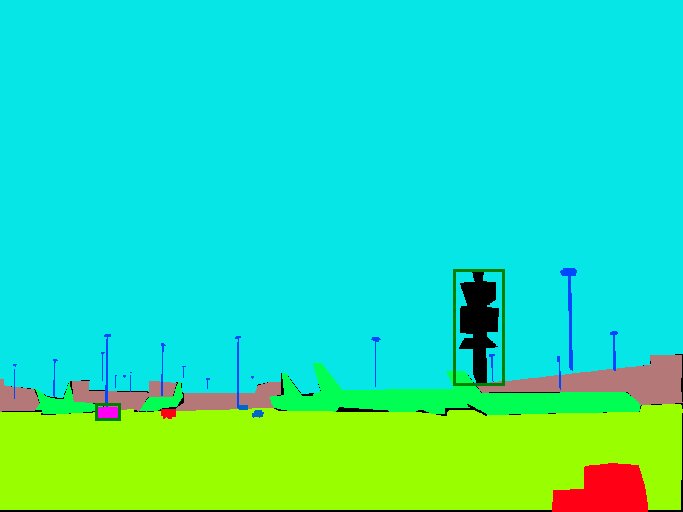}%
\includegraphics[width=0.165\linewidth,clip,trim=200 0 0 40]{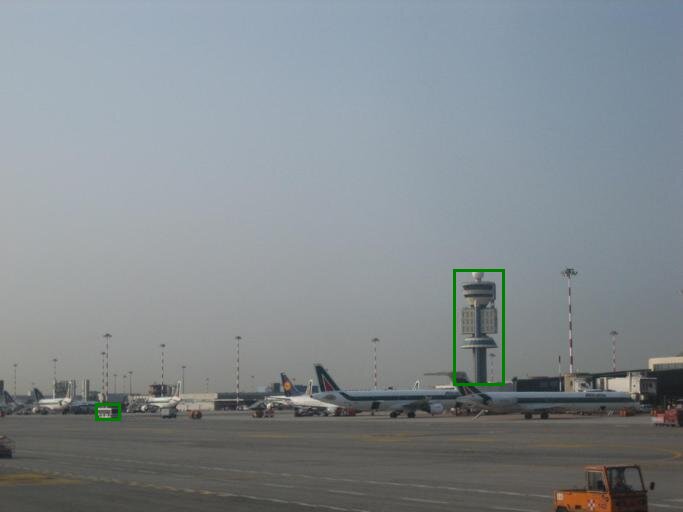}%
\includegraphics[width=0.165\linewidth,clip,trim=435 40 0 0]{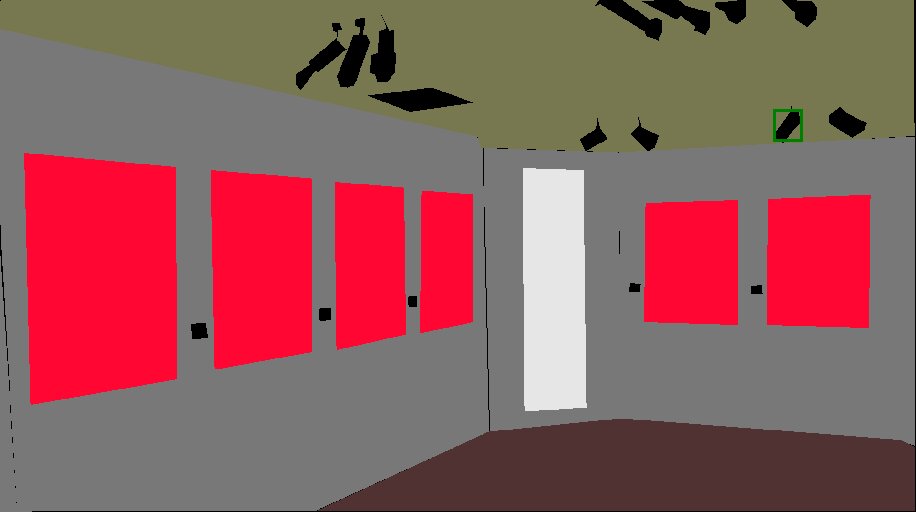}%
\includegraphics[width=0.165\linewidth,clip,trim=435 40 0 0]{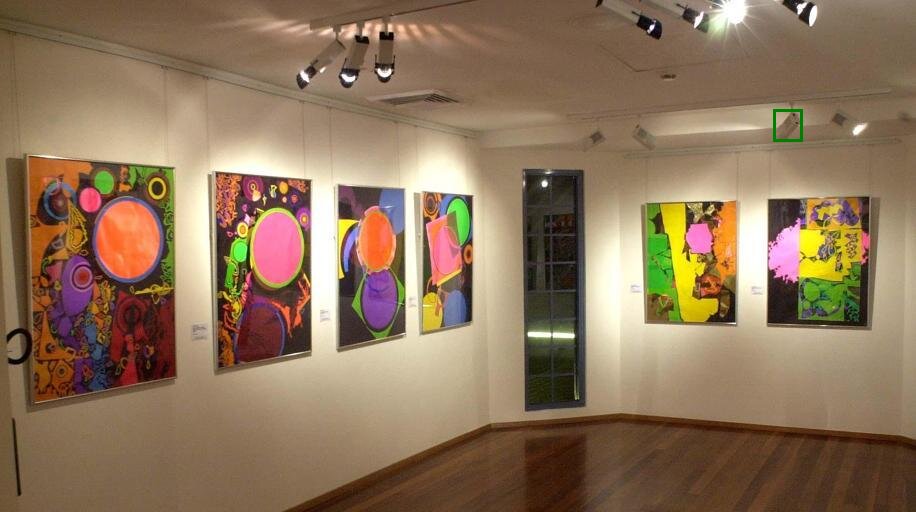}
\hspace{\imgsep}%
\includegraphics[width=0.165\linewidth,clip,trim=50 0 50 250]{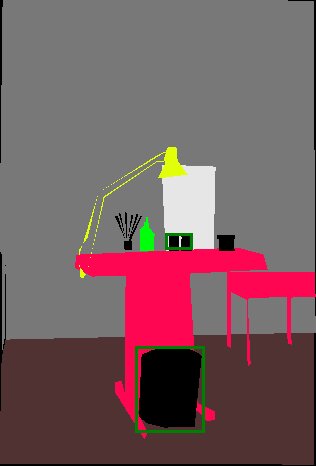}%
\includegraphics[width=0.165\linewidth,clip,trim=50 0 50 250]{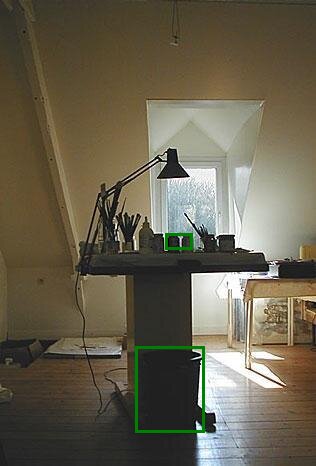}%
\includegraphics[width=0.165\linewidth,clip,trim=360 130 30 0]{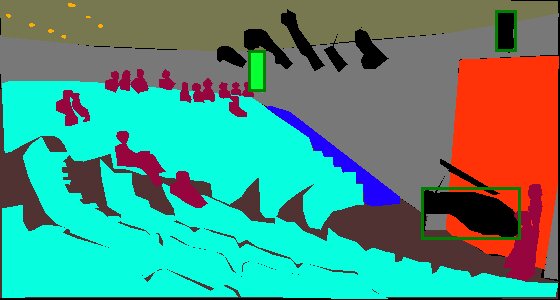}%
\includegraphics[width=0.165\linewidth,clip,trim=360 130 30 0]{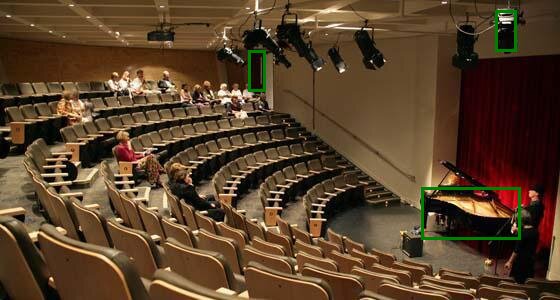}%
\includegraphics[width=0.165\linewidth,clip,trim=270 0 0 145]{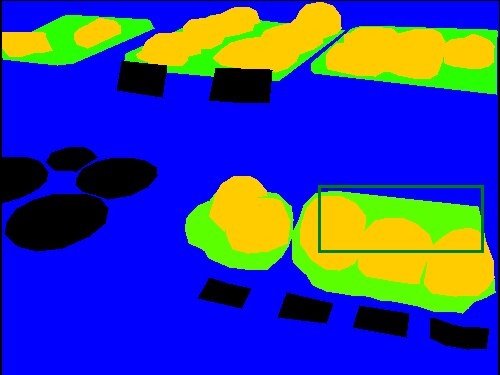}%
\includegraphics[width=0.165\linewidth,clip,trim=270 0 0 145]{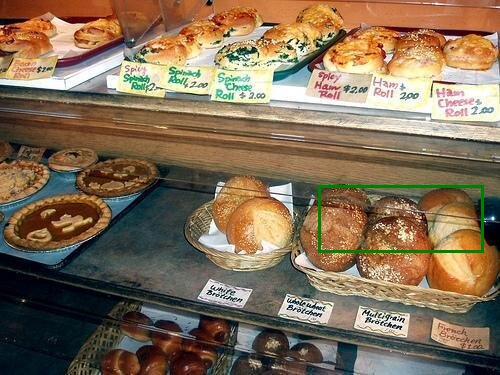}
\hspace{\imgsep}%
\includegraphics[width=0.165\linewidth,clip,trim=250 150 0 70]{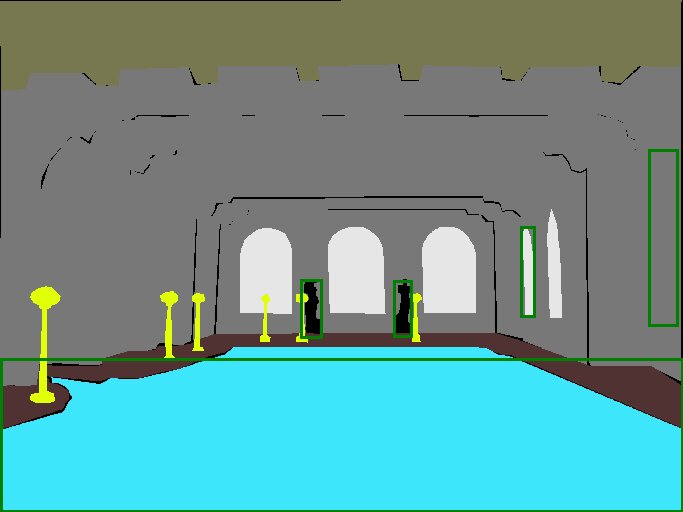}%
\includegraphics[width=0.165\linewidth,clip,trim=250 150 0 70]{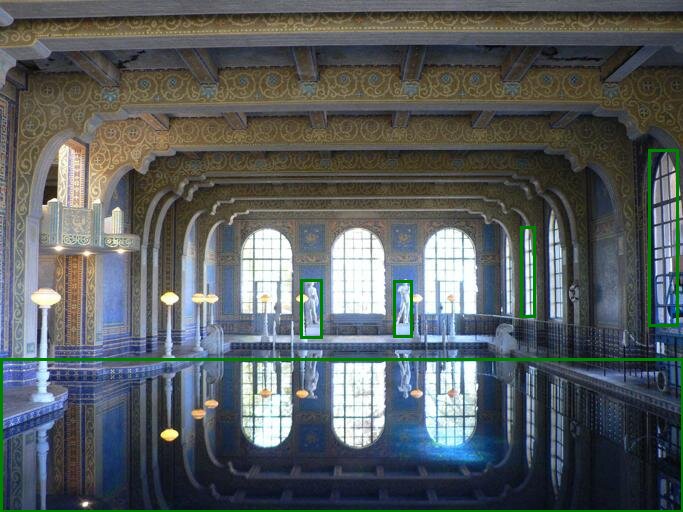}%
\includegraphics[width=0.165\linewidth,clip,trim=0 0 50 90]{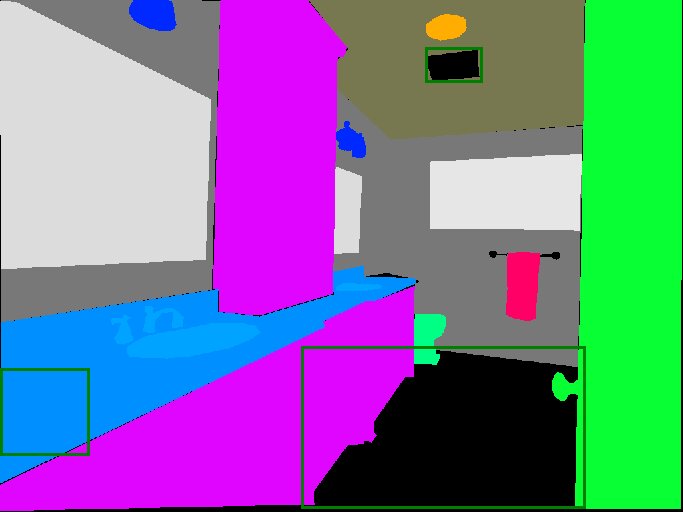}%
\includegraphics[width=0.165\linewidth,clip,trim=0 0 50 90]{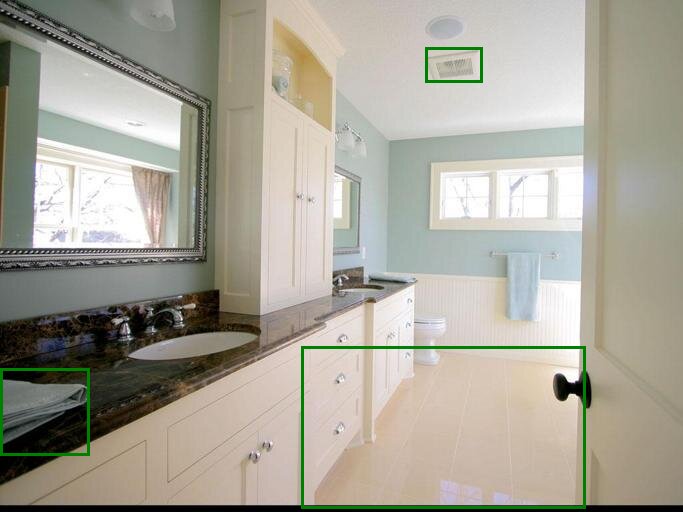}%
\includegraphics[width=0.165\linewidth,clip,trim=0 0 0 350]{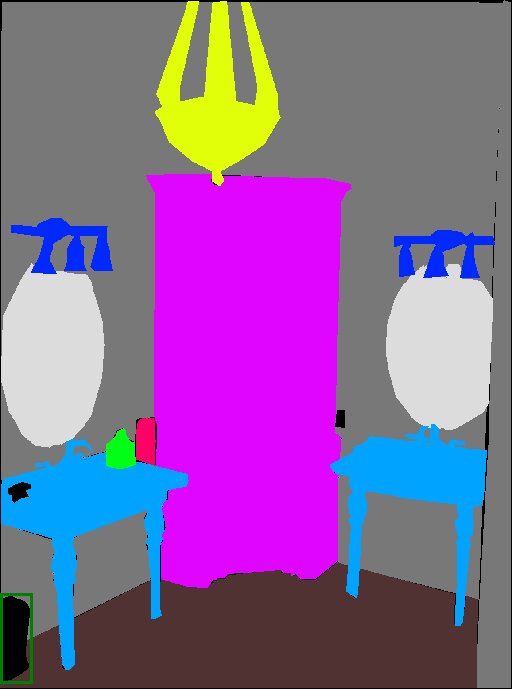}%
\includegraphics[width=0.165\linewidth,clip,trim=0 0 0 350]{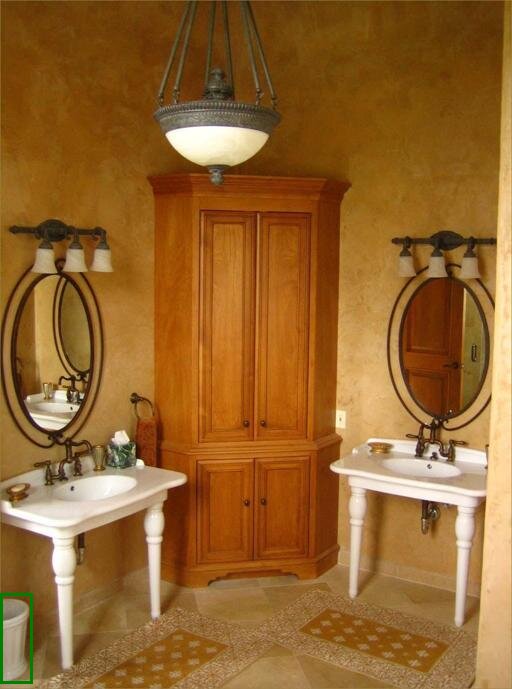}
\hspace{\imgsep}%
\includegraphics[width=0.165\linewidth,clip,trim=0 0 0 250]{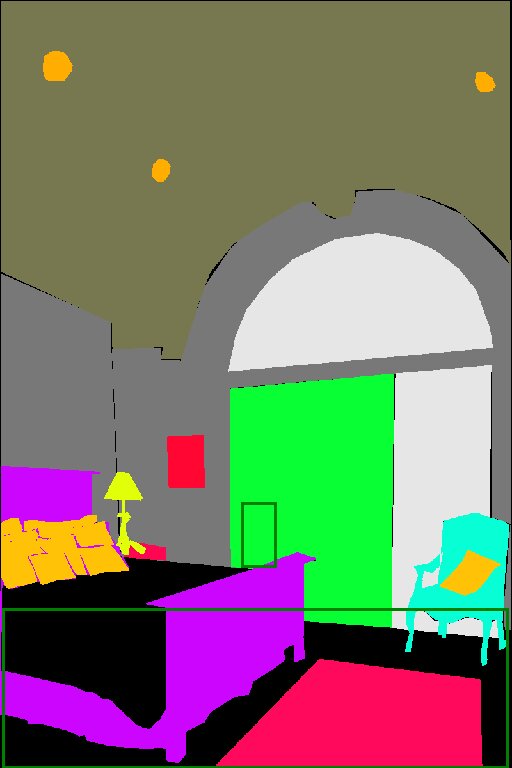}%
\includegraphics[width=0.165\linewidth,clip,trim=0 0 0 250]{images/real_errs_ade/00000138_rgb.jpg}%
\includegraphics[width=0.165\linewidth,clip,trim=360 0 0 100]{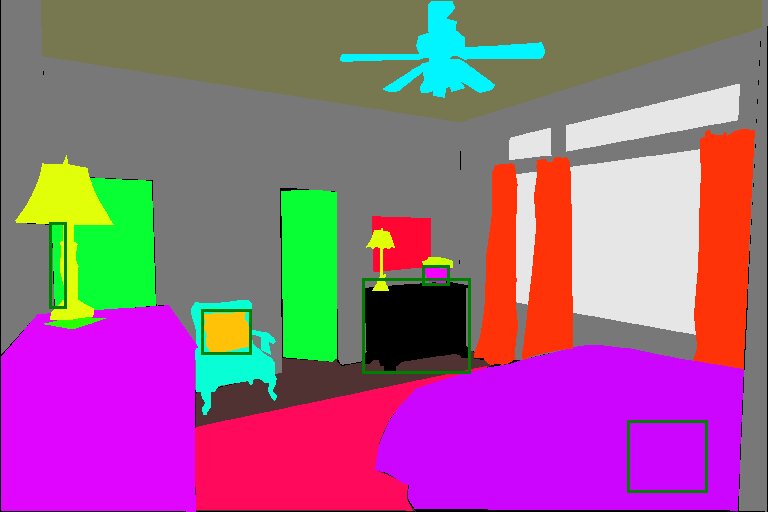}%
\includegraphics[width=0.165\linewidth,clip,trim=360 0 0 100]{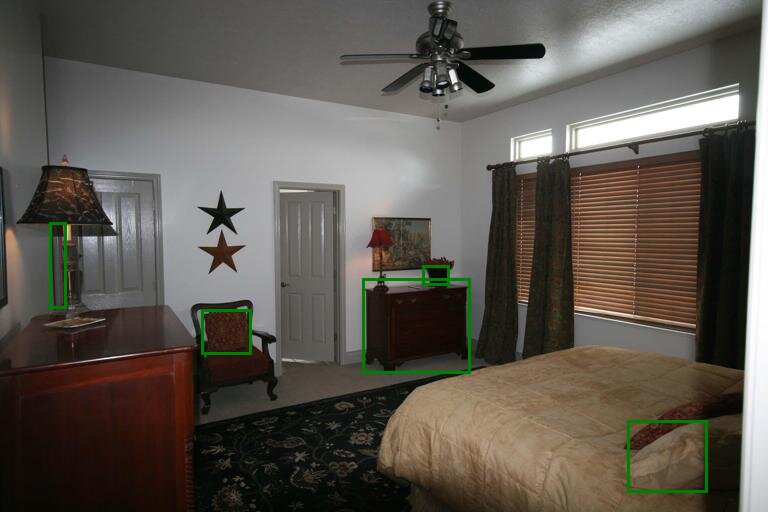}%
\includegraphics[width=0.165\linewidth,clip,trim=115 100 150 0]{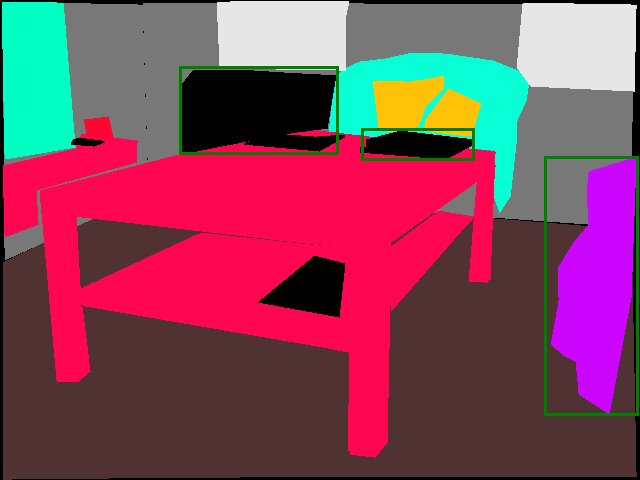}%
\includegraphics[width=0.165\linewidth,clip,trim=115 100 150 0]{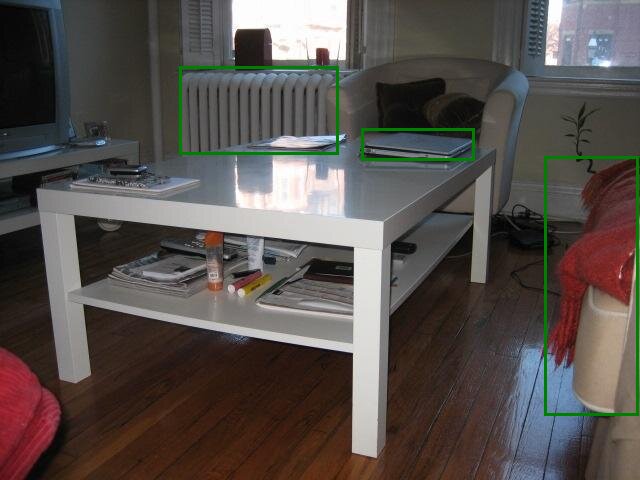}
\hspace{\imgsep}%
\includegraphics[width=0.165\linewidth,clip,trim=0 0 0 0]{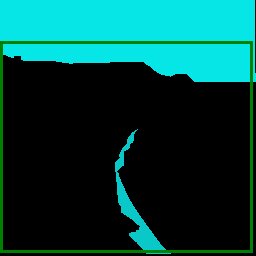}%
\includegraphics[width=0.165\linewidth,clip,trim=0 0 0 0]{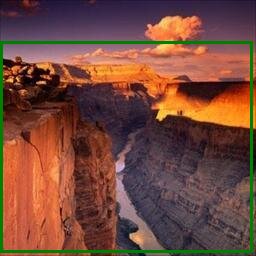}%
\includegraphics[width=0.165\linewidth,clip,trim=240 0 0 0]{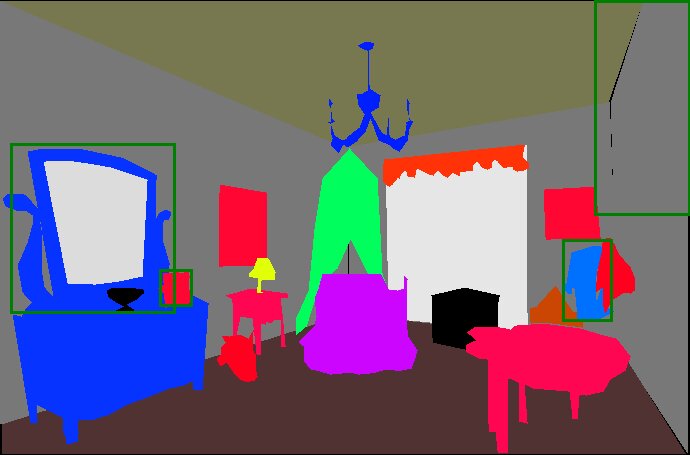}%
\includegraphics[width=0.165\linewidth,clip,trim=240 0 0 0]{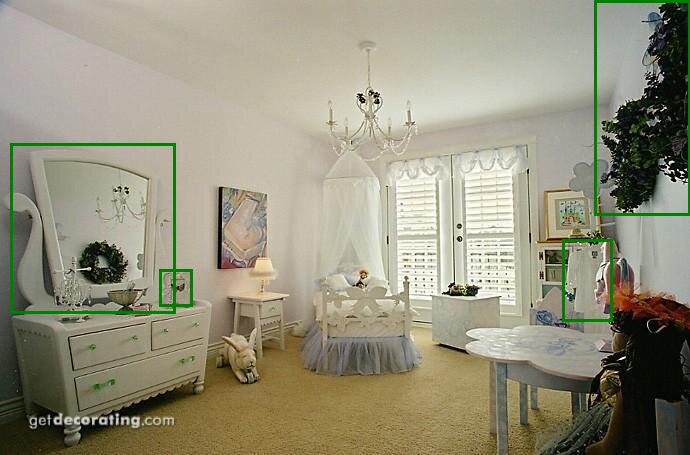}%
\includegraphics[width=0.165\linewidth,clip,trim=0 0 0 165]{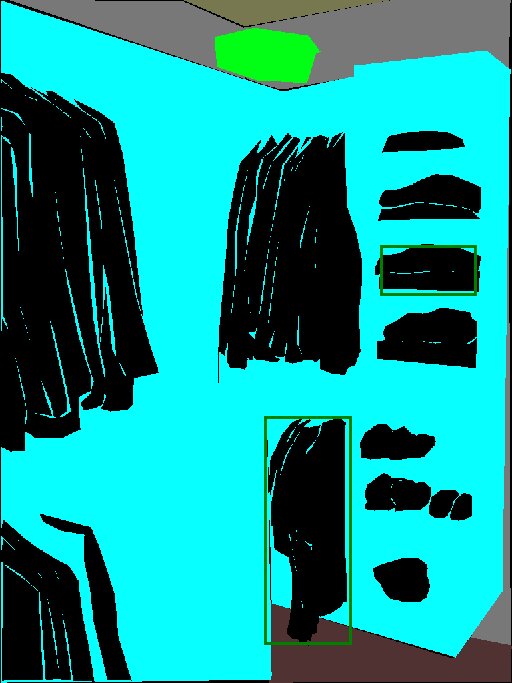}%
\includegraphics[width=0.165\linewidth,clip,trim=0 0 0 165]{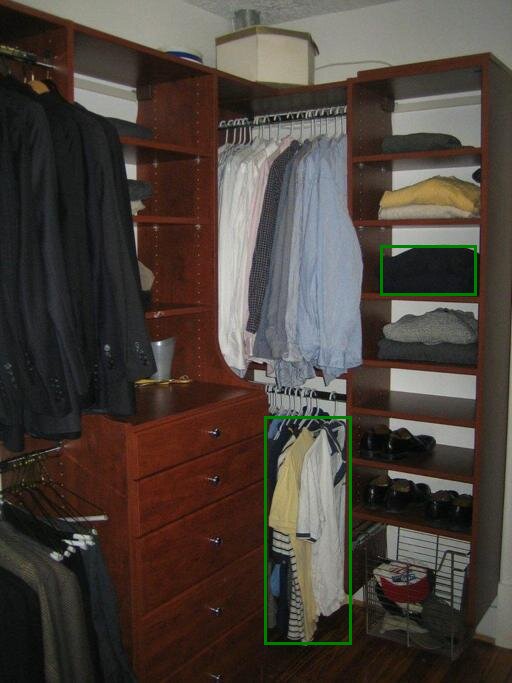}
\hspace{\imgsep}%
\includegraphics[width=0.165\linewidth,clip,trim=0 0 200 100]{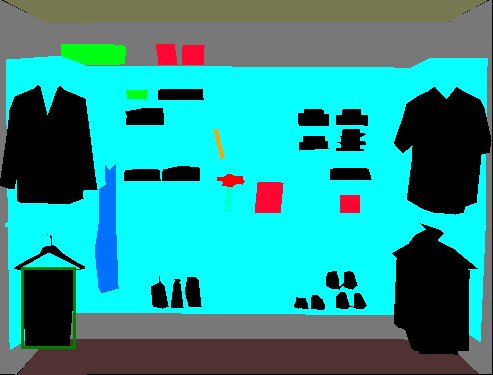}%
\includegraphics[width=0.165\linewidth,clip,trim=0 0 200 100]{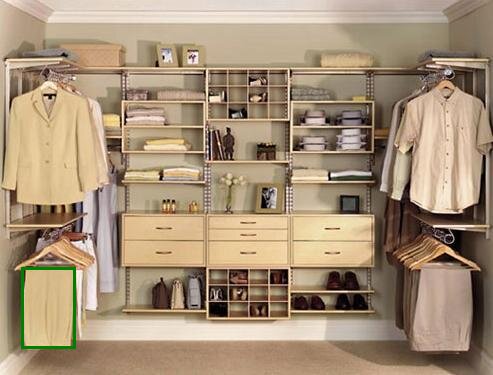}%
\includegraphics[width=0.165\linewidth,clip,trim=0 0 0 200]{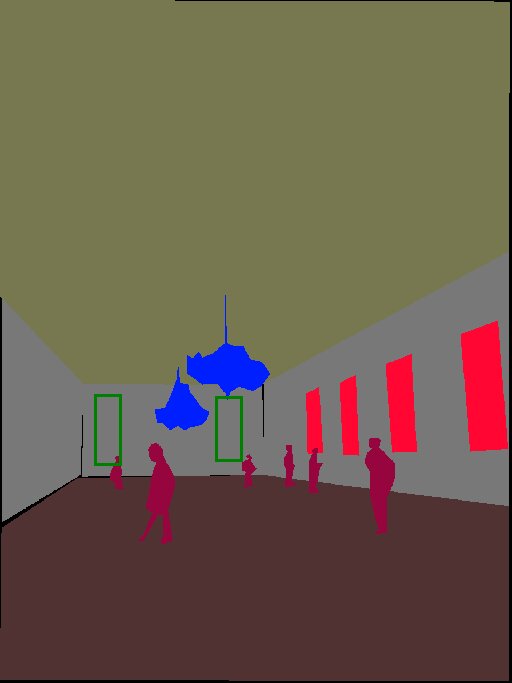}%
\includegraphics[width=0.165\linewidth,clip,trim=0 0 0 200]{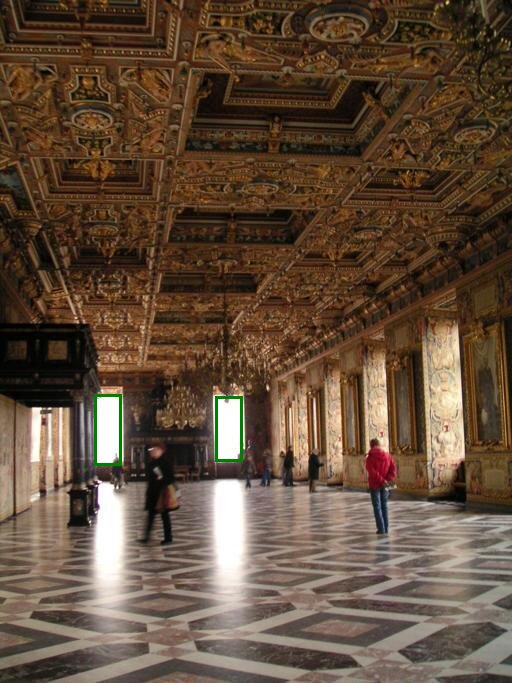}%
\includegraphics[width=0.165\linewidth,clip,trim=220 0 0 0]{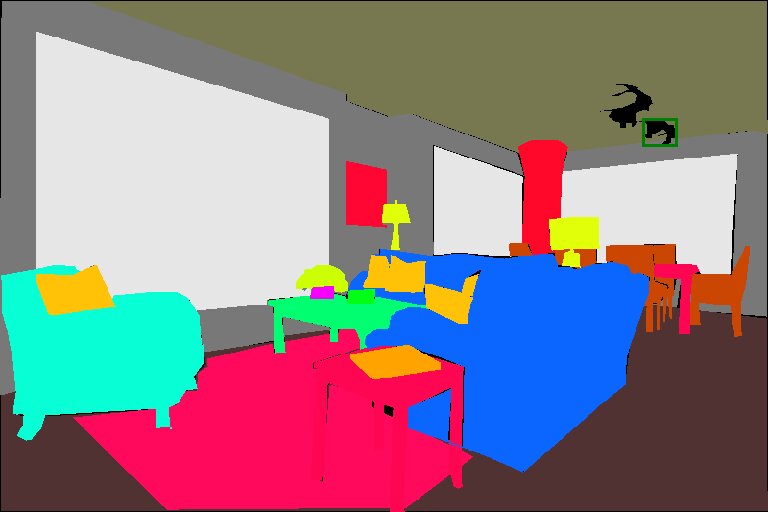}%
\includegraphics[width=0.165\linewidth,clip,trim=220 0 0 0]{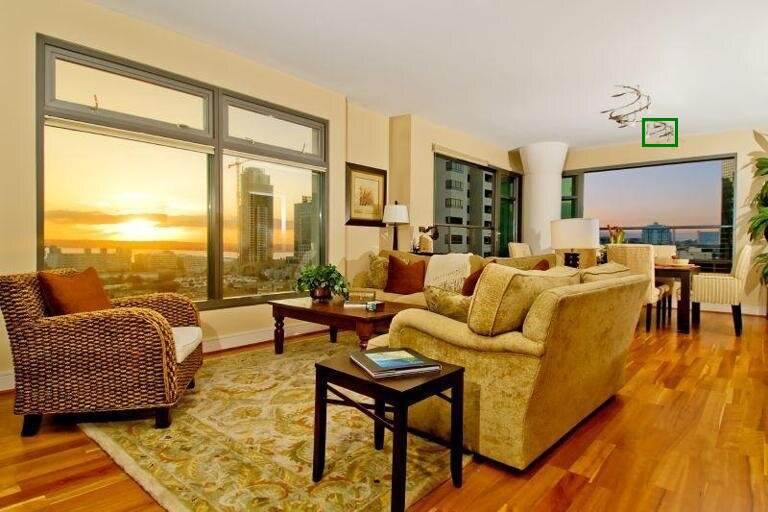}

\caption{A collection of label errors present in ADE20K. The segmentation masks show the ground truth annotations and in green are the prediction of our label error detection method.}
\label{fig:real_ade}
\end{figure}

\clearpage
\begin{figure}
\centering
\newcommand{\imgsep}{1cm} 
\includegraphics[width=0.24\linewidth,clip,trim=350 450 340 10]{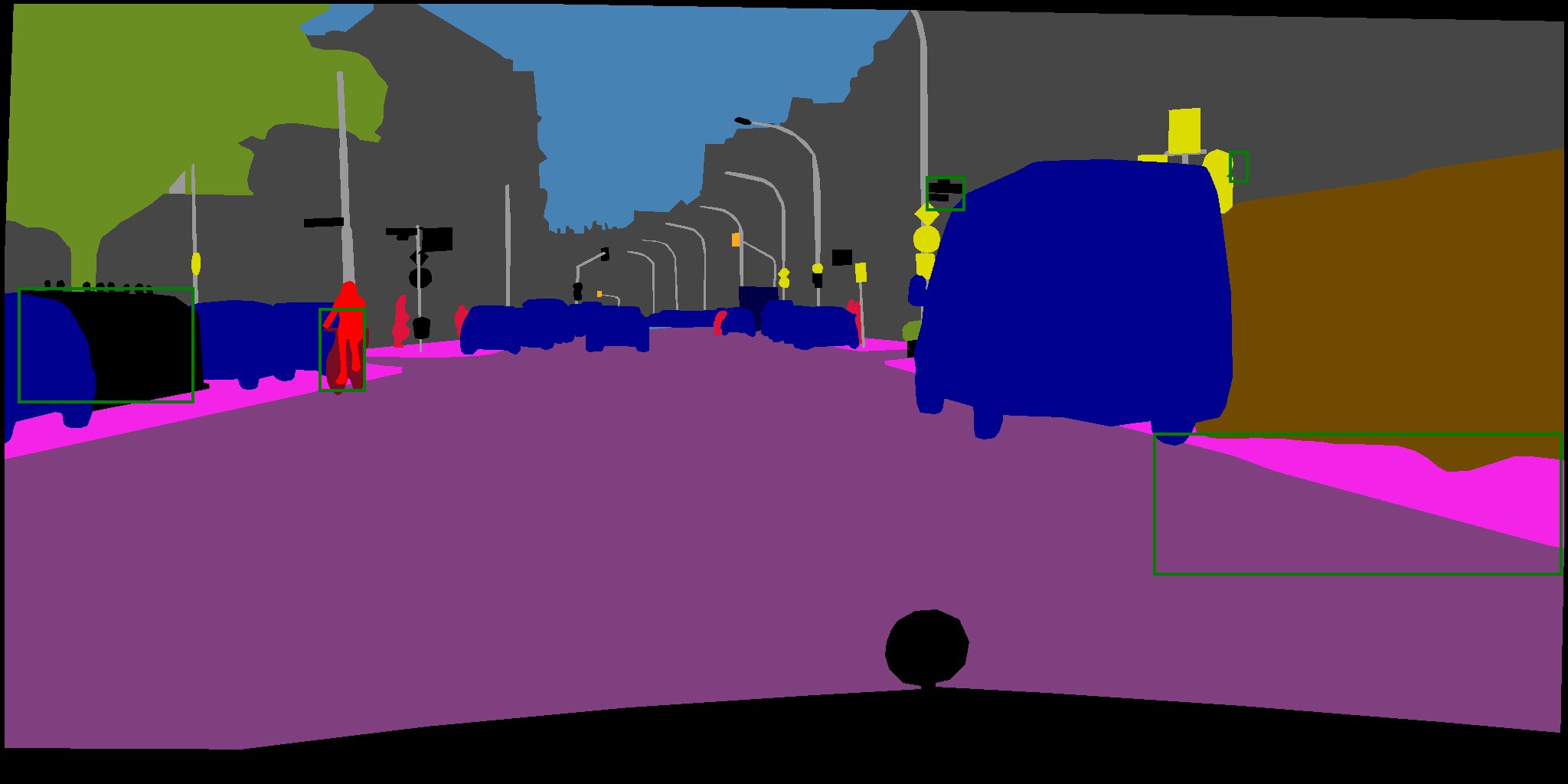}%
\includegraphics[width=0.24\linewidth,clip,trim=350 450 340 10]{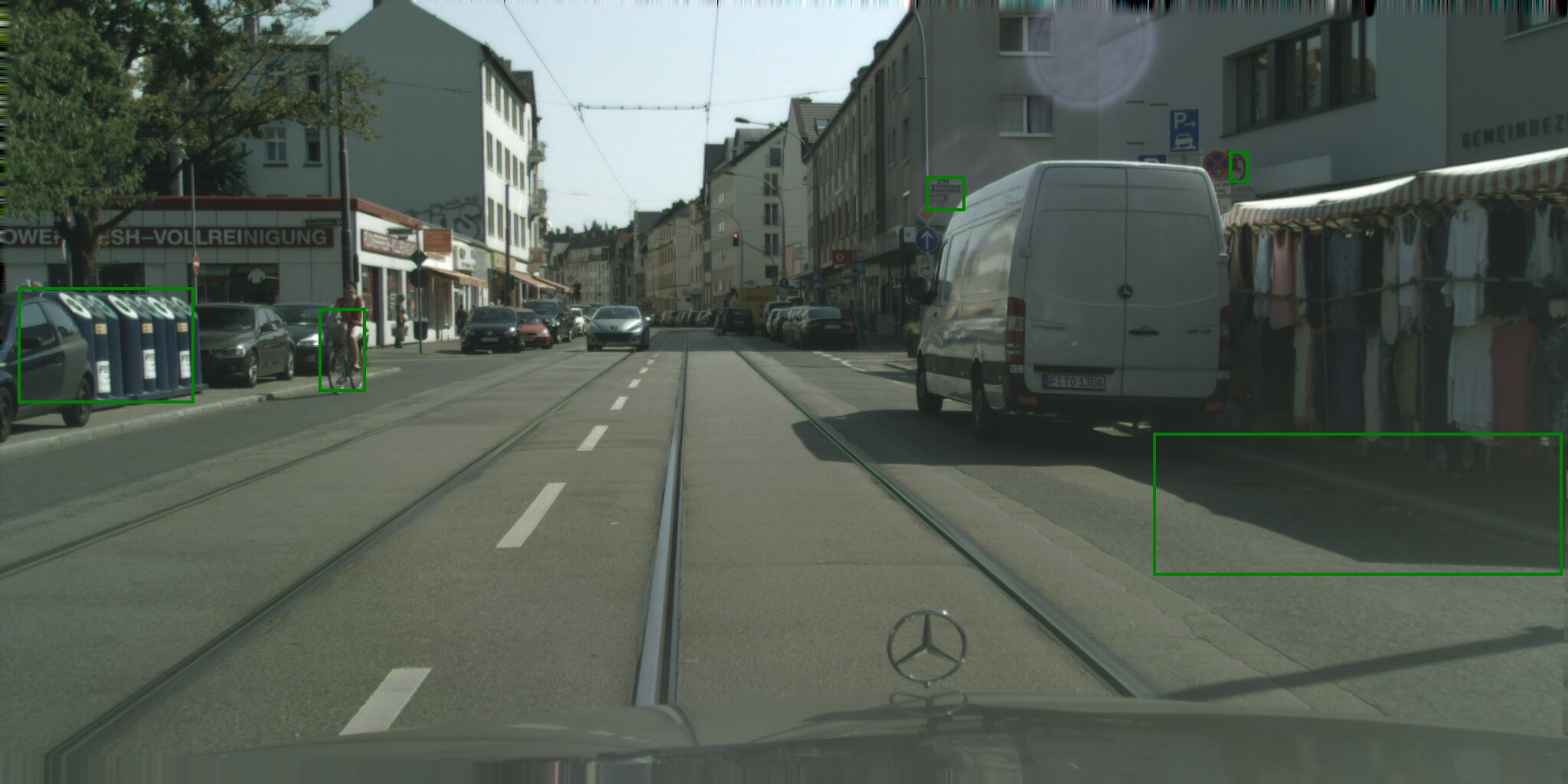}%
\includegraphics[width=0.24\linewidth,clip,trim=200 450 450 0]{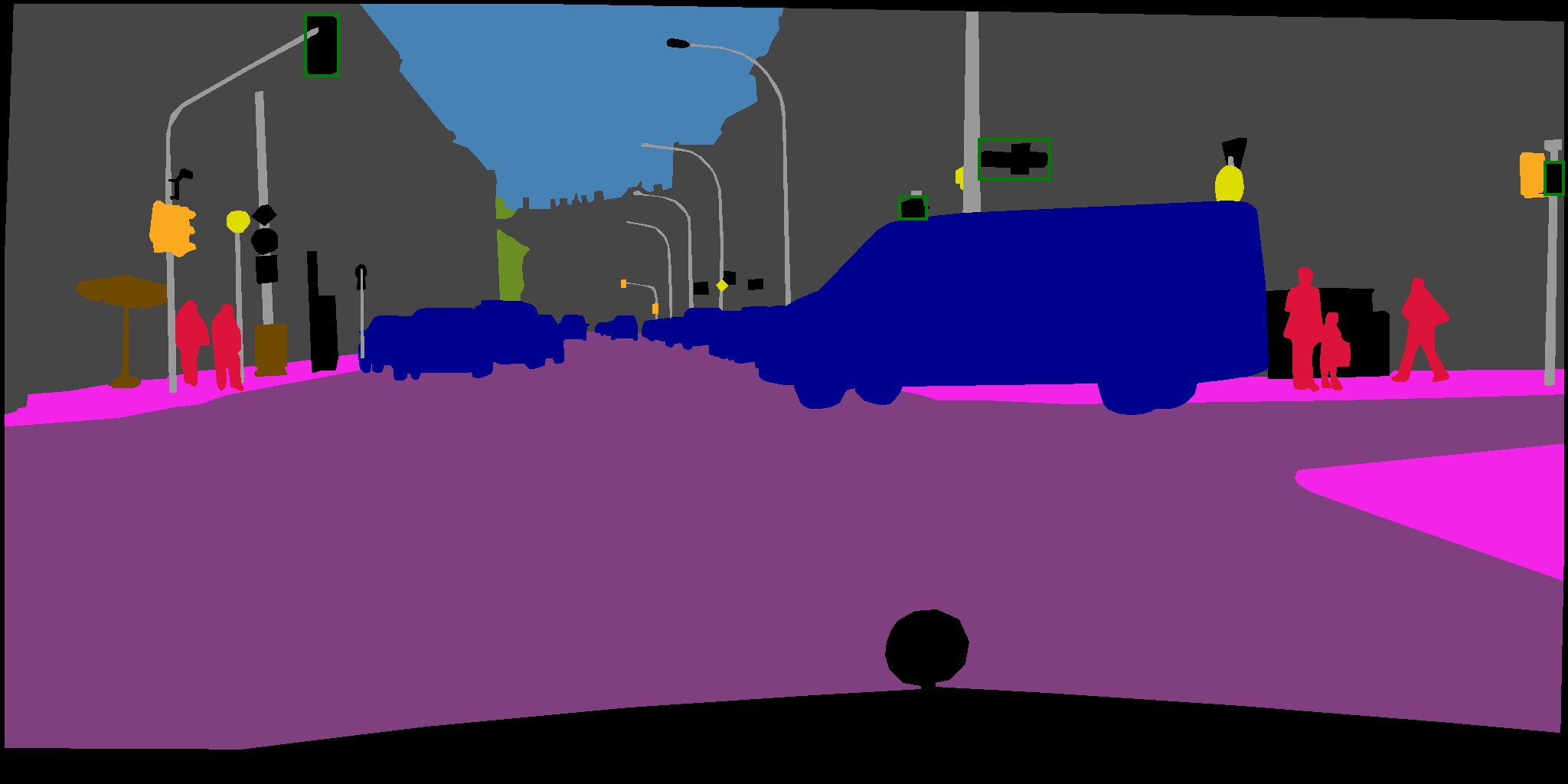}%
\includegraphics[width=0.24\linewidth,clip,trim=200 450 450 0]{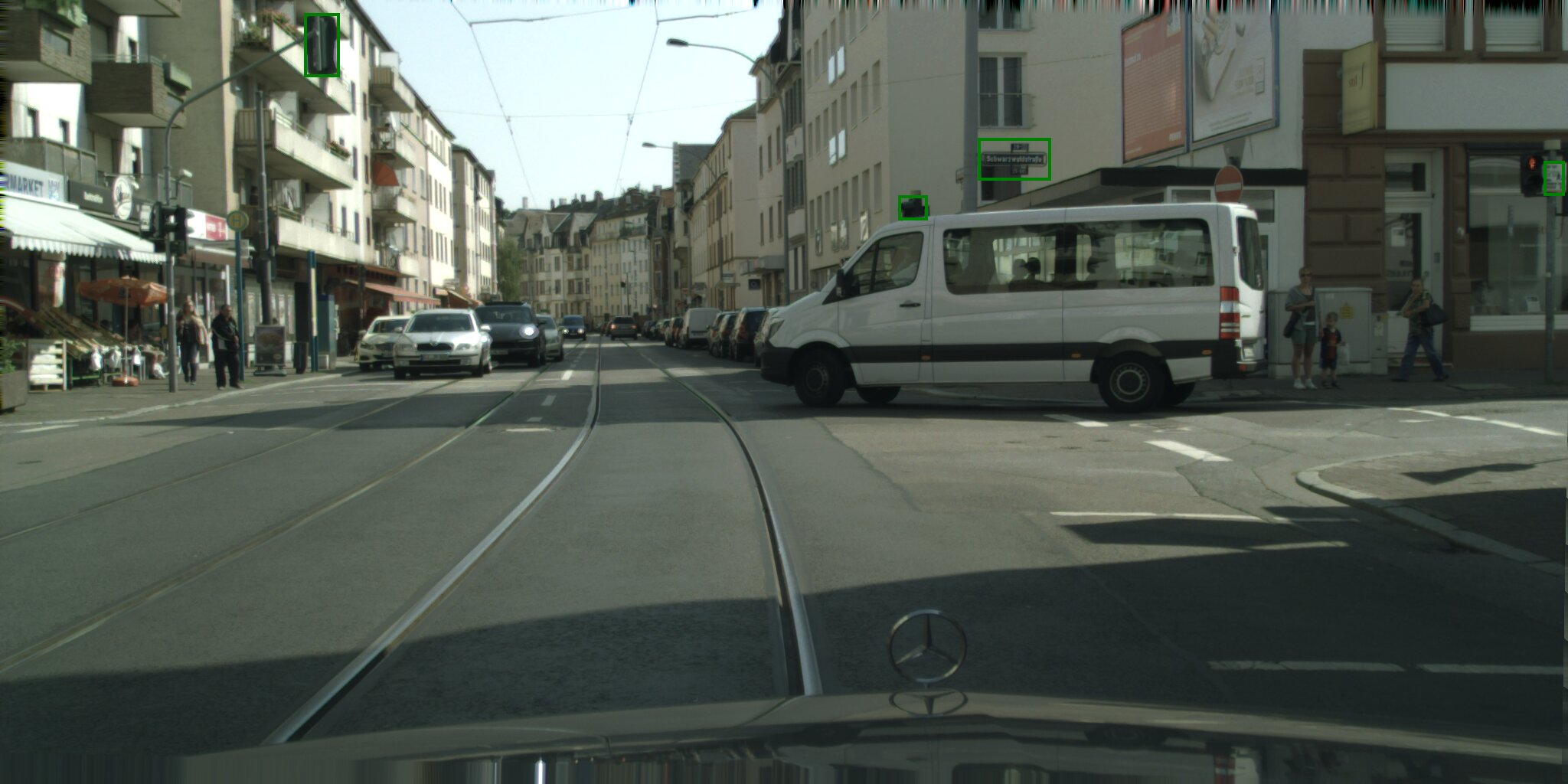}
\hspace{\imgsep}%
\includegraphics[width=0.24\linewidth,clip,trim=880 500 420 160]{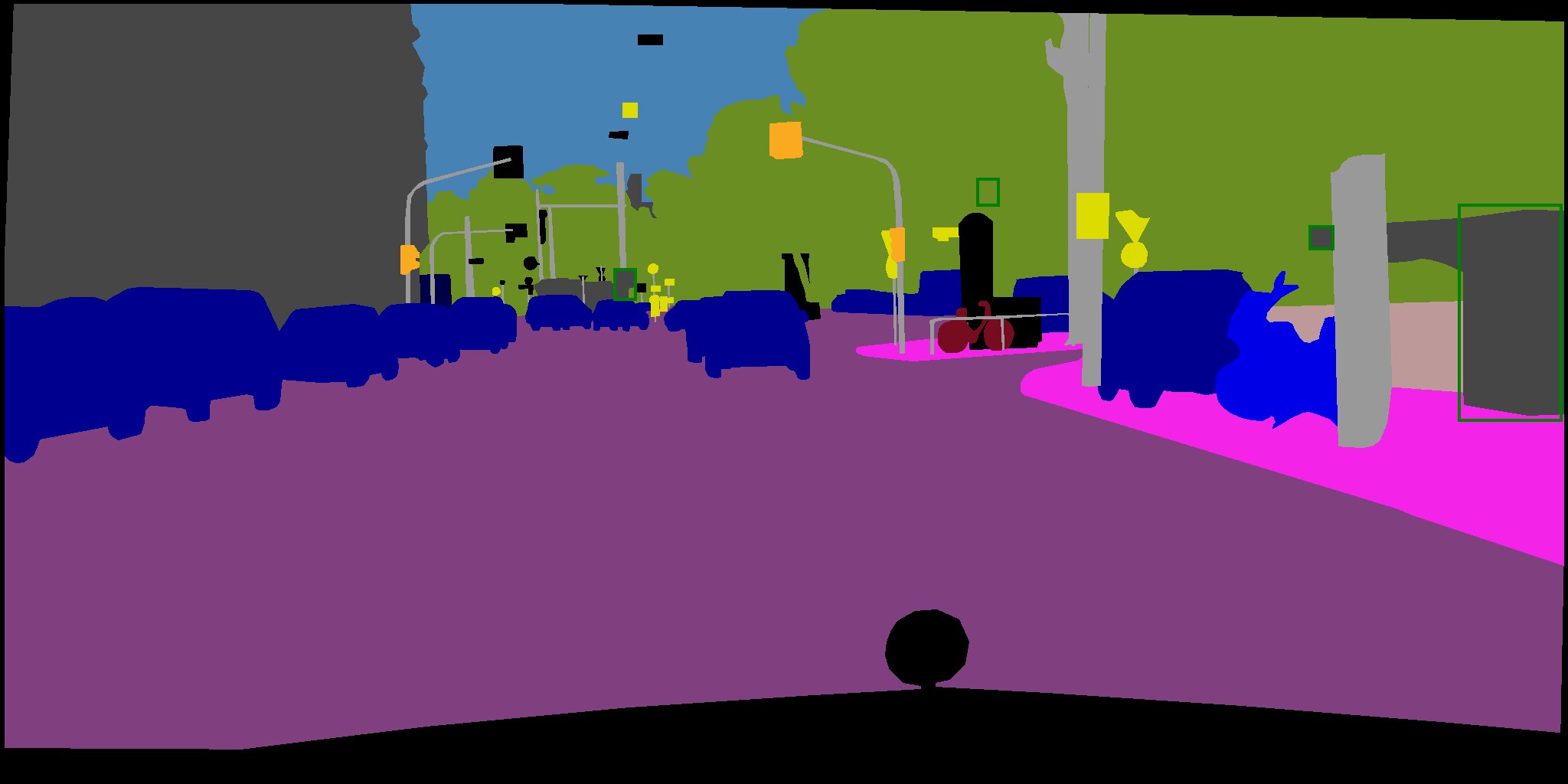}%
\includegraphics[width=0.24\linewidth,clip,trim=880 500 420 160]{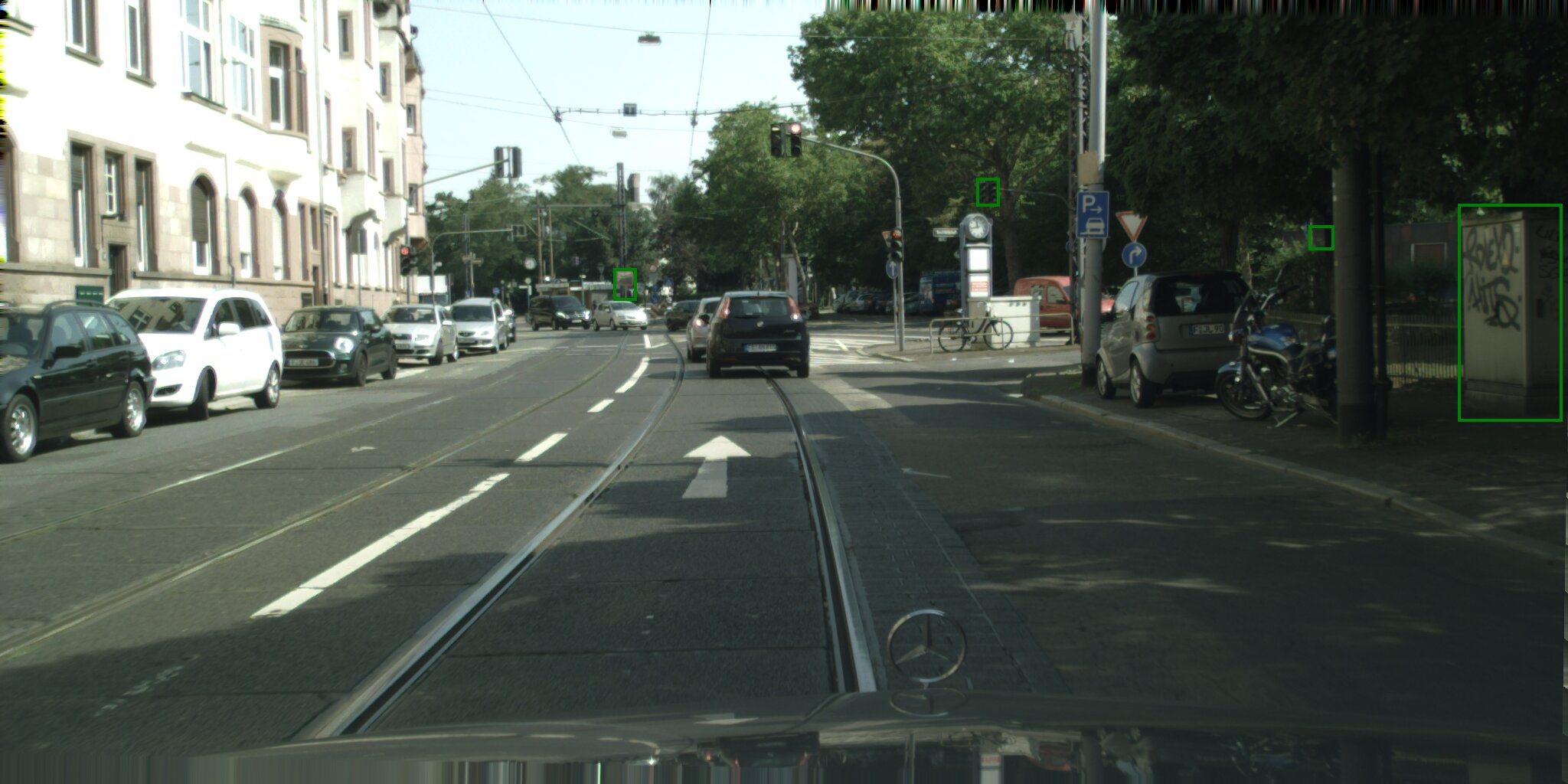}%
\includegraphics[width=0.24\linewidth,clip,trim=1000 500 0 0]{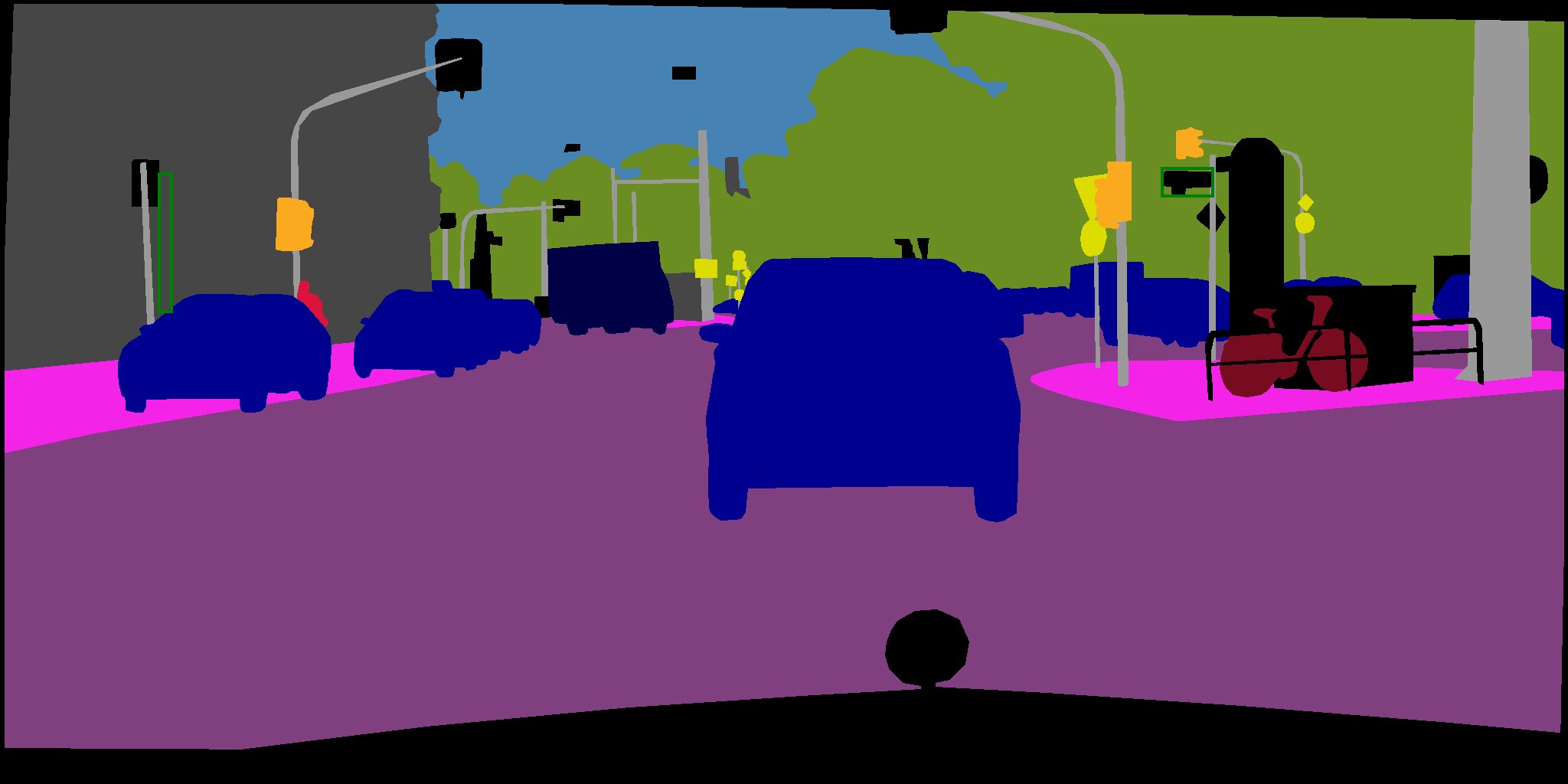}%
\includegraphics[width=0.24\linewidth,clip,trim=1000 500 0 0]{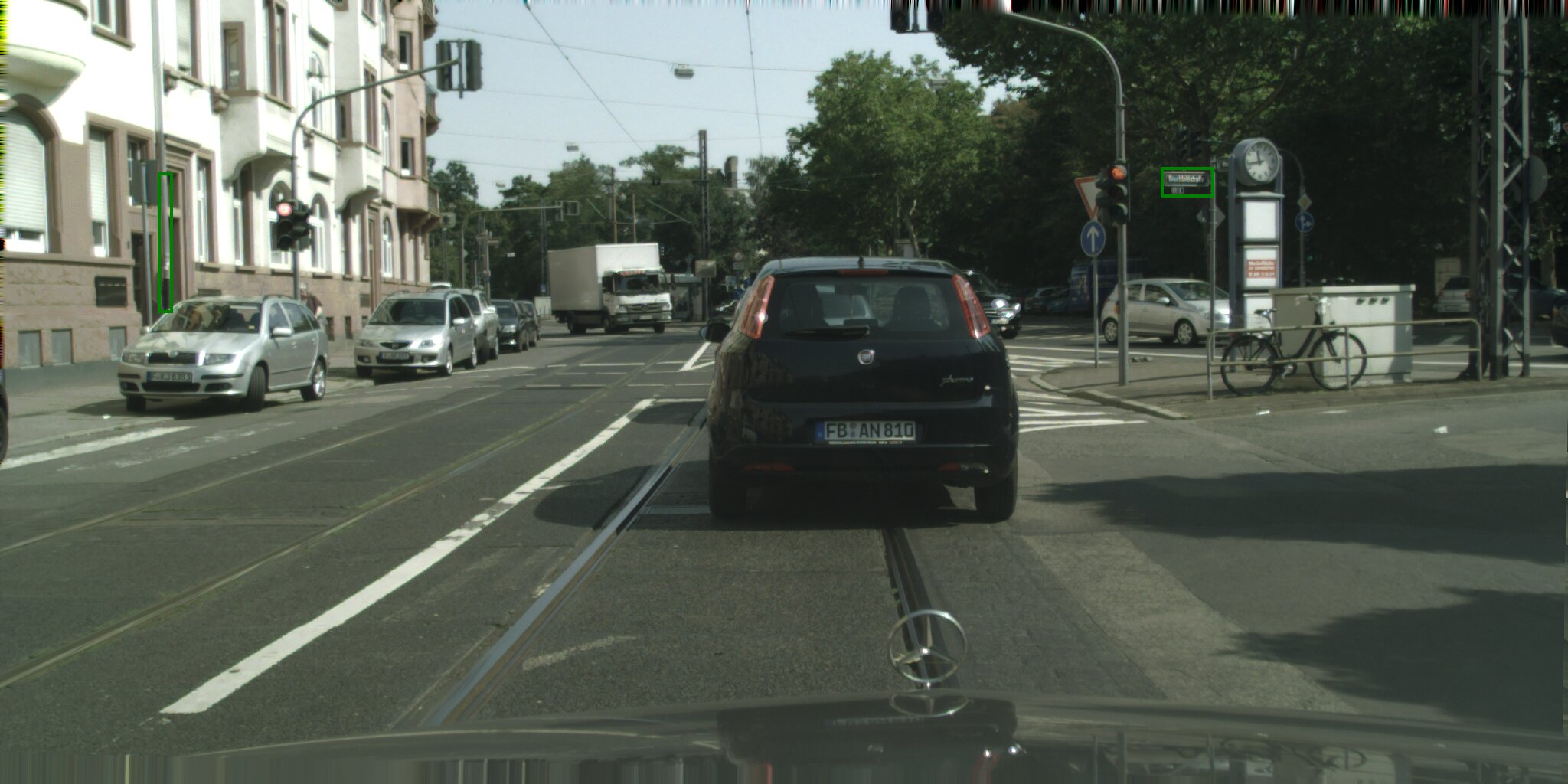}
\hspace{\imgsep}%
\includegraphics[width=0.24\linewidth,clip,trim=400 470 1080 270]{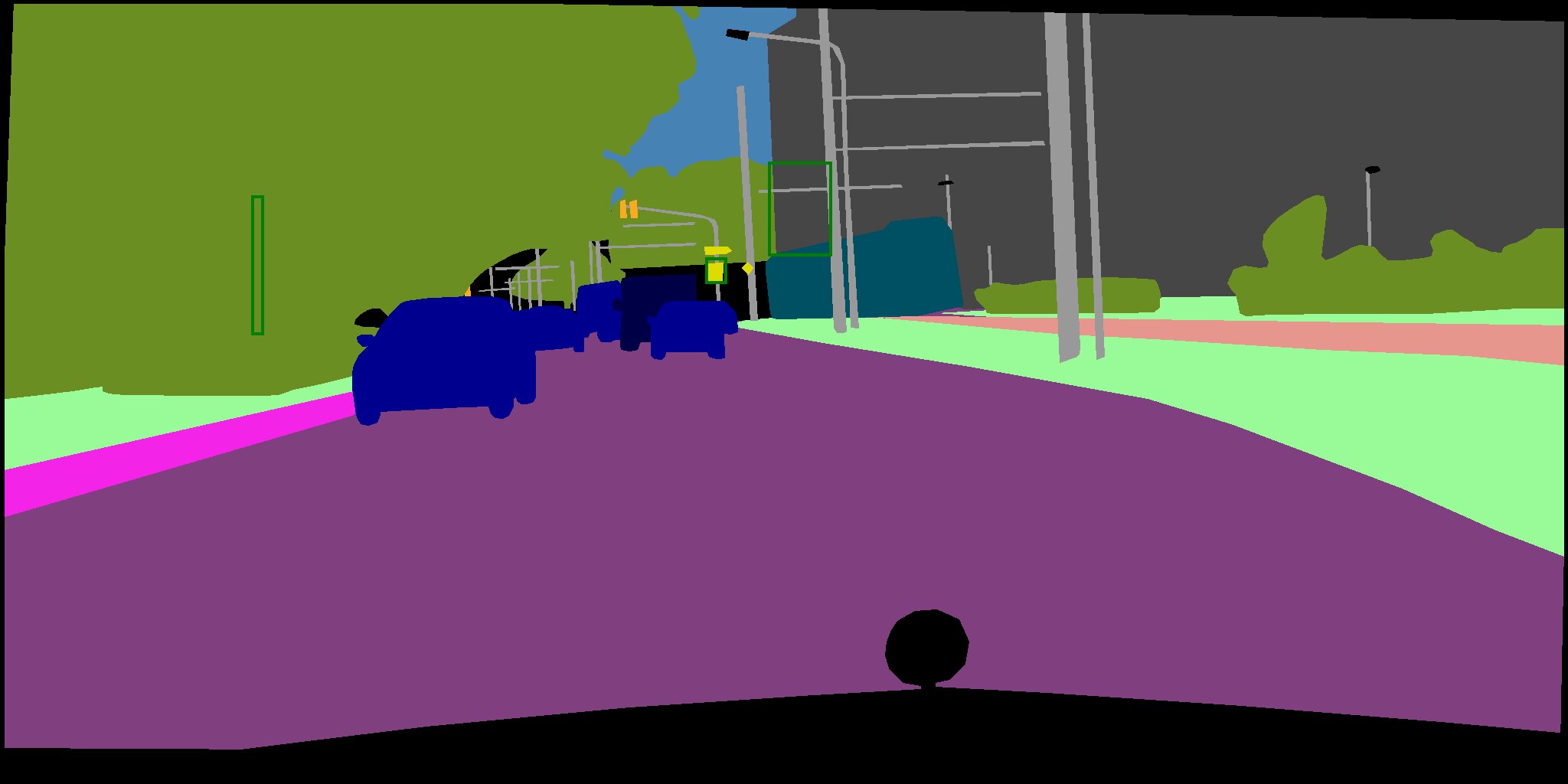}%
\includegraphics[width=0.24\linewidth,clip,trim=400 470 1080 270]{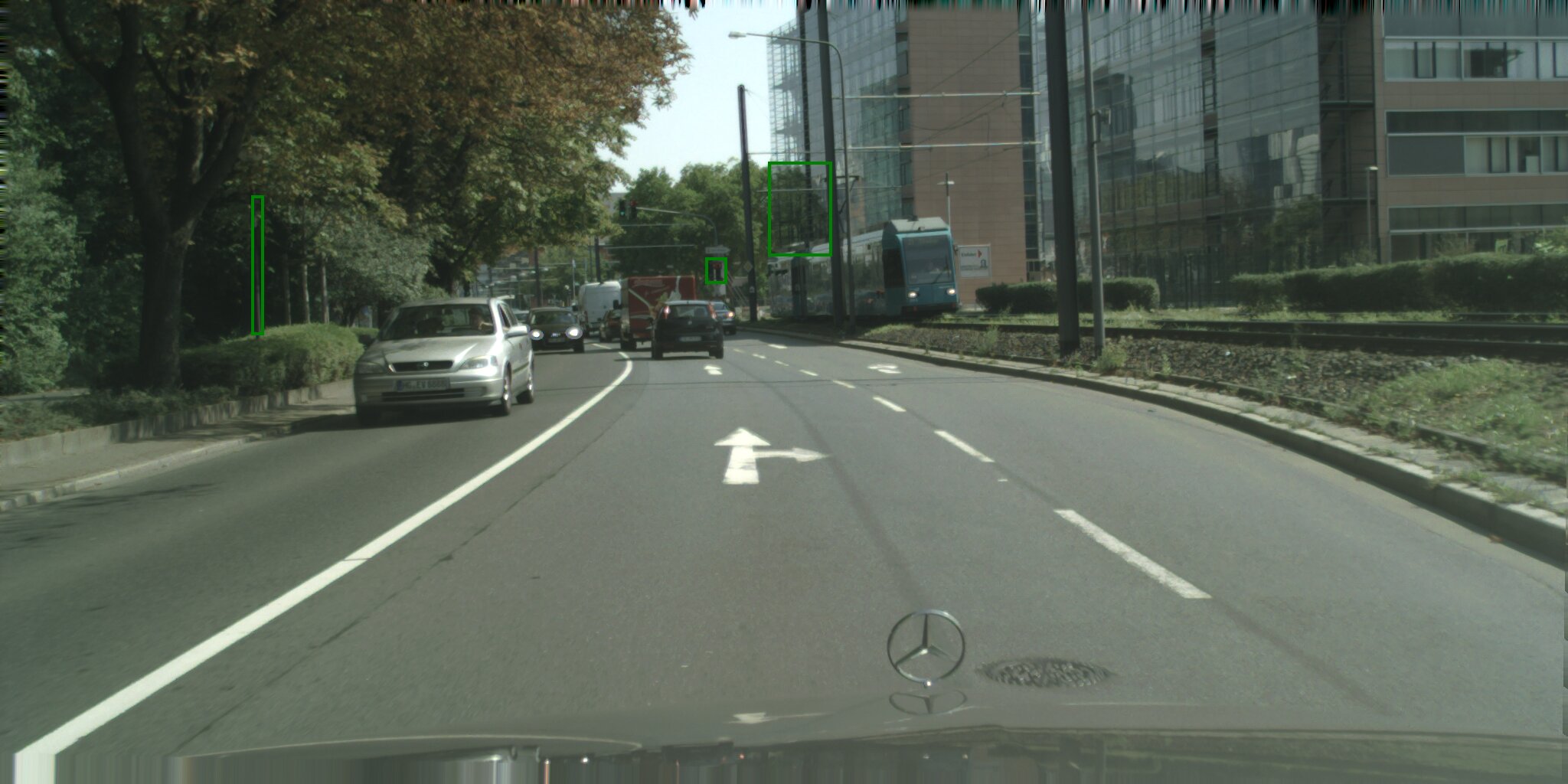}%
\includegraphics[width=0.24\linewidth,clip,trim=0 300 800 100]{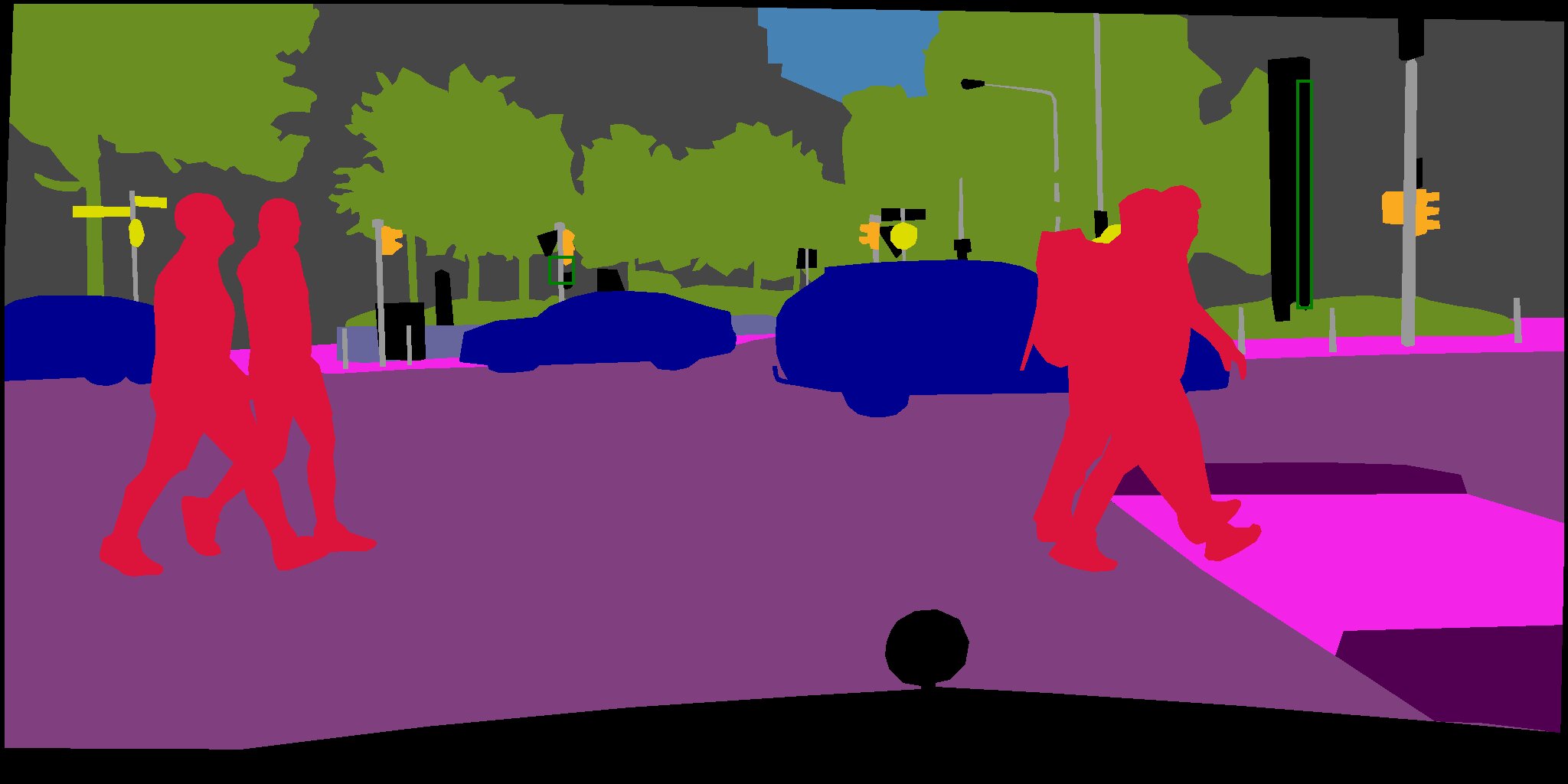}%
\includegraphics[width=0.24\linewidth,clip,trim=0 300 800 100]{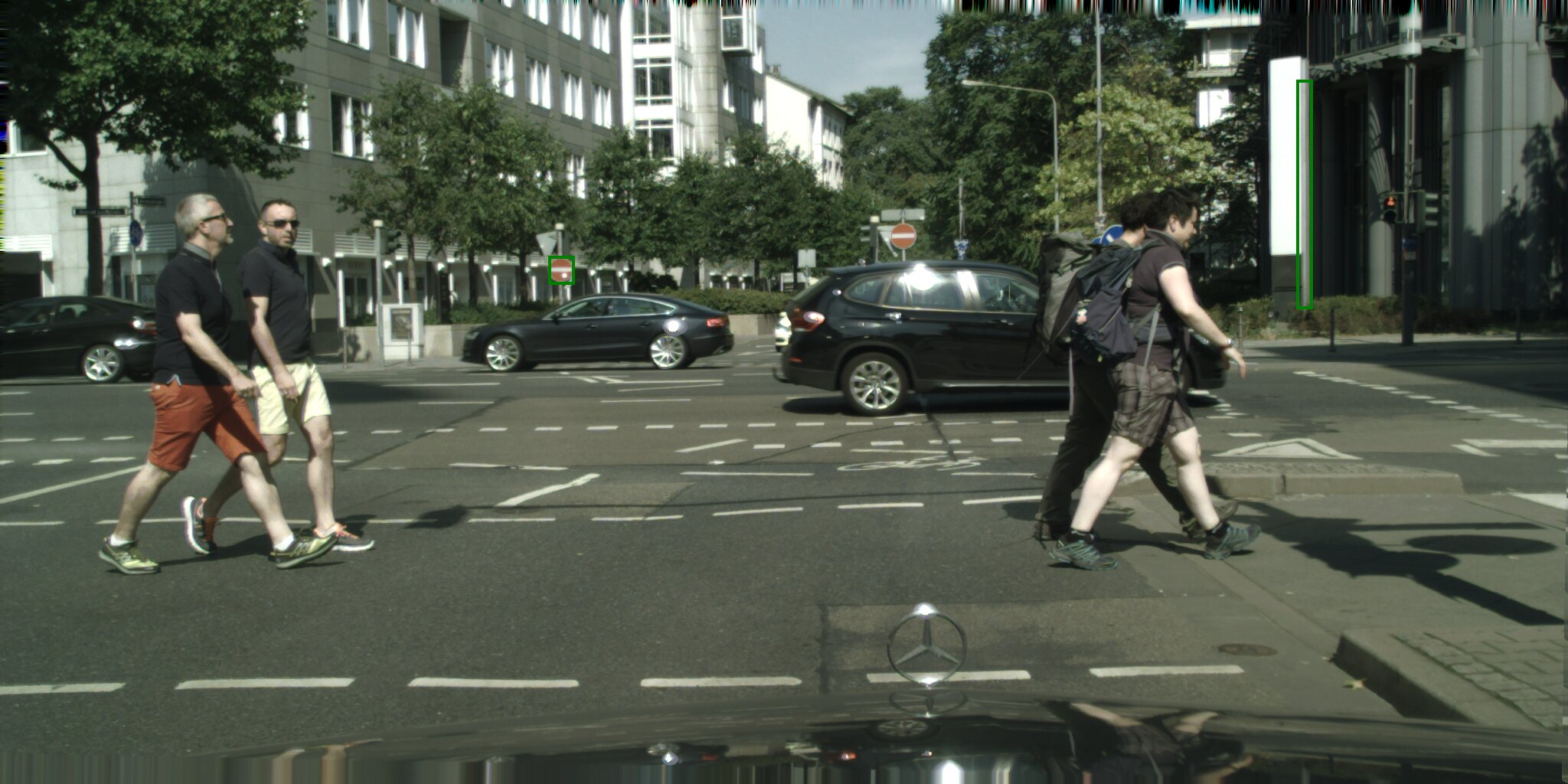}
\hspace{\imgsep}%
\includegraphics[width=0.24\linewidth,clip,trim=1250 400 300 300]{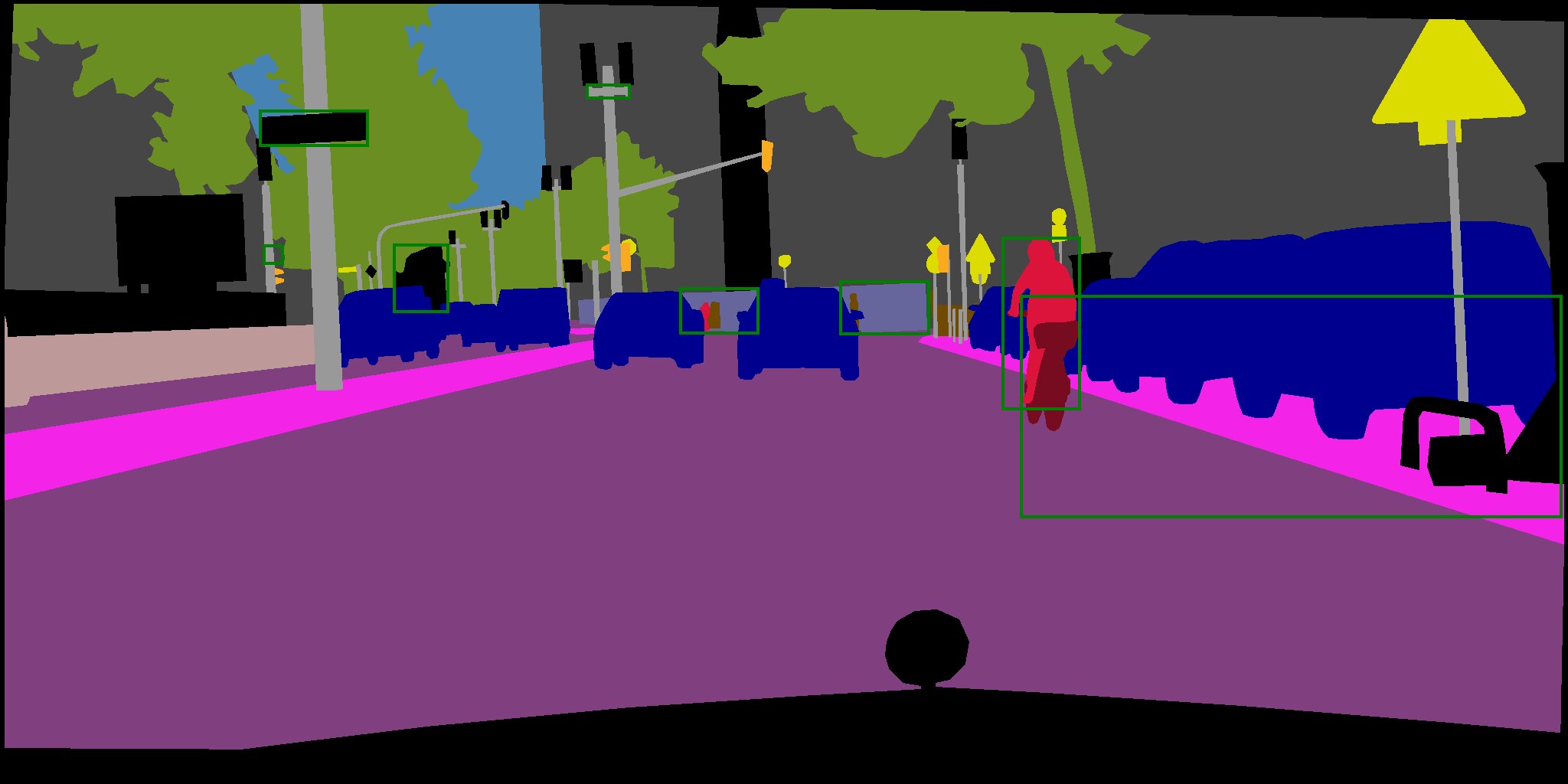}%
\includegraphics[width=0.24\linewidth,clip,trim=1250 400 300 300]{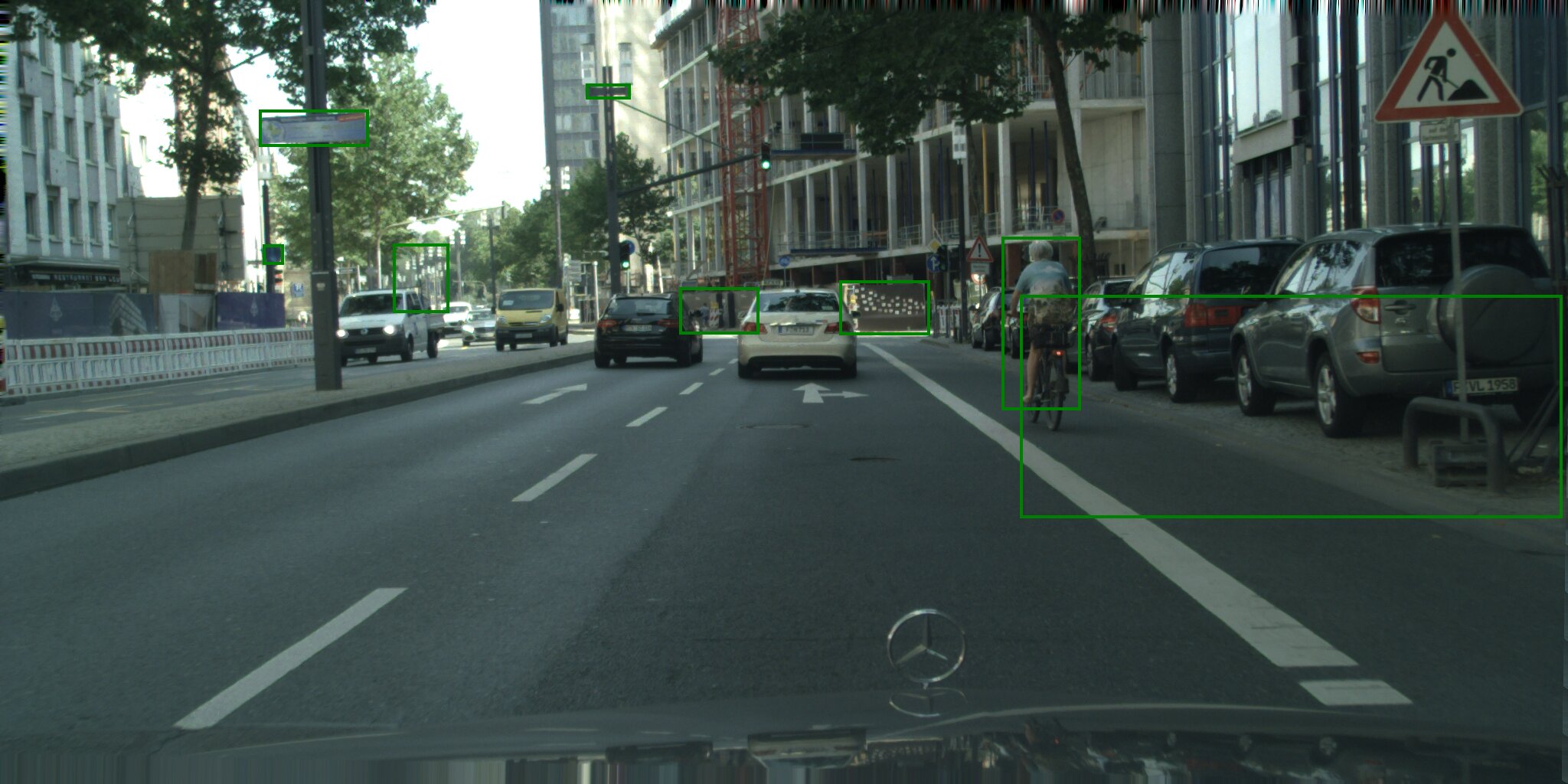}%
\includegraphics[width=0.24\linewidth,clip,trim=1000 340 300 200]{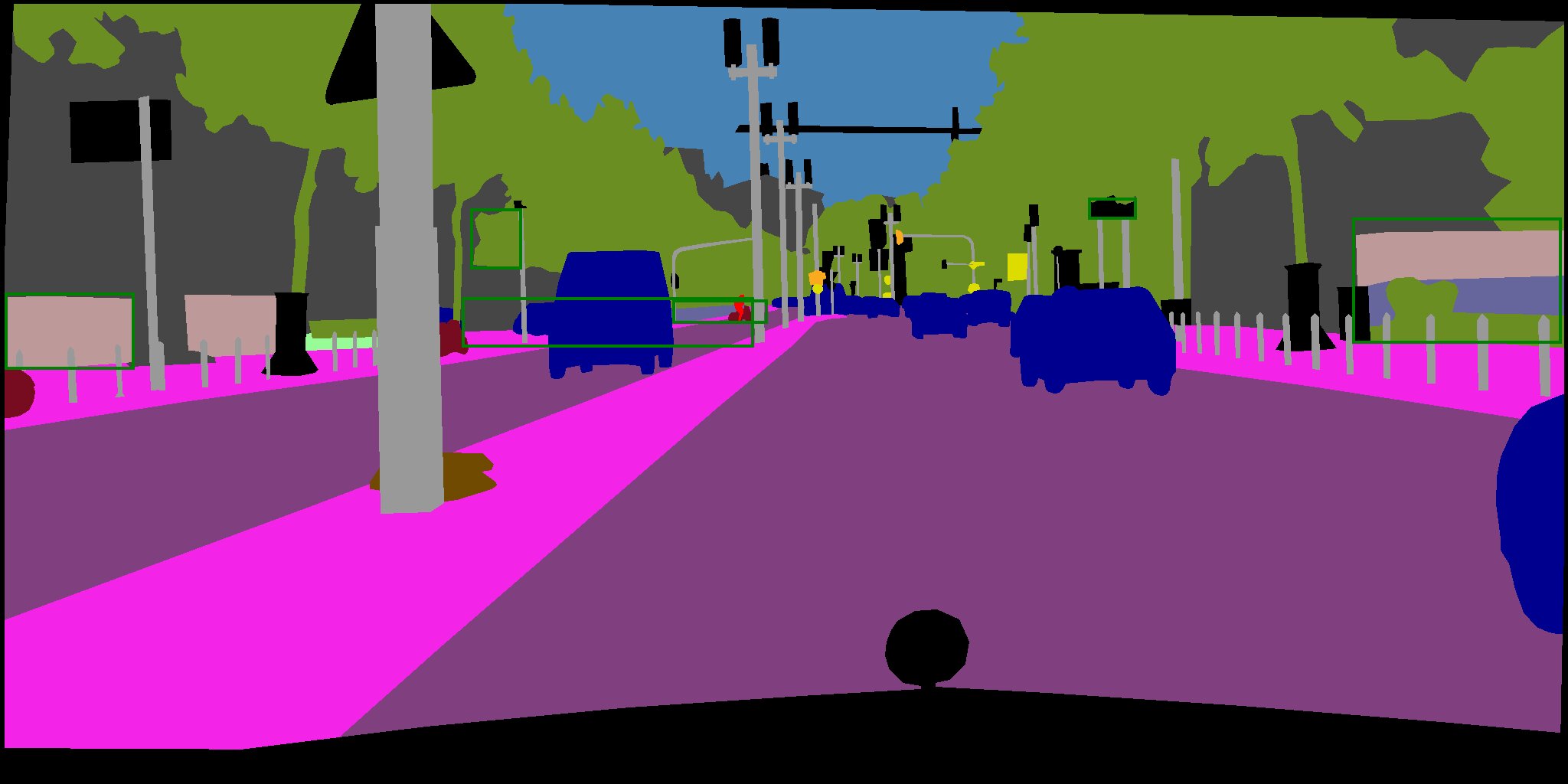}%
\includegraphics[width=0.24\linewidth,clip,trim=1000 340 300 200]{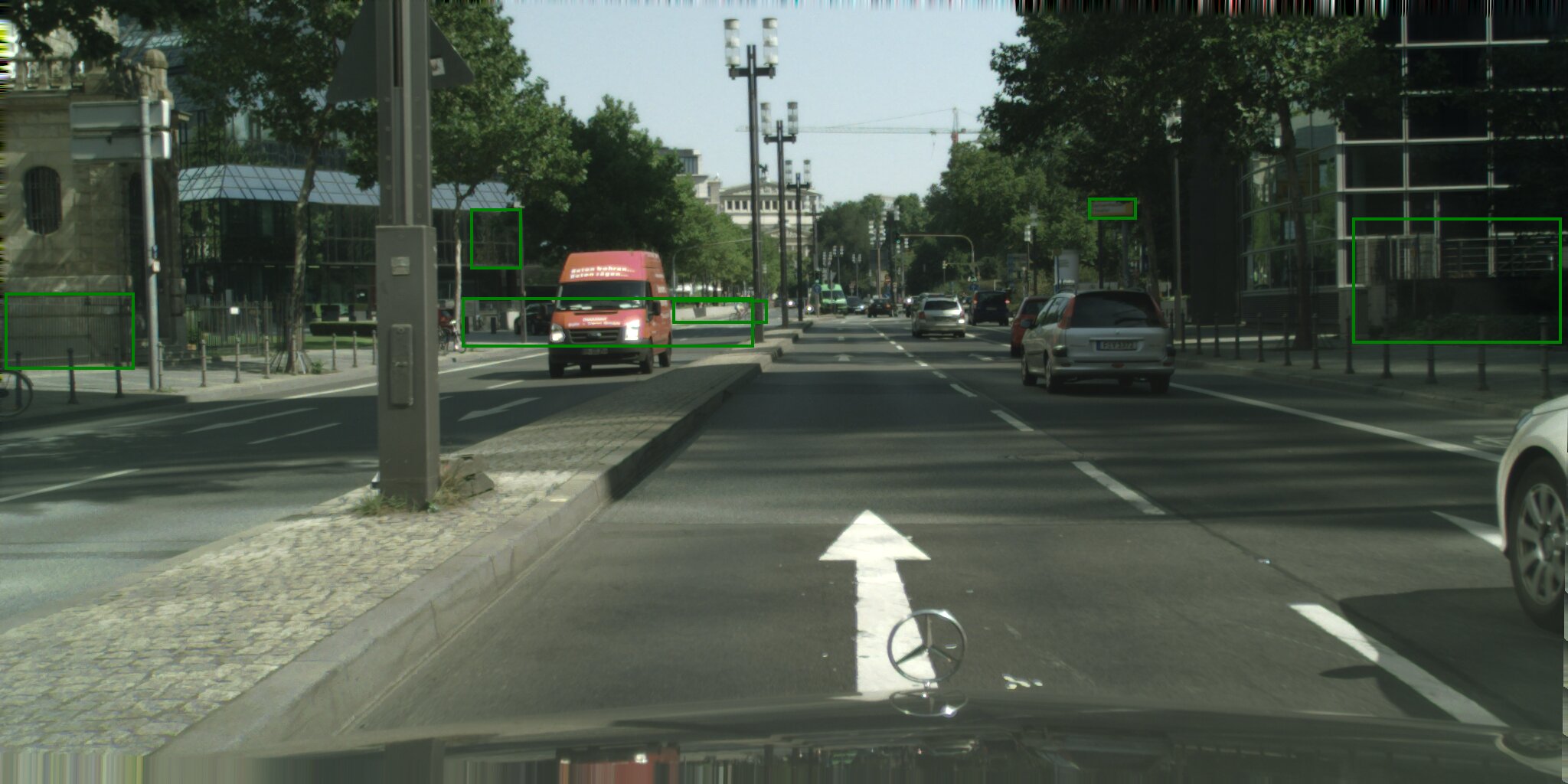}
\hspace{\imgsep}%
\includegraphics[width=0.24\linewidth,clip,trim=700 440 600 100]{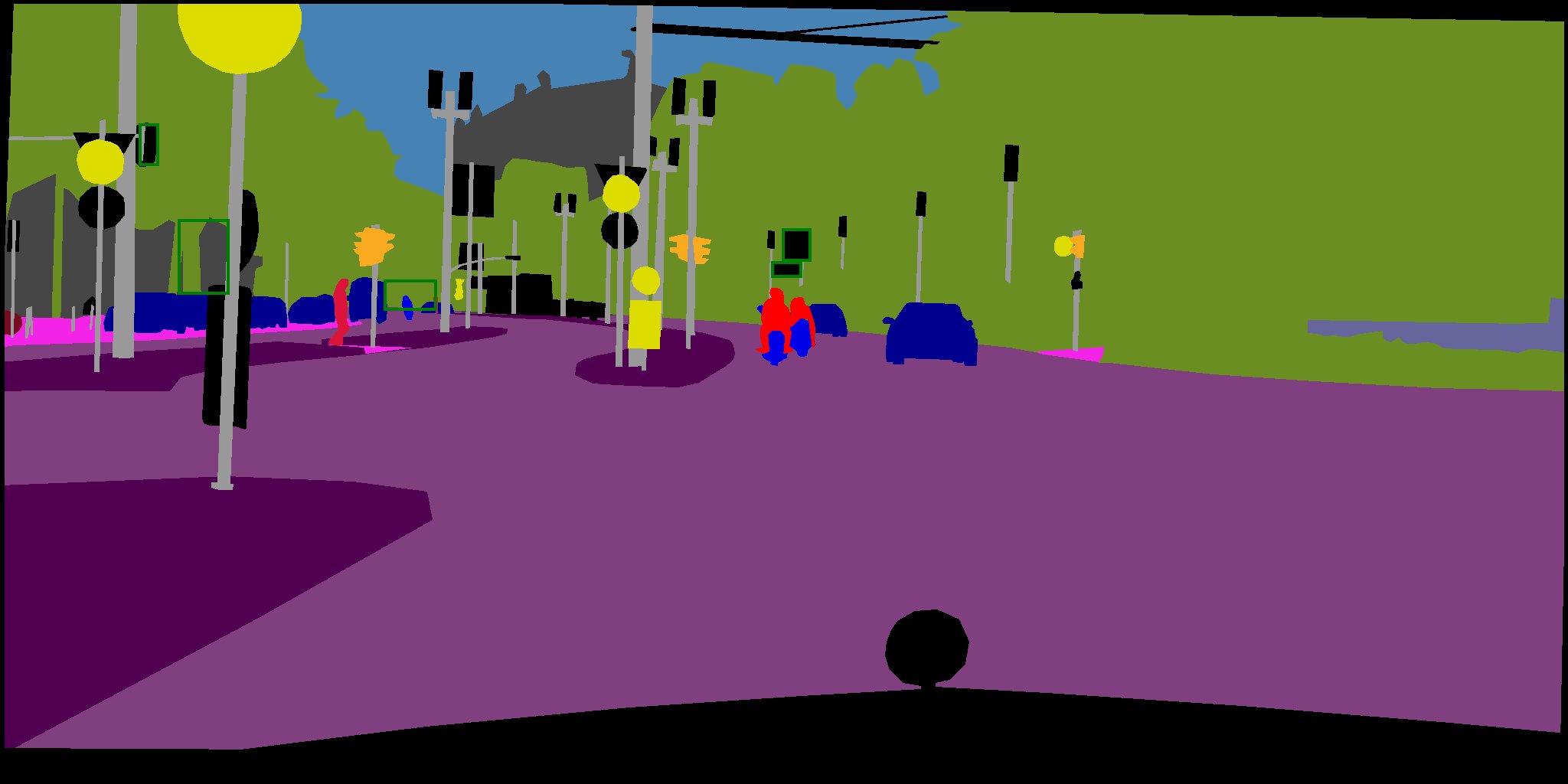}%
\includegraphics[width=0.24\linewidth,clip,trim=700 440 600 100]{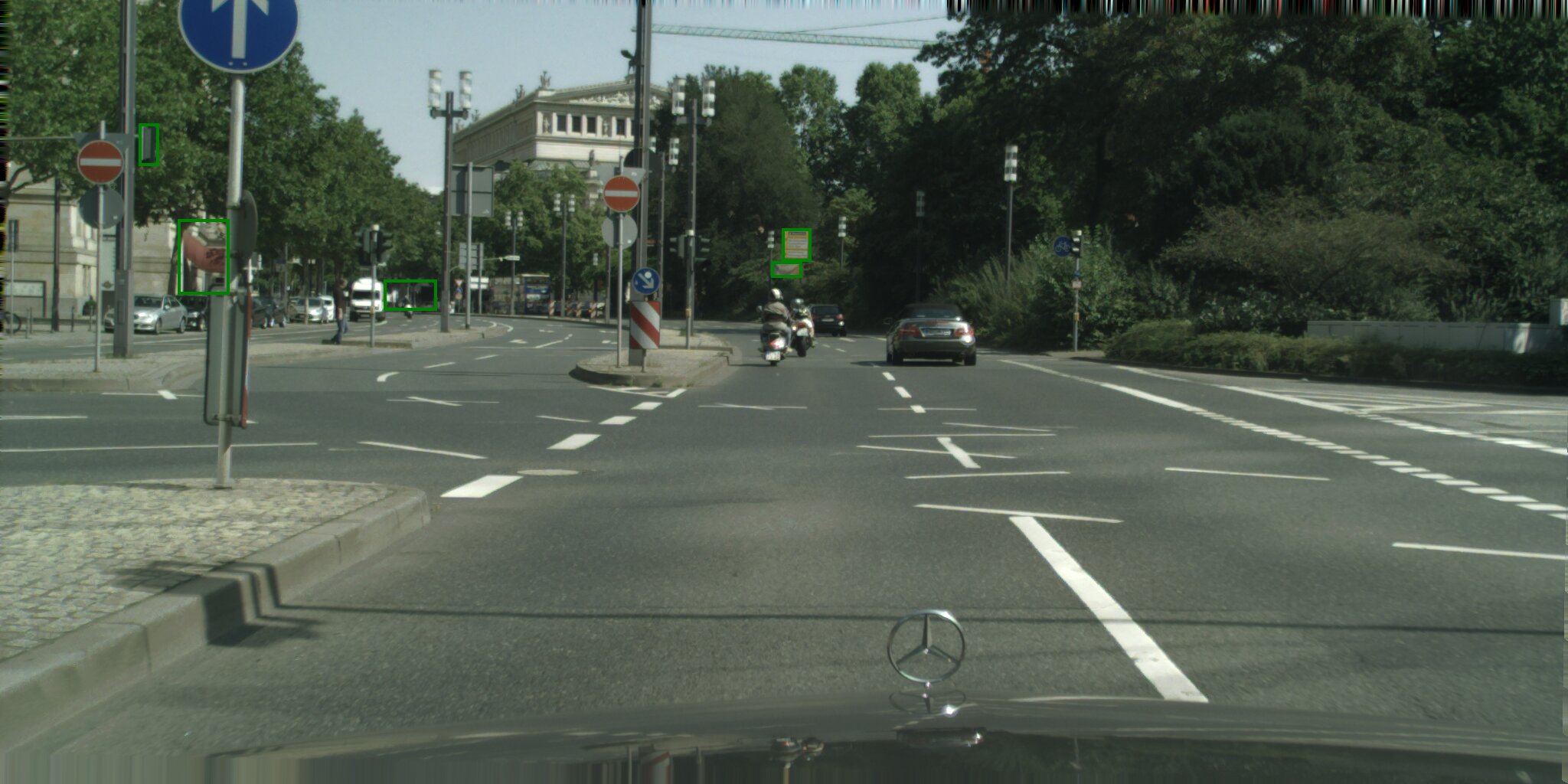}%
\includegraphics[width=0.24\linewidth,clip,trim=1000 360 0 0]{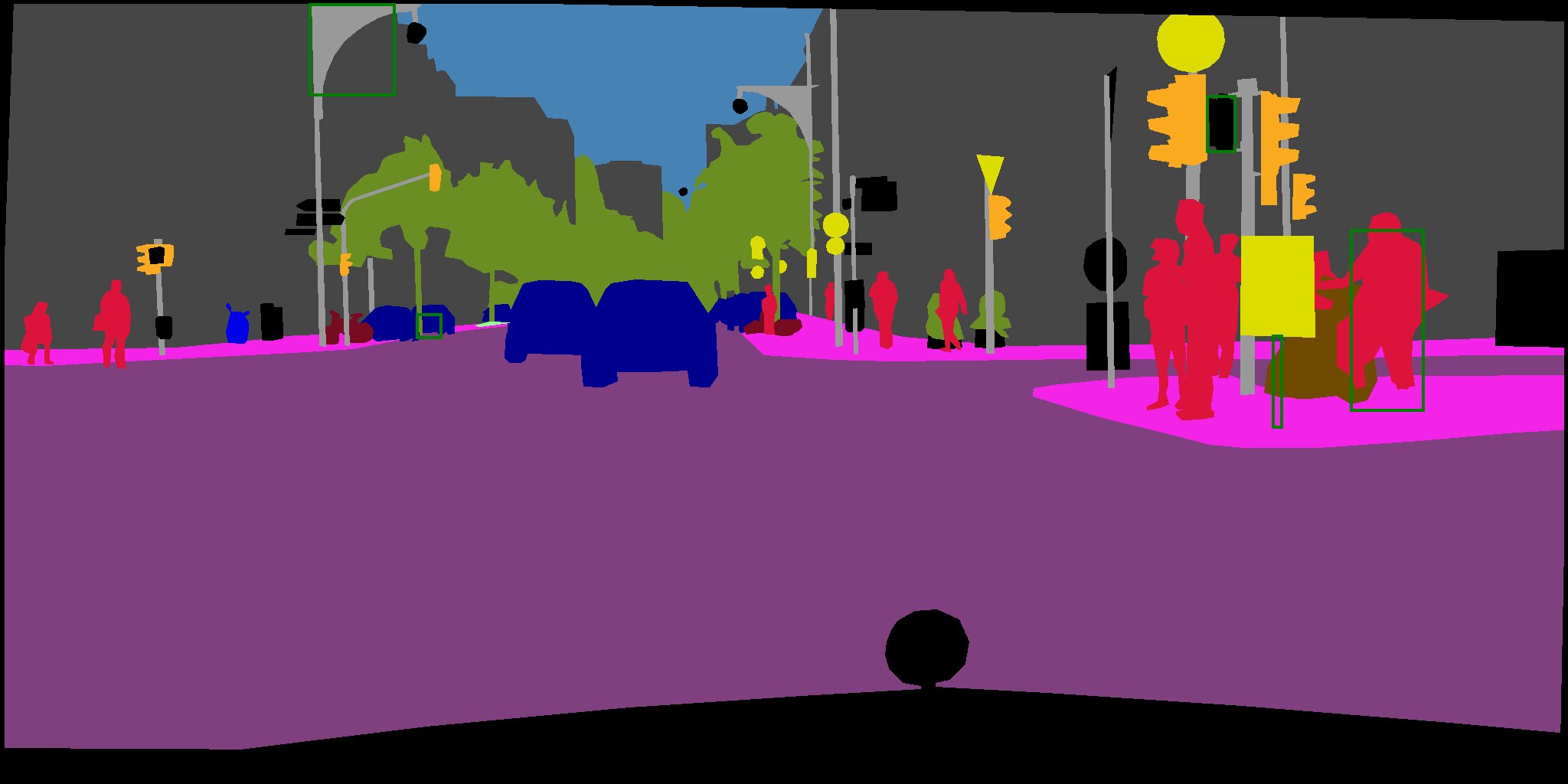}%
\includegraphics[width=0.24\linewidth,clip,,trim=1000 360 0 0]{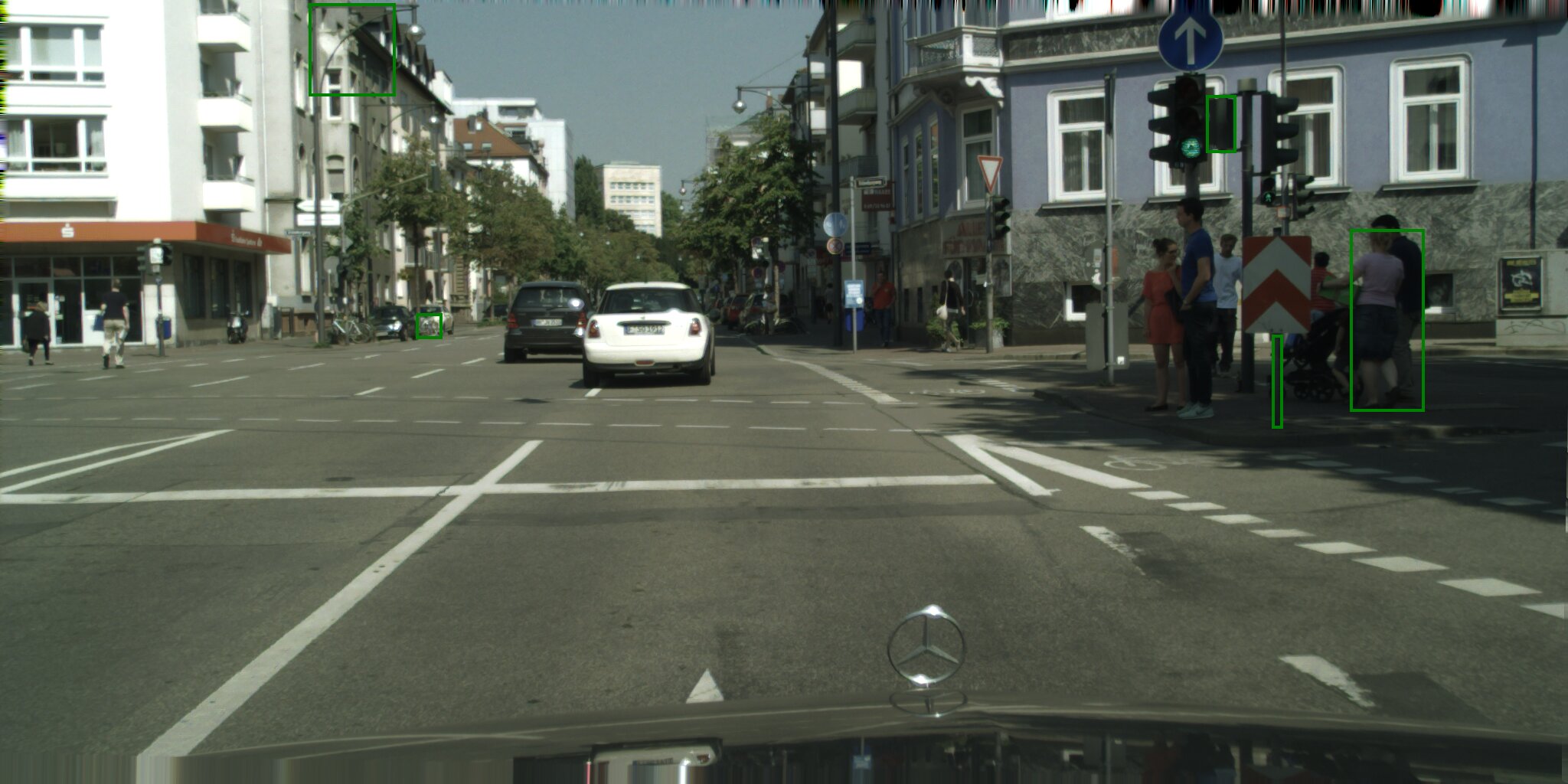}
\hspace{\imgsep}%
\includegraphics[width=0.24\linewidth,clip,trim=320 440 1010 200]{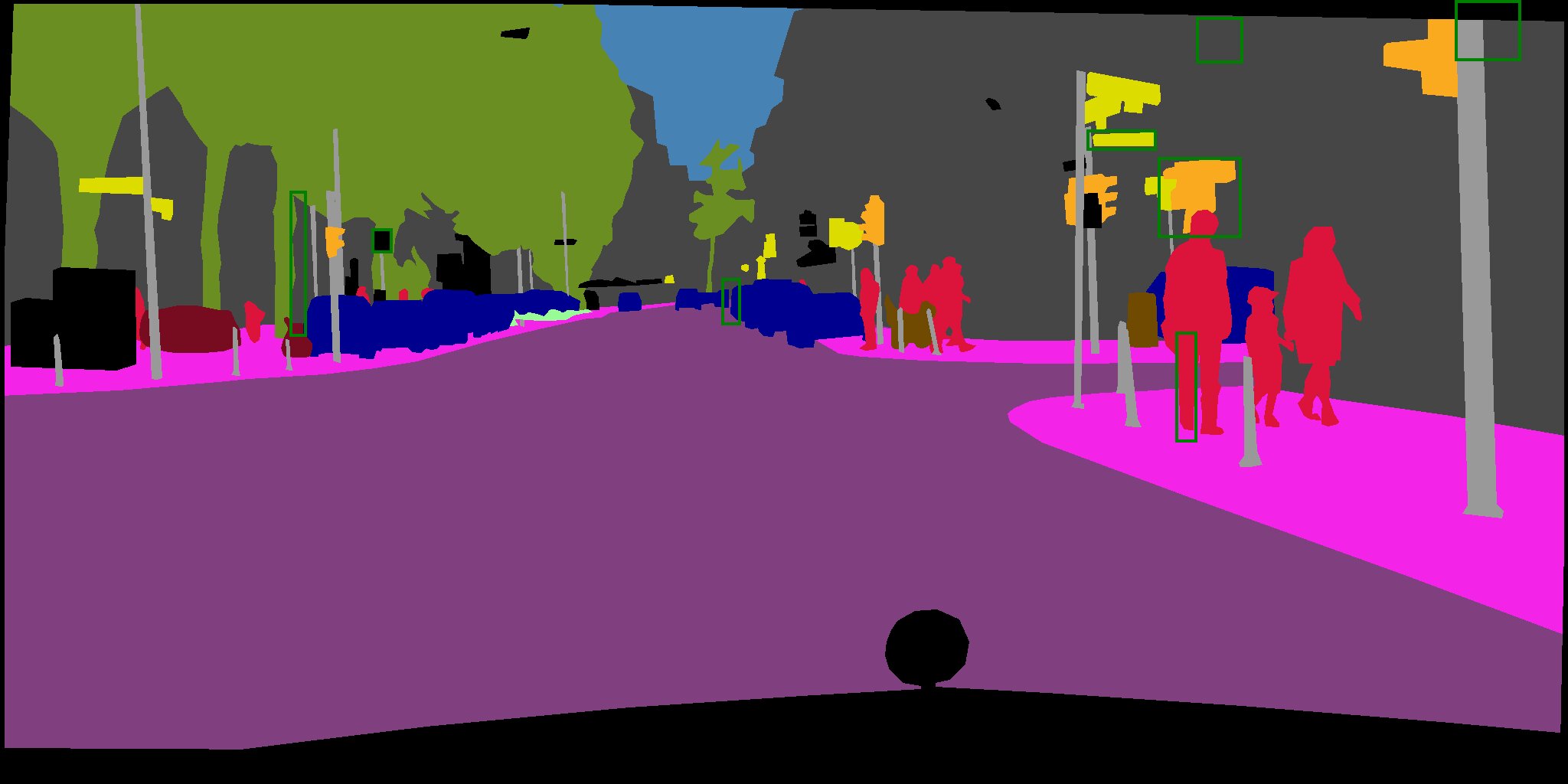}%
\includegraphics[width=0.24\linewidth,clip,trim=320 440 1010 200]{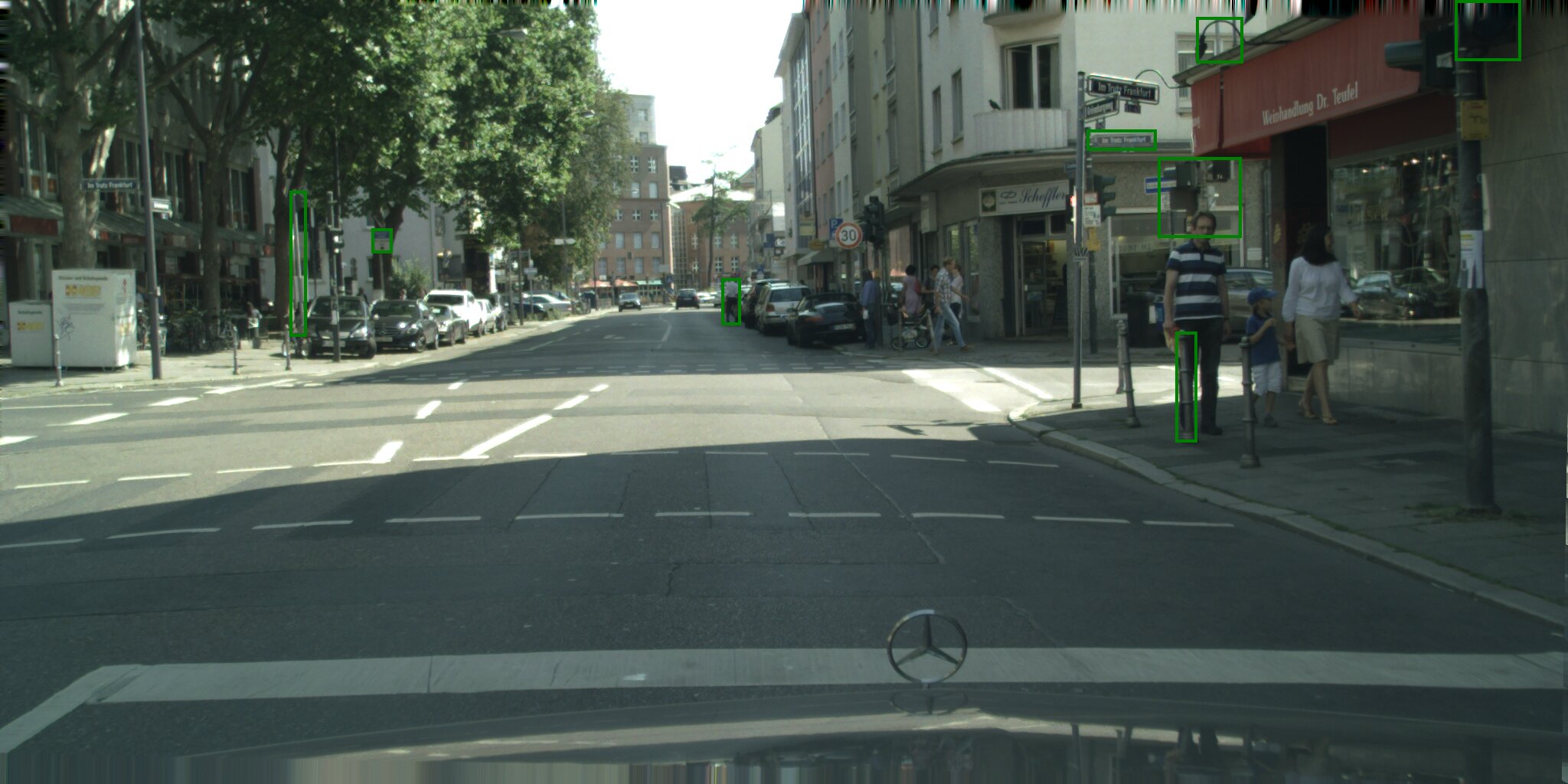}%
\includegraphics[width=0.24\linewidth,clip,trim=1000 470 40 0]{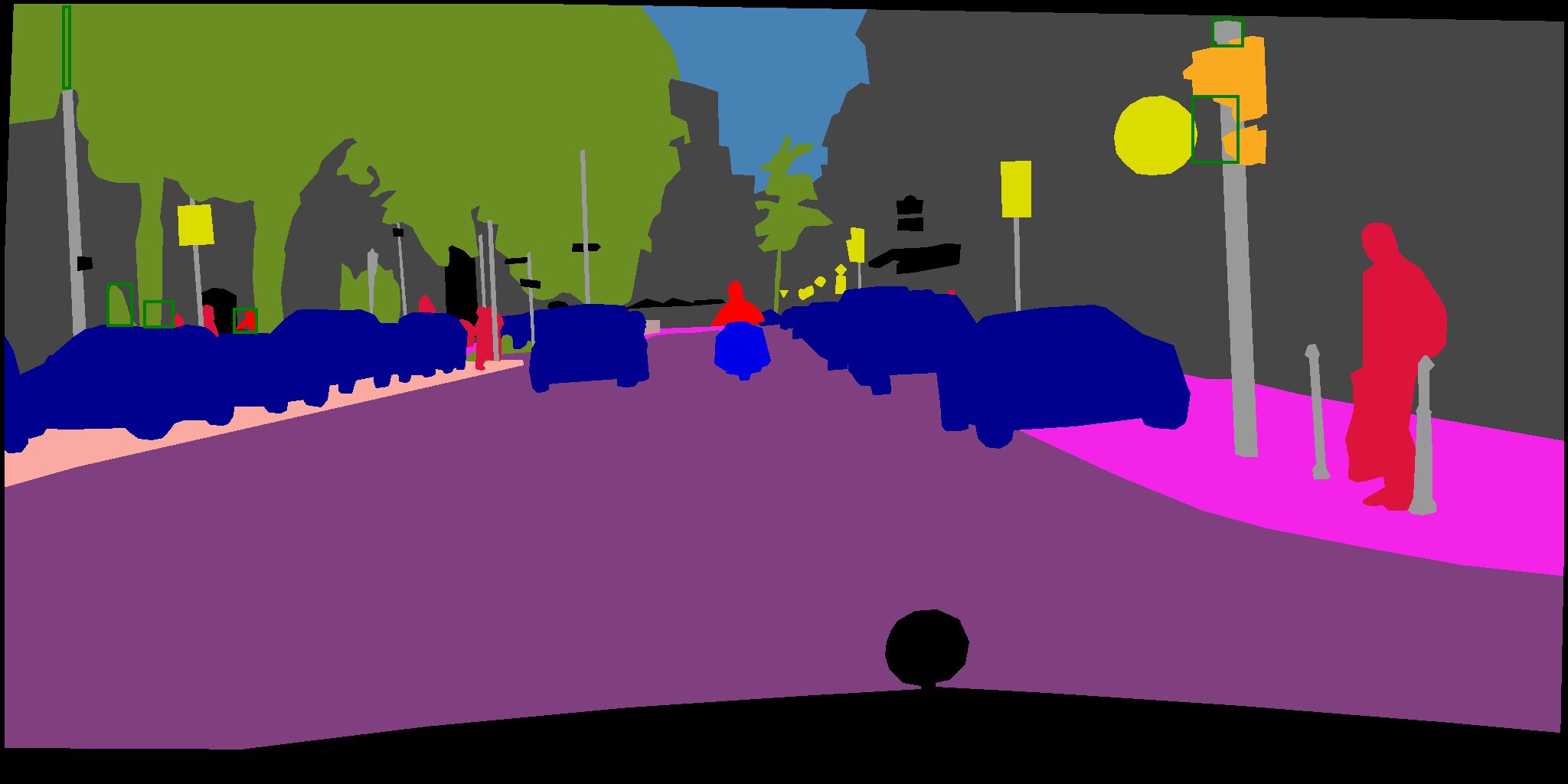}%
\includegraphics[width=0.24\linewidth,clip,trim=1000 470 40 0]{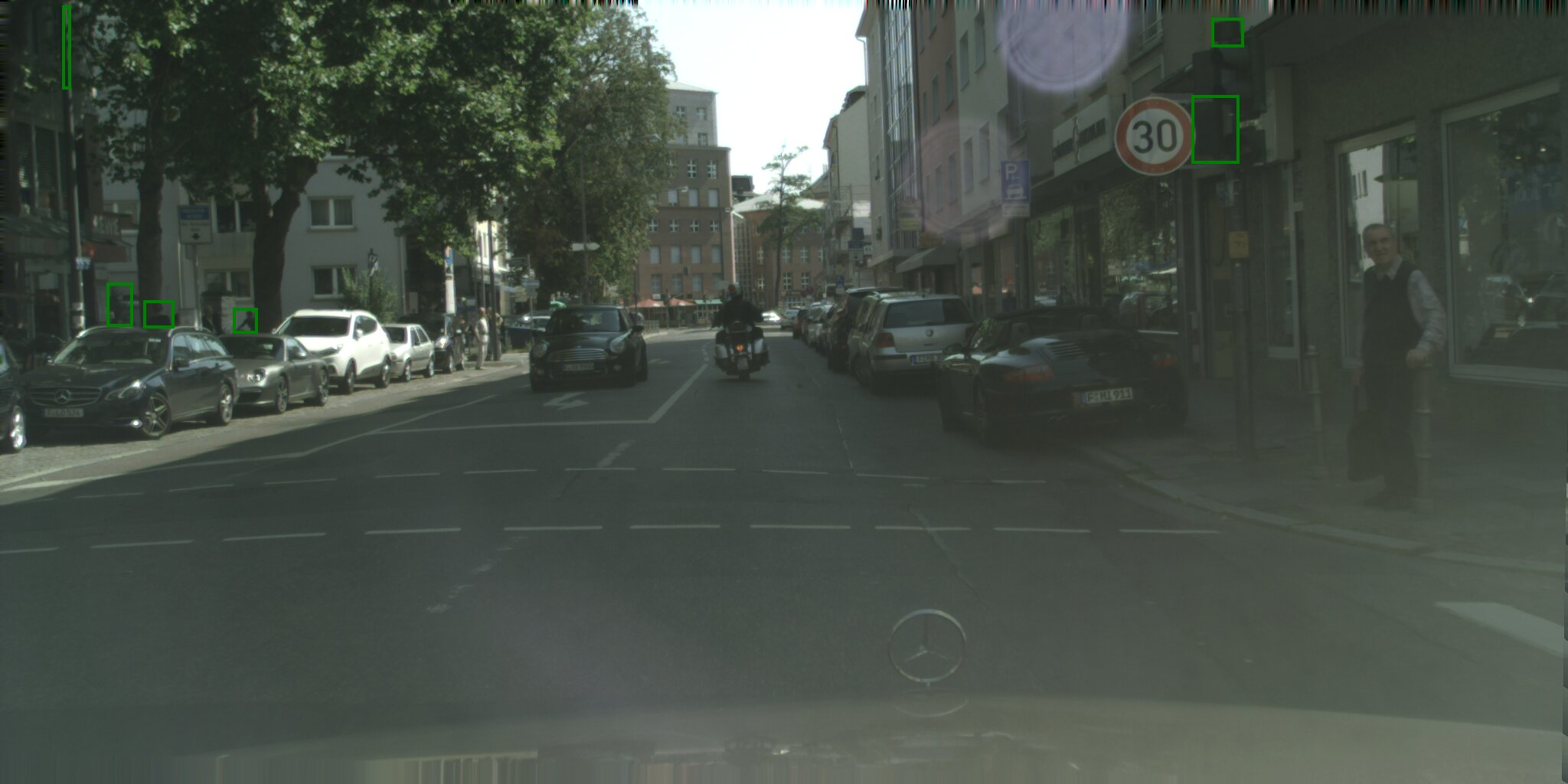}
\hspace{\imgsep}%
\includegraphics[width=0.24\linewidth,clip,trim=800 450 200 115]{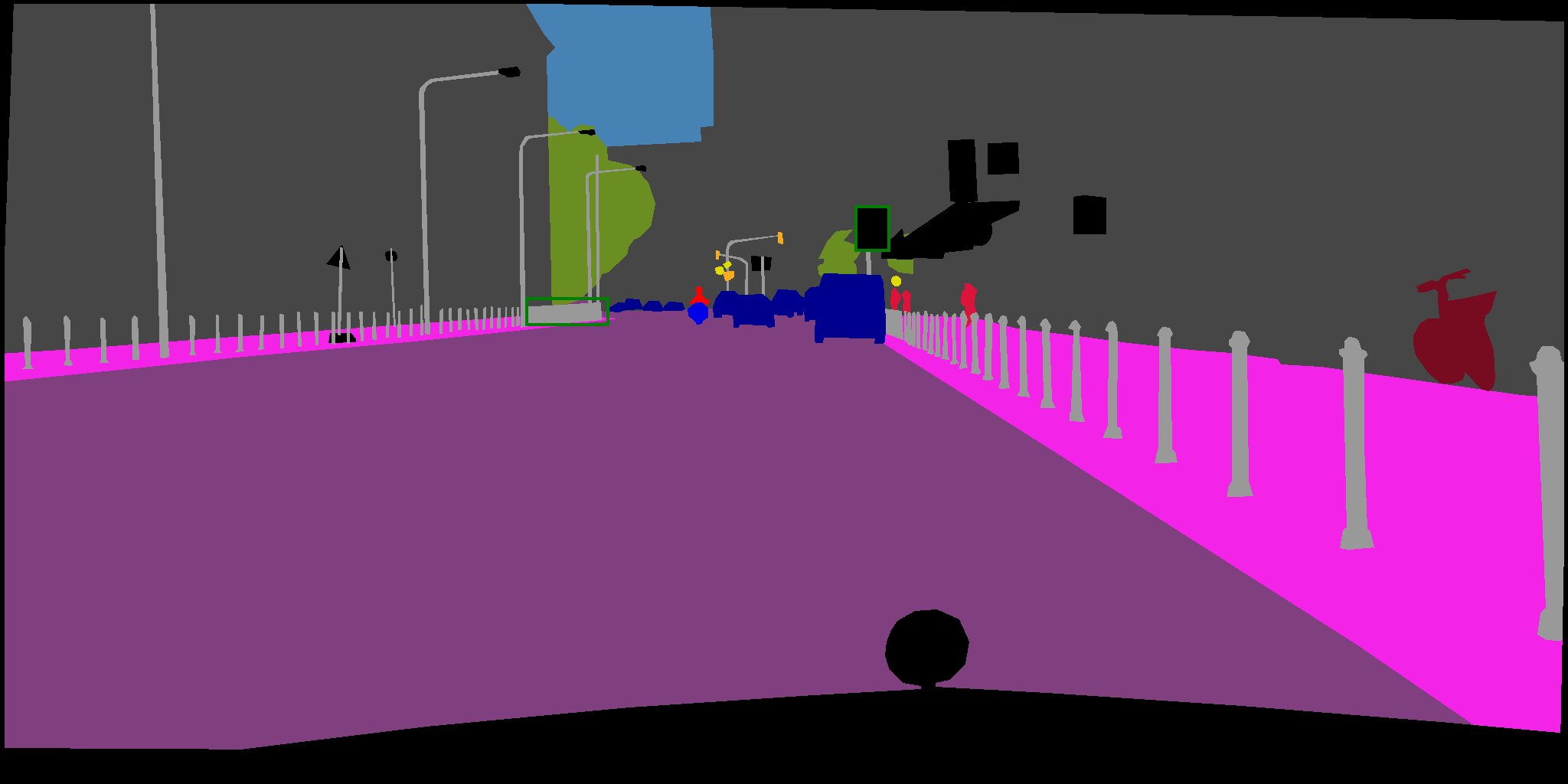}%
\includegraphics[width=0.24\linewidth,clip,trim=800 450 200 115]{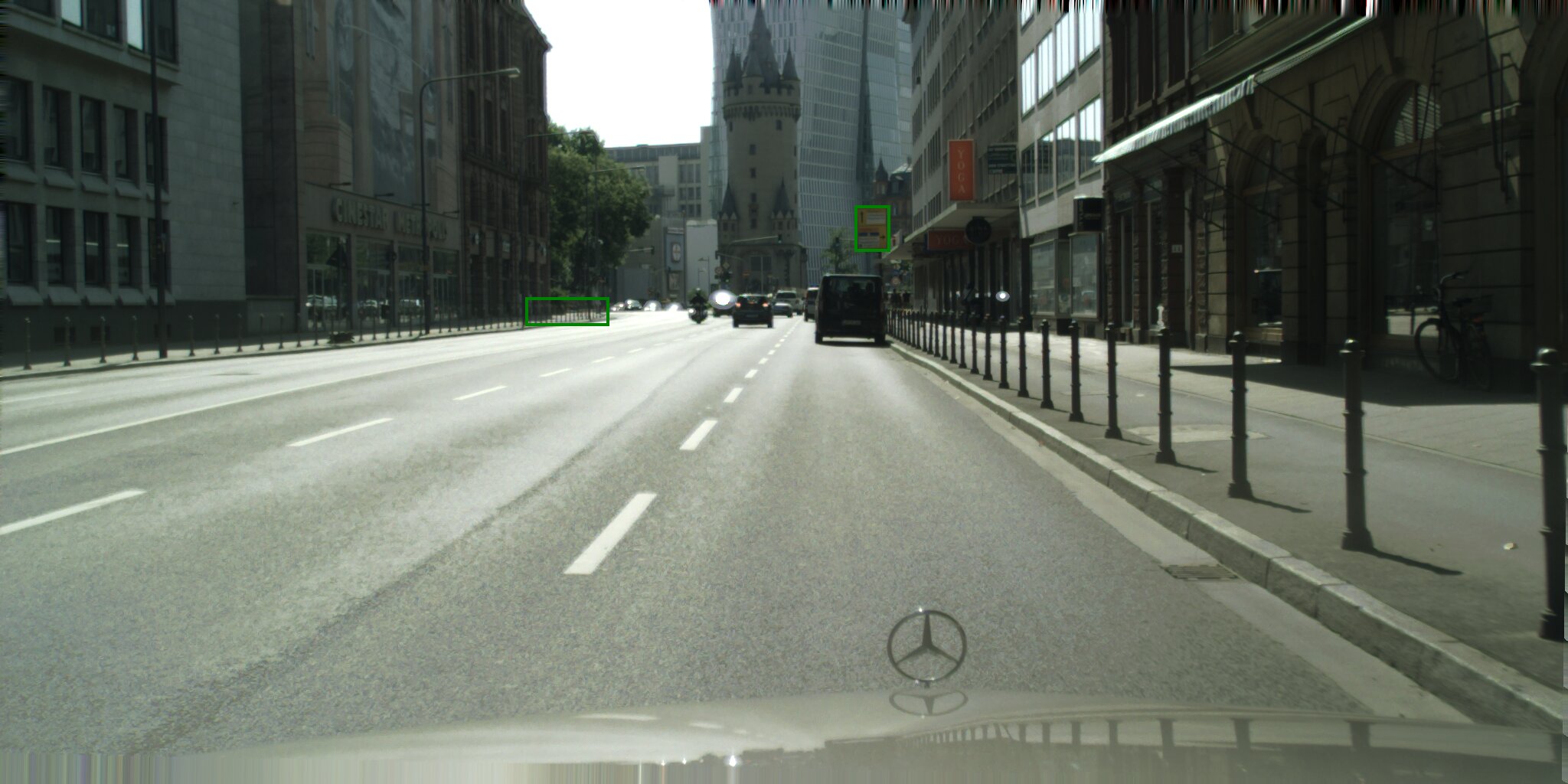}%
\includegraphics[width=0.24\linewidth,clip,trim=0 580 1030 0]{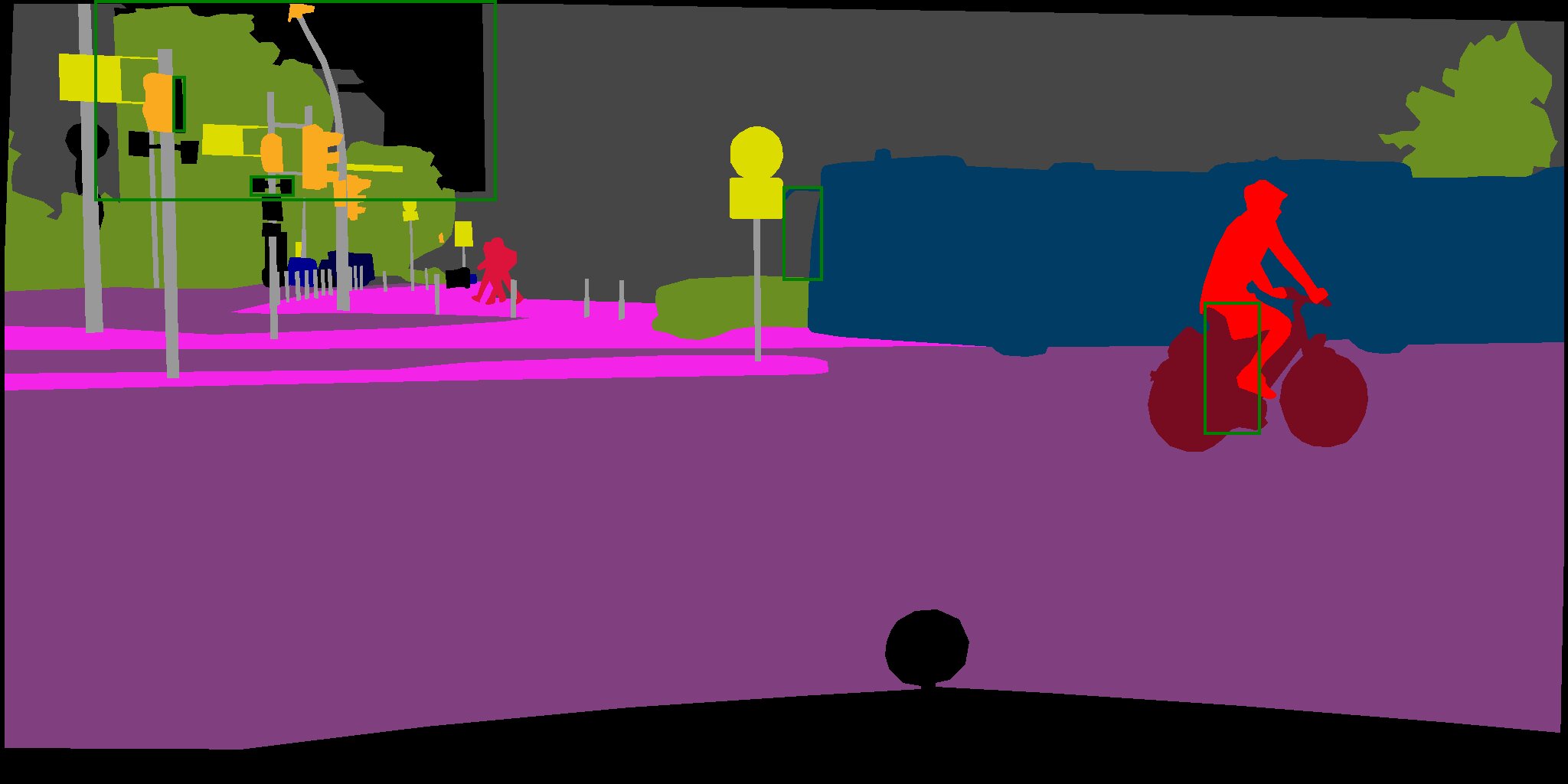}%
\includegraphics[width=0.24\linewidth,clip,trim=0 580 1030 0]{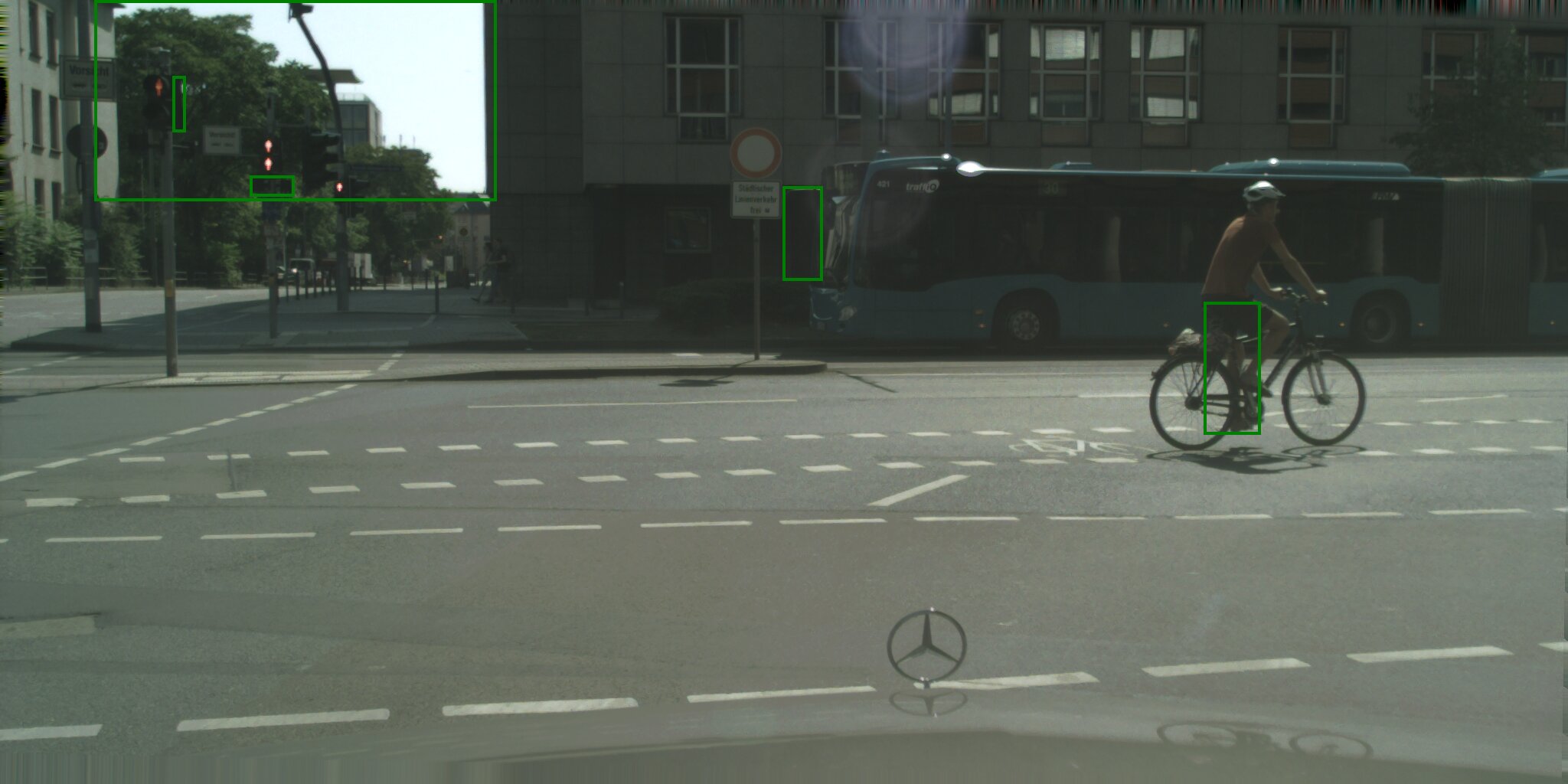}
\hspace{\imgsep}%
\includegraphics[width=0.24\linewidth,clip,trim=570 620 550 0]{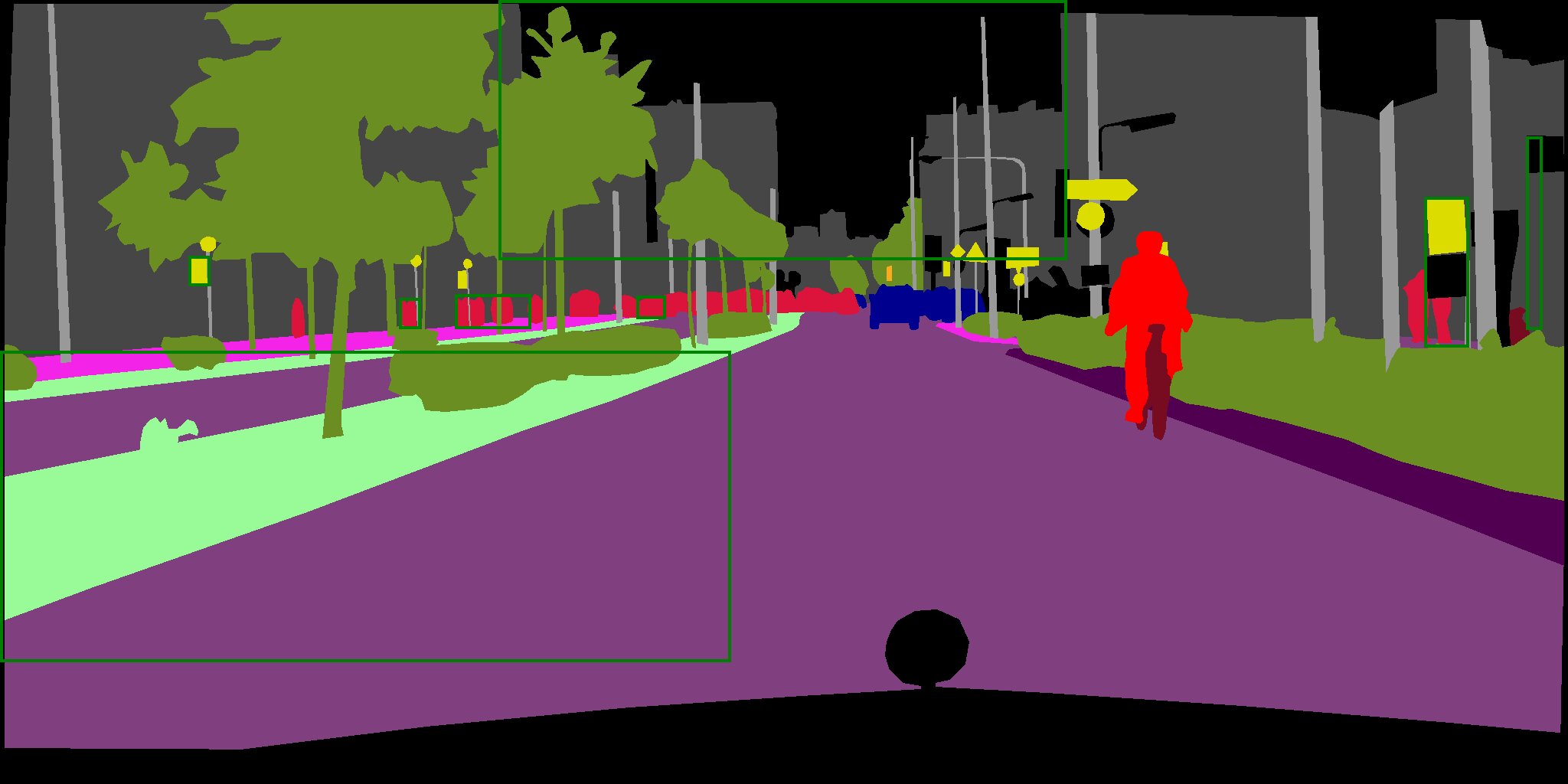}%
\includegraphics[width=0.24\linewidth,clip,trim=570 620 550 0]{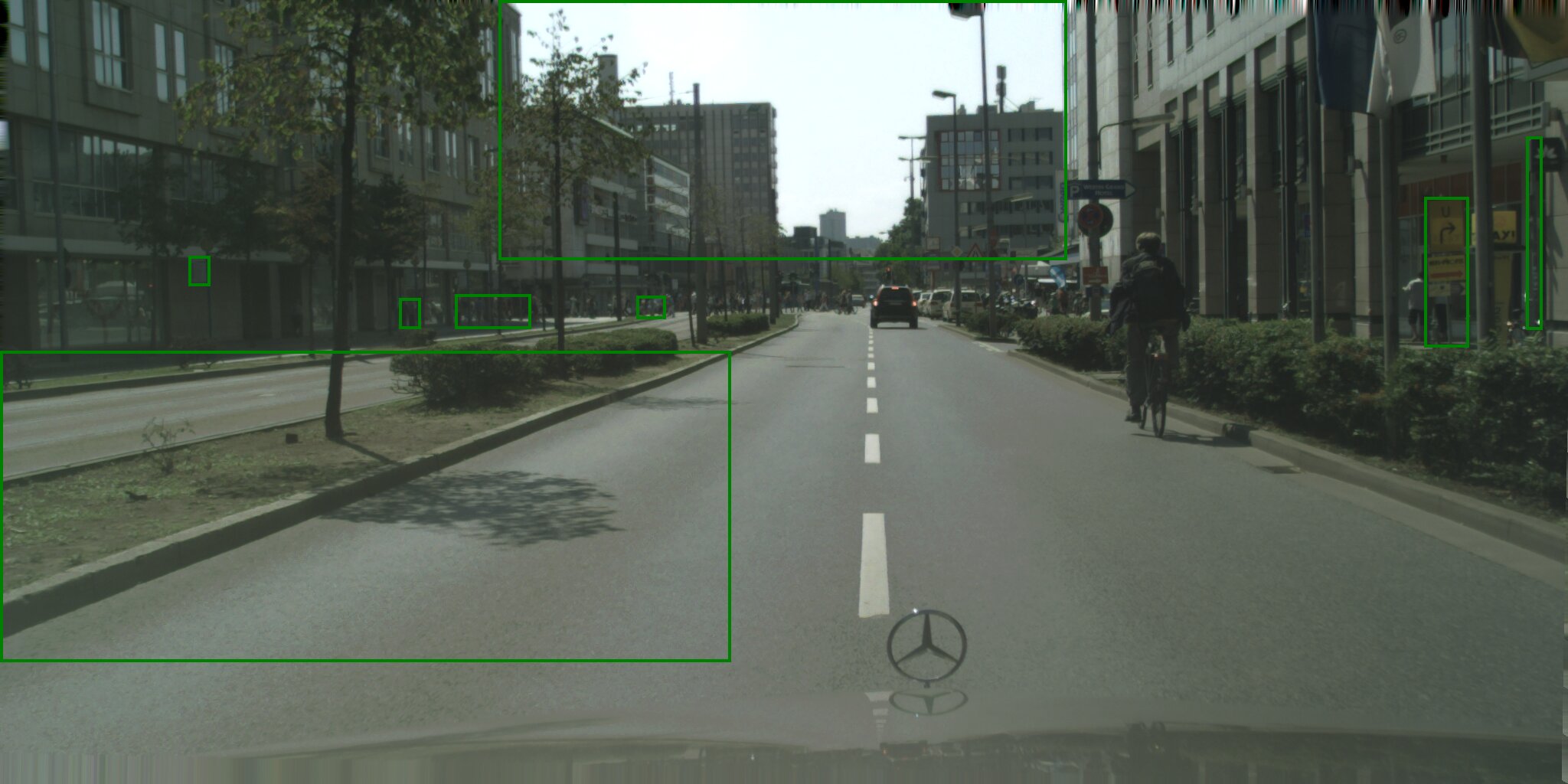}%
\includegraphics[width=0.24\linewidth,clip,trim=600 550 1100 320]{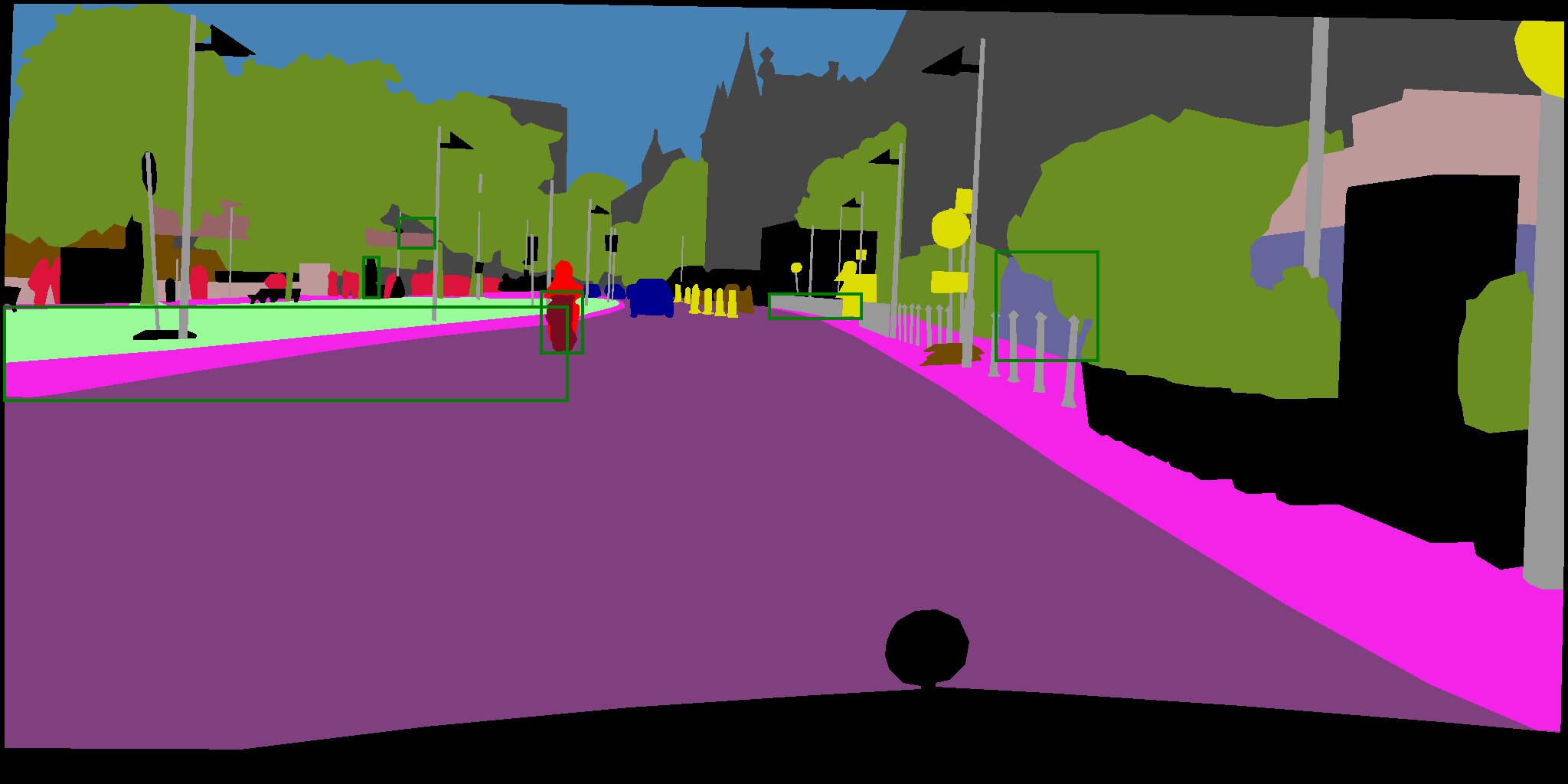}%
\includegraphics[width=0.24\linewidth,clip,trim=600 550 1100 320]{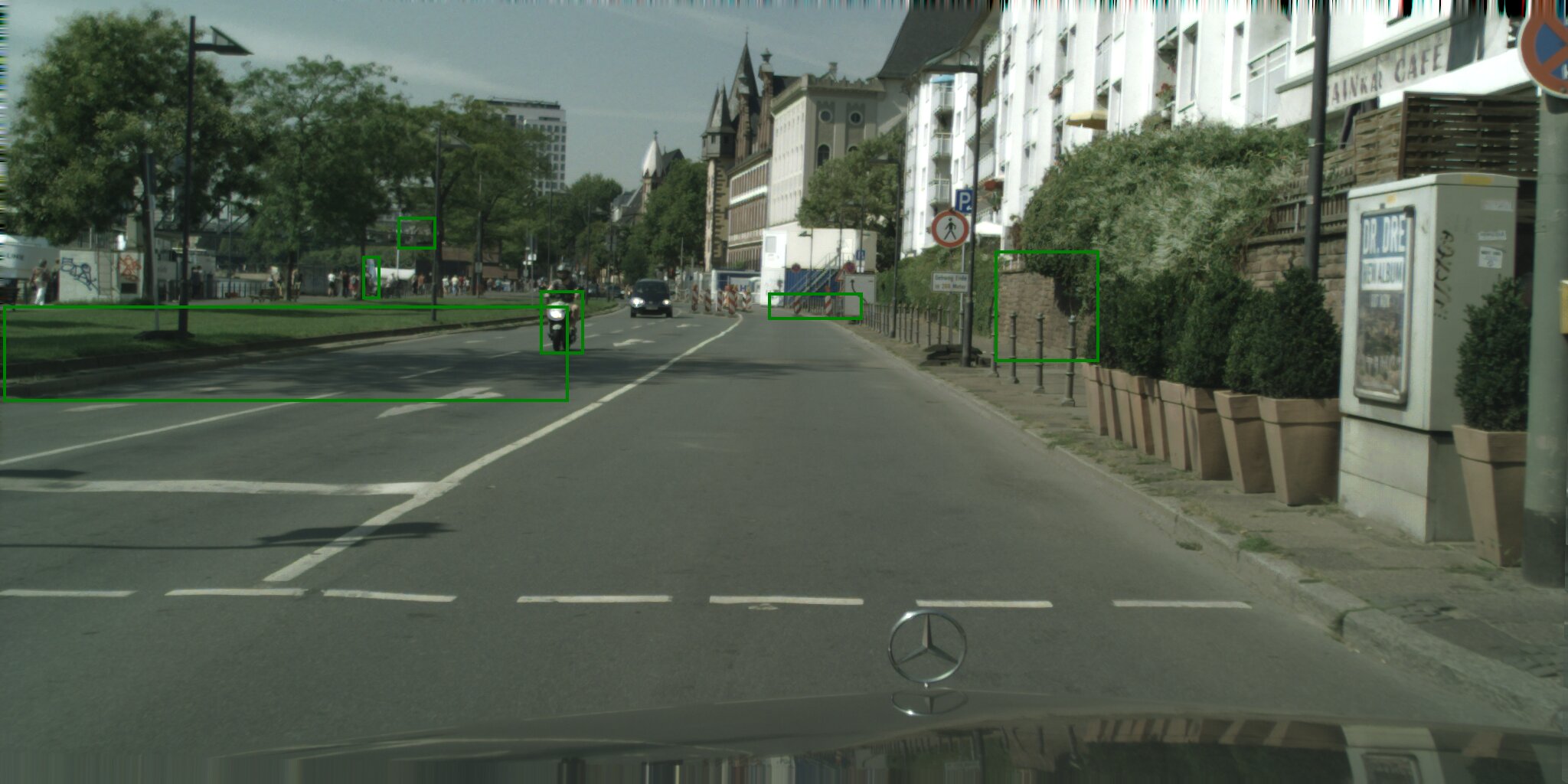}
\hspace{\imgsep}%
\includegraphics[width=0.24\linewidth,clip,trim=0 400 1220 100]{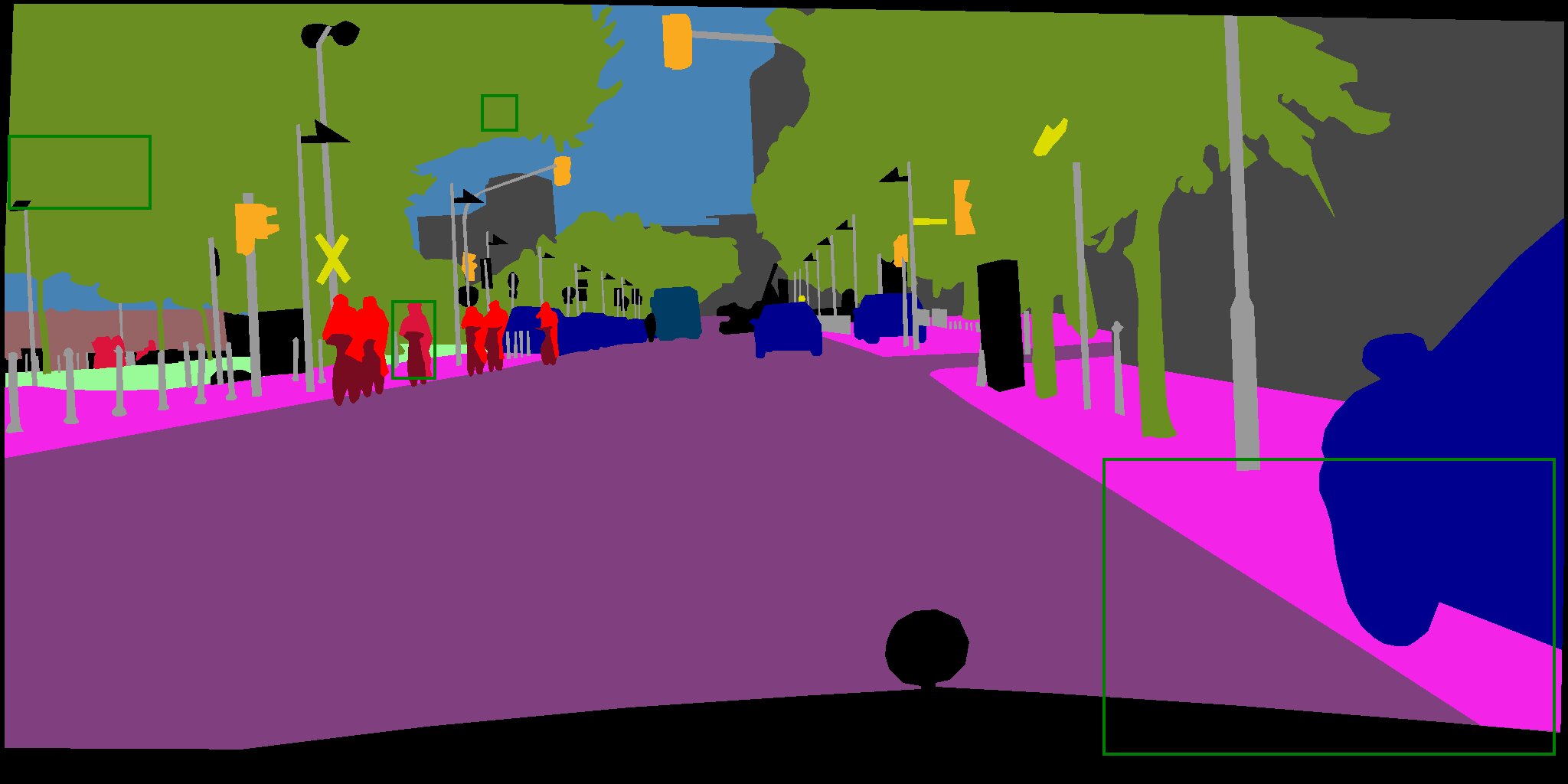}%
\includegraphics[width=0.24\linewidth,clip,trim=0 400 1220 100]{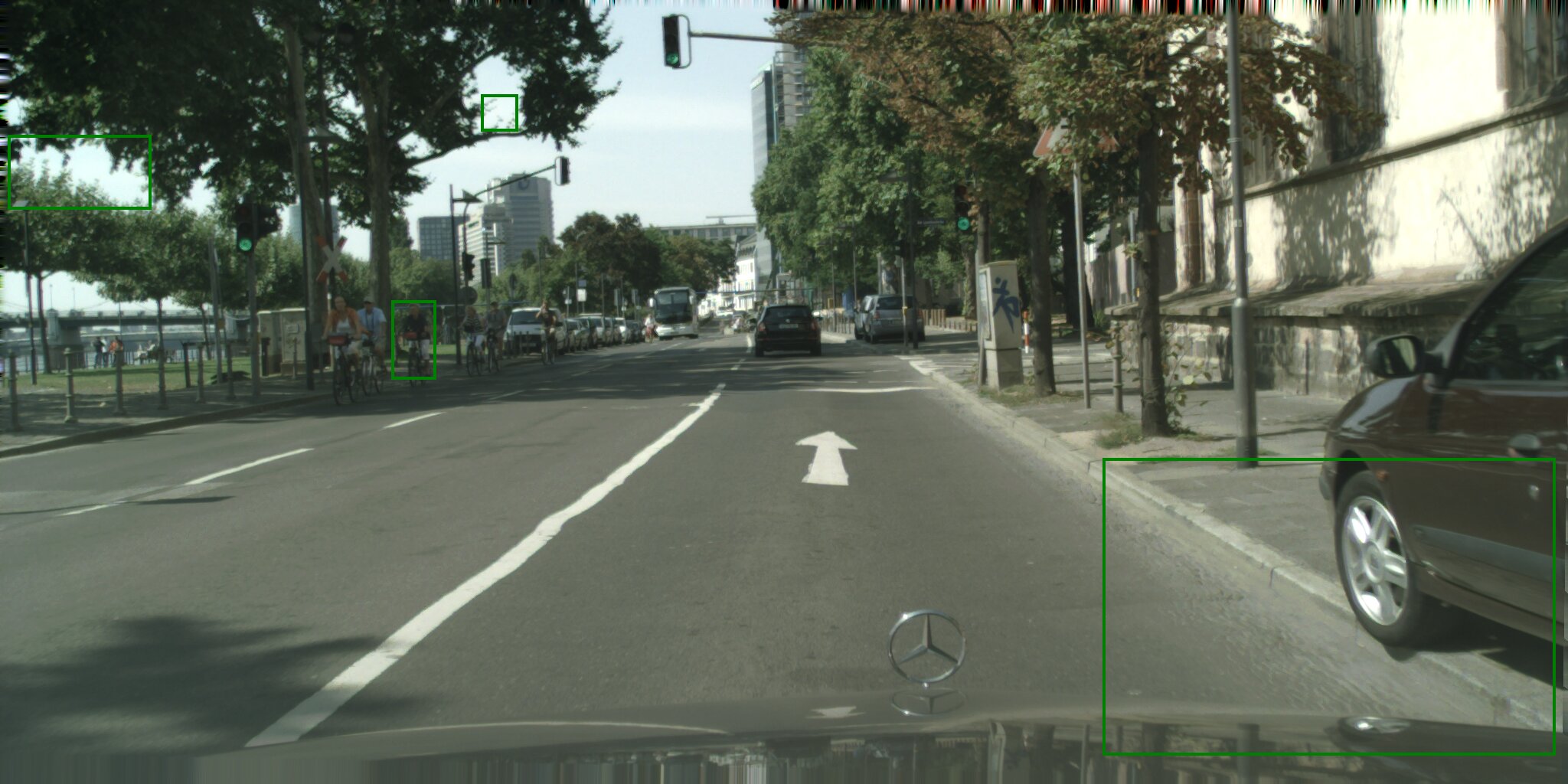}%
\includegraphics[width=0.24\linewidth,clip,trim=900 630 1000 300]{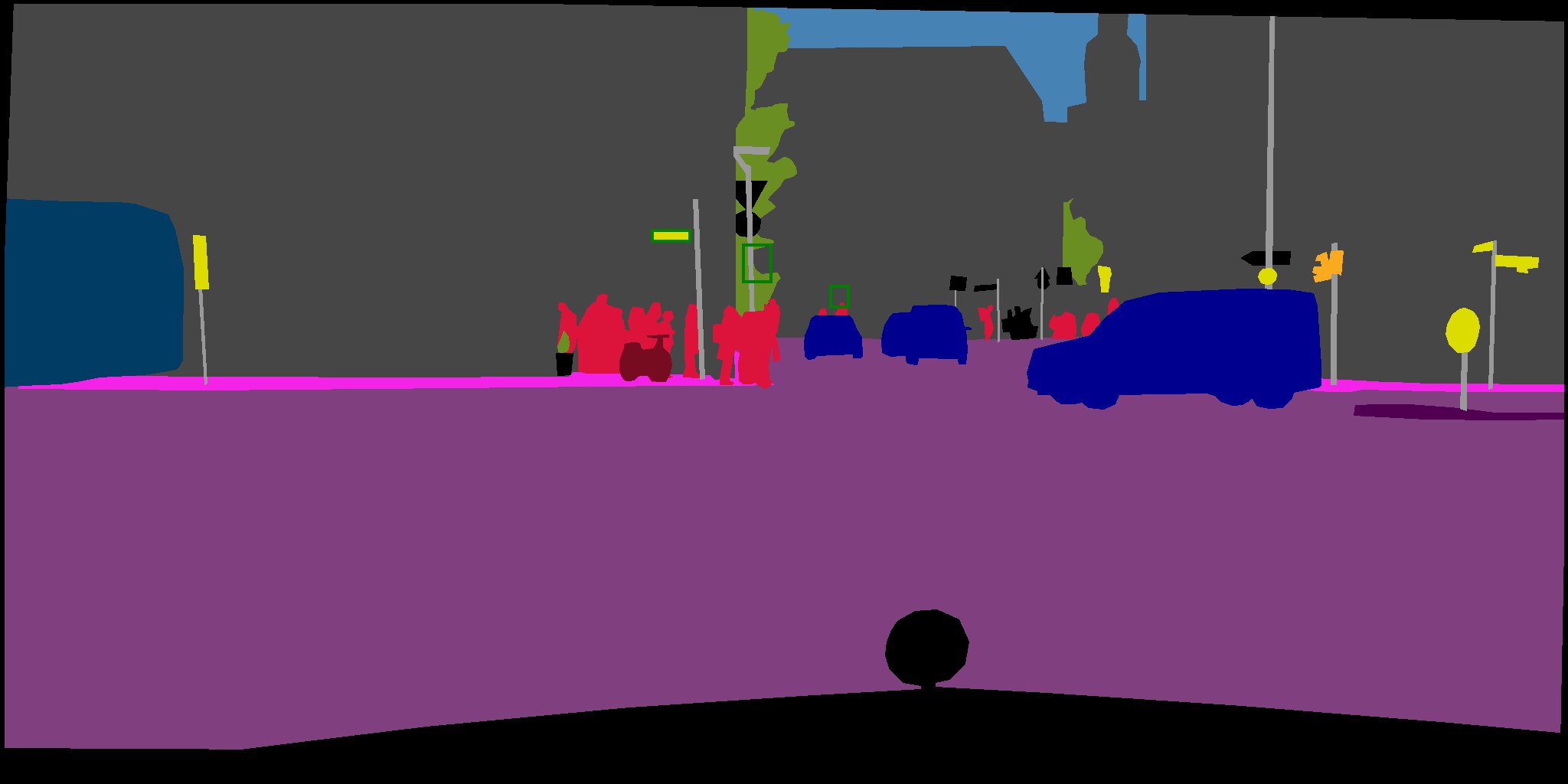}%
\includegraphics[width=0.24\linewidth,clip,trim=900 630 1000 300]{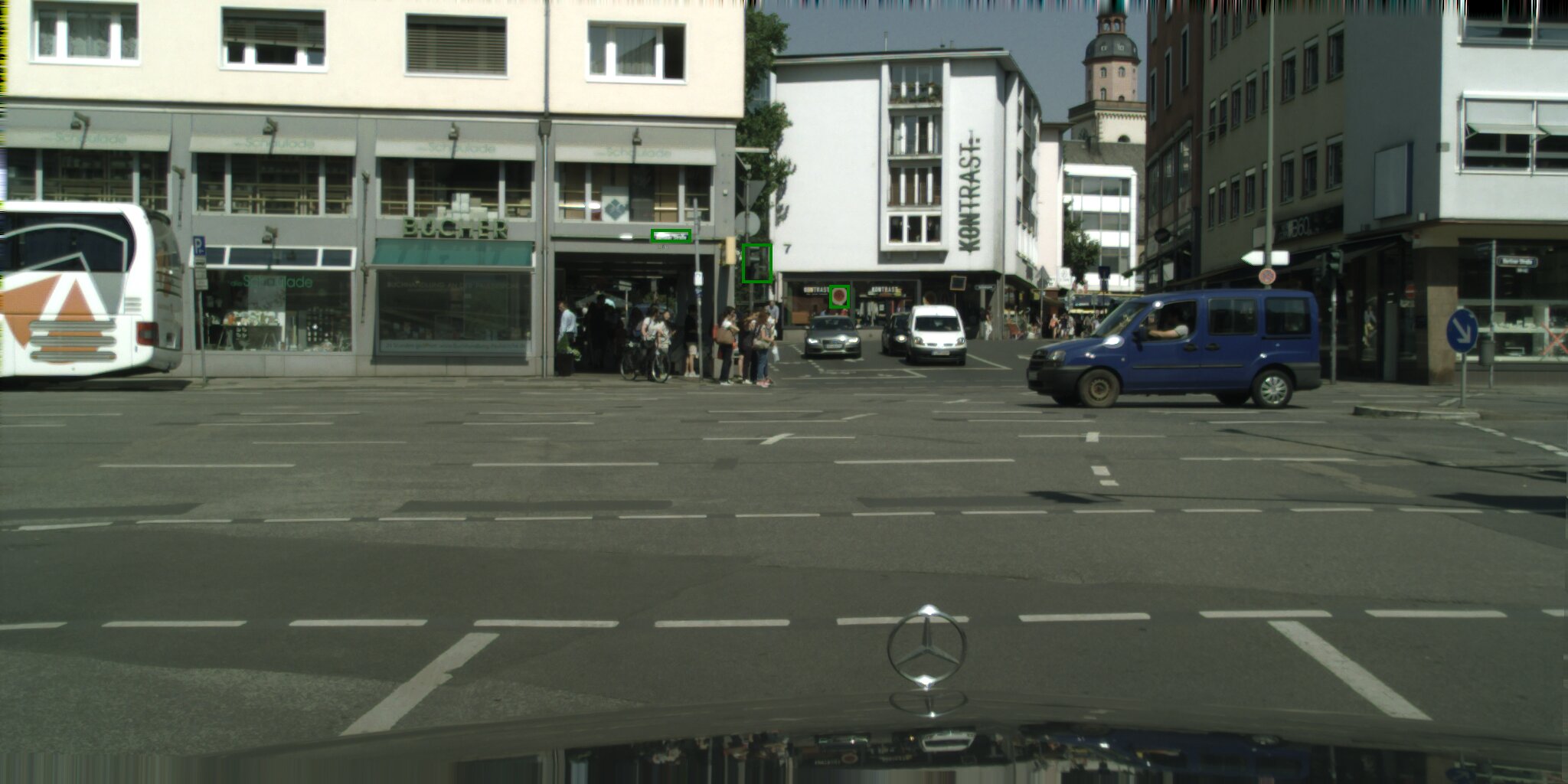}

\caption{A collection of label errors present in Cityscapes. The segmentation masks show the ground truth annotations and the green boxes are the predictions of our label error detection method.}
\label{fig:real_city}
\end{figure}

\clearpage
\begin{figure}
\centering
\newcommand{\imgsep}{1cm} 
\includegraphics[width=0.165\linewidth,clip,trim=300 210 100 0]{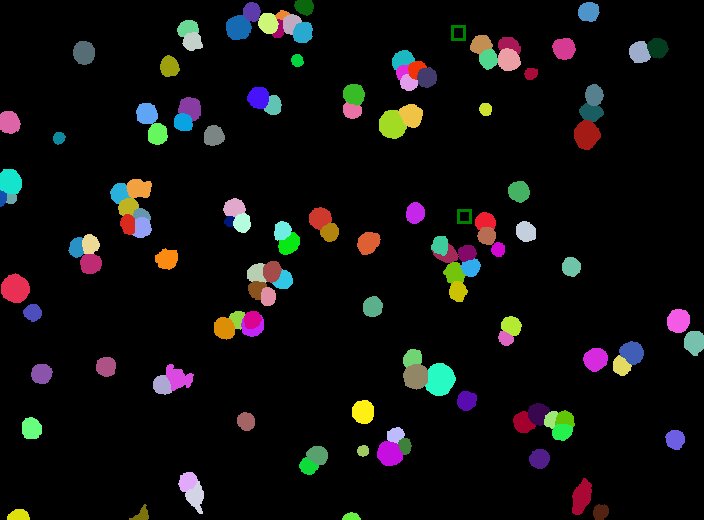}%
\includegraphics[width=0.165\linewidth,clip,trim=300 210 100 0]{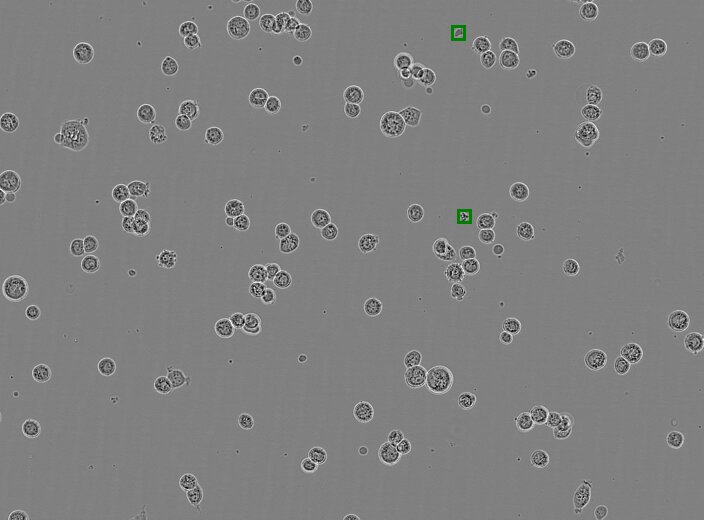}%
\includegraphics[width=0.165\linewidth,clip,trim=30 20 180 0]{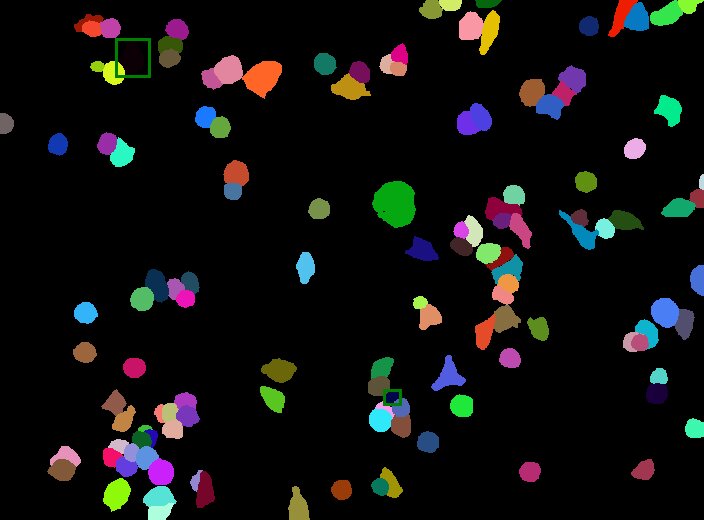}%
\includegraphics[width=0.165\linewidth,clip,trim=30 20 180 0]{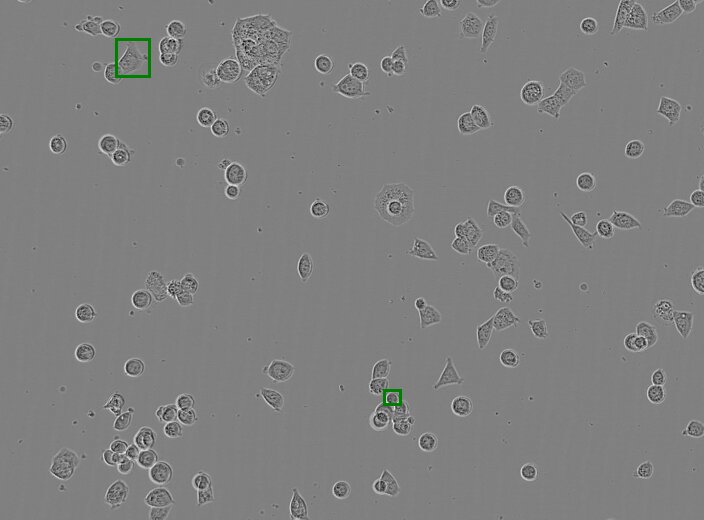}%
\includegraphics[width=0.165\linewidth,clip,trim=0 0 260 70]{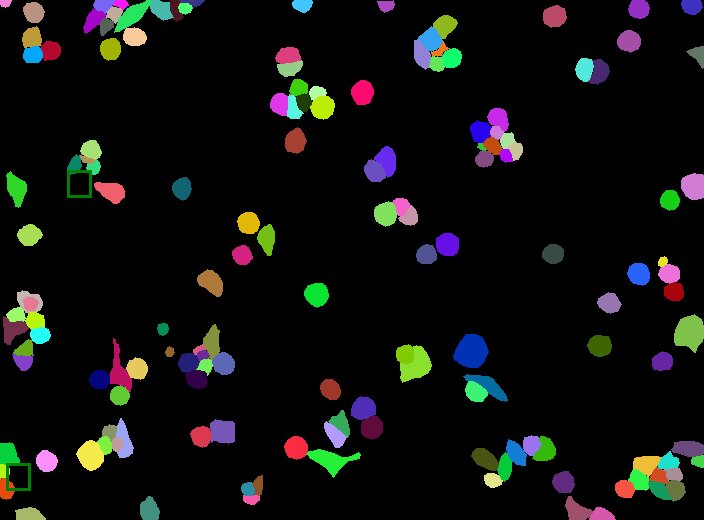}%
\includegraphics[width=0.165\linewidth,clip,trim=0 0 260 70]{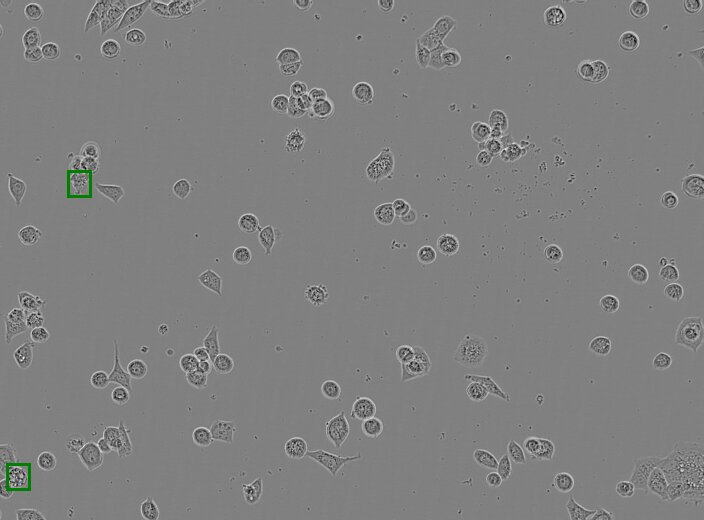}%
\hspace{\imgsep}%
\includegraphics[width=0.165\linewidth,clip,trim=100 140 200 0]{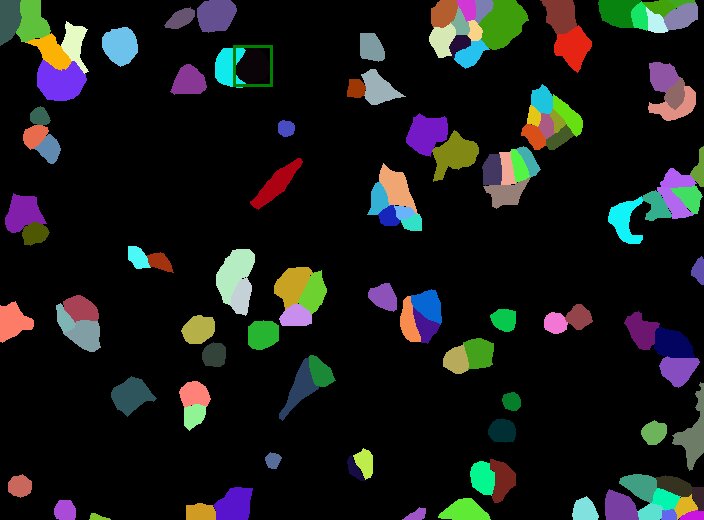}%
\includegraphics[width=0.165\linewidth,clip,trim=100 140 200 0]{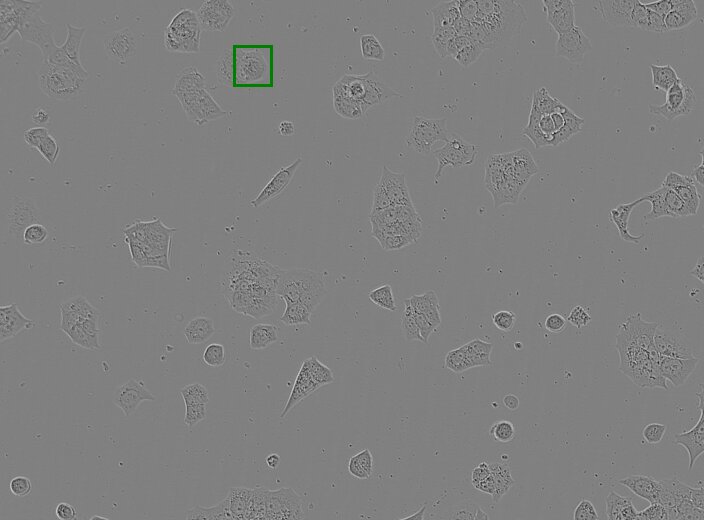}%
\includegraphics[width=0.165\linewidth,clip,trim=100 0 140 90]{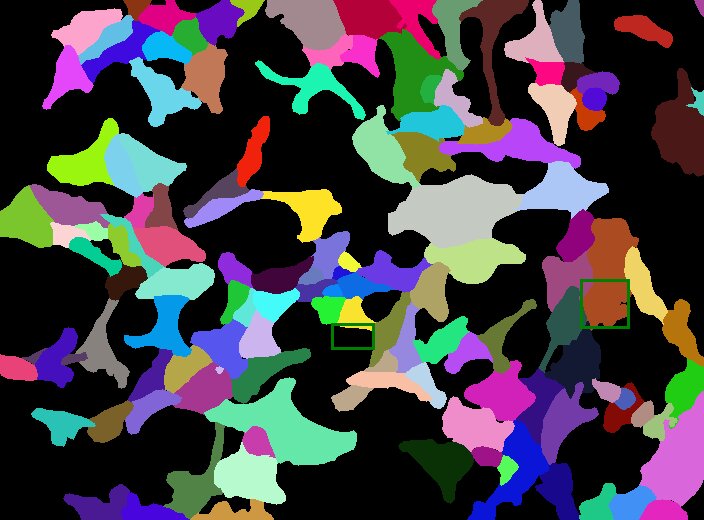}%
\includegraphics[width=0.165\linewidth,clip,trim=100 0 140 90]{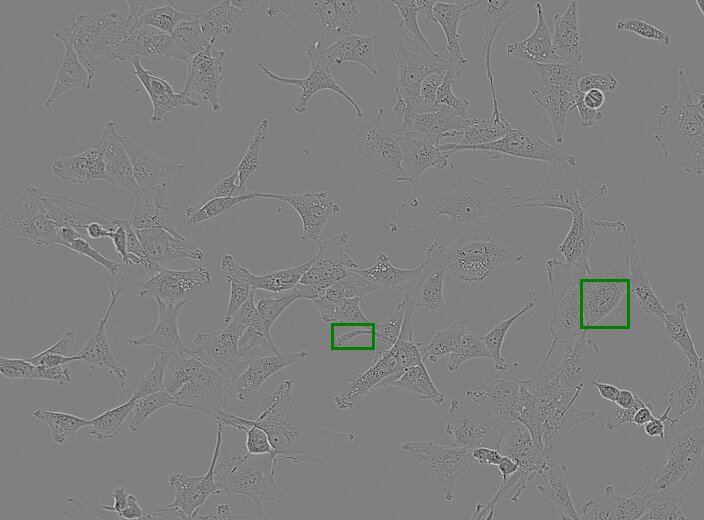}%
\includegraphics[width=0.165\linewidth,clip,trim=180 230 210 0]{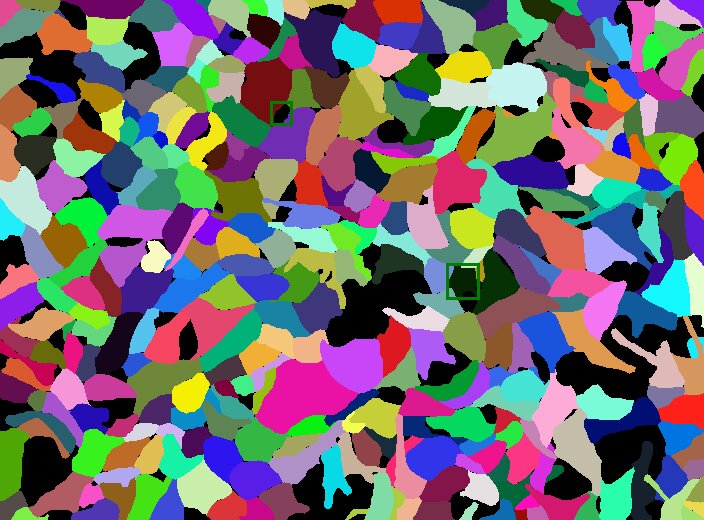}%
\includegraphics[width=0.165\linewidth,clip,trim=180 230 210 0]{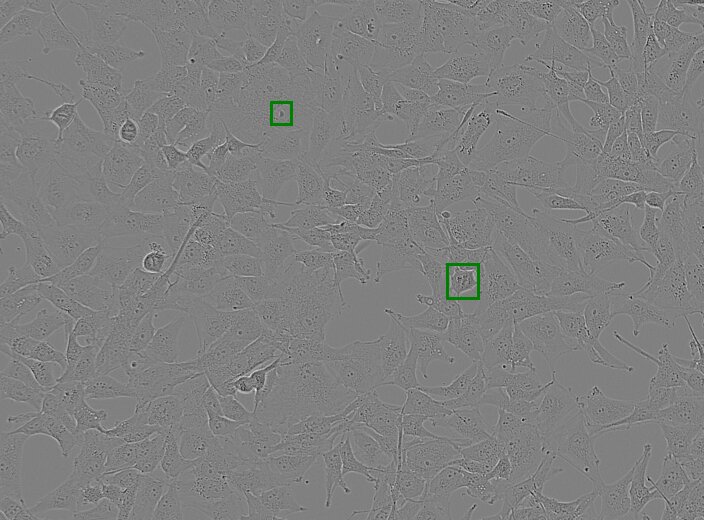}
\hspace{\imgsep}%
\includegraphics[width=0.165\linewidth,clip,trim=300 0 0 170]{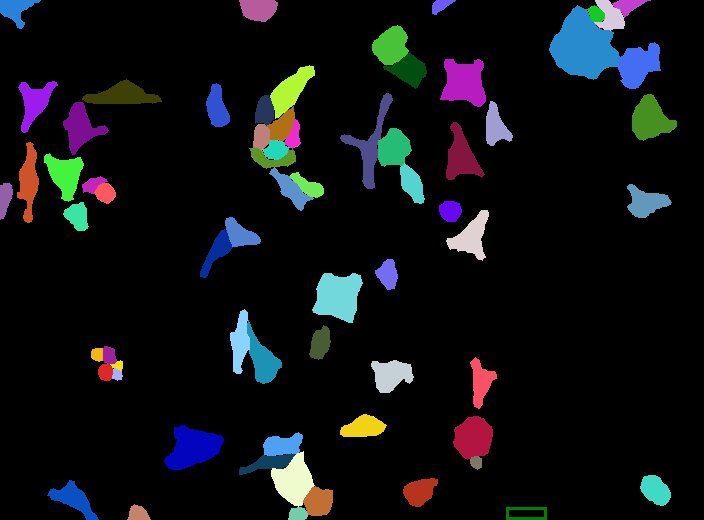}%
\includegraphics[width=0.165\linewidth,clip,trim=300 0 0 170]{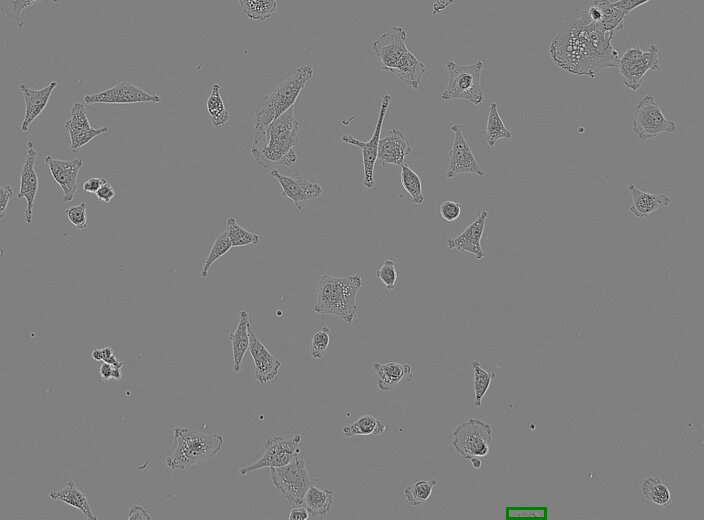}%
\includegraphics[width=0.165\linewidth,clip,trim=300 0 0 170]{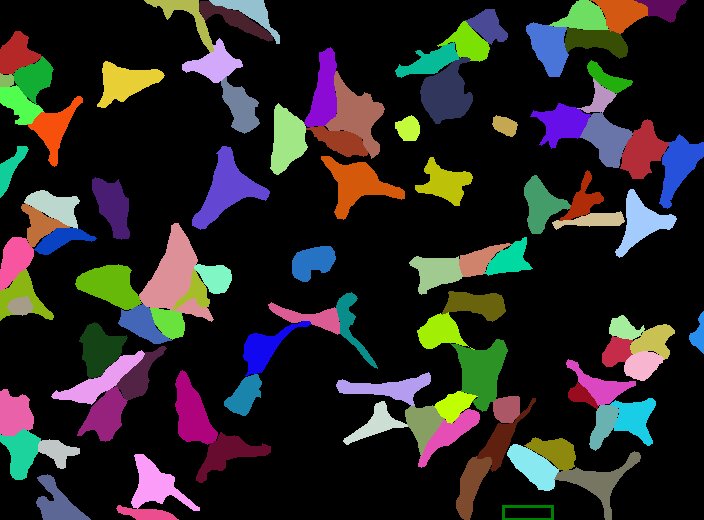}%
\includegraphics[width=0.165\linewidth,clip,trim=300 0 0 170]{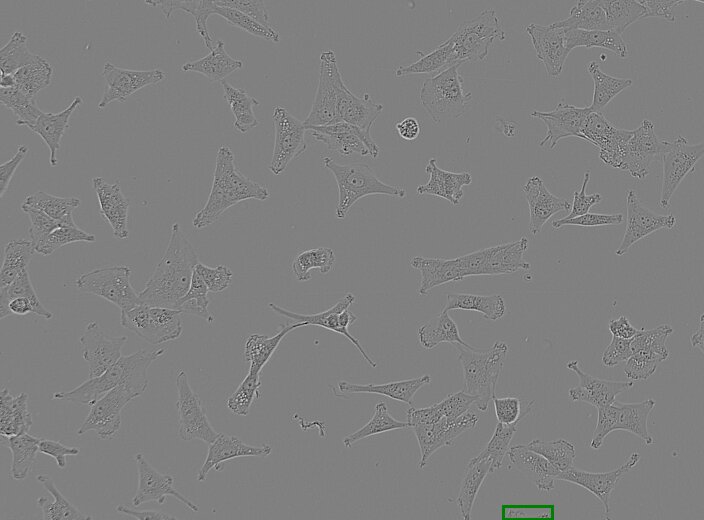}%
\includegraphics[width=0.165\linewidth,clip,trim=0 170 300 0]{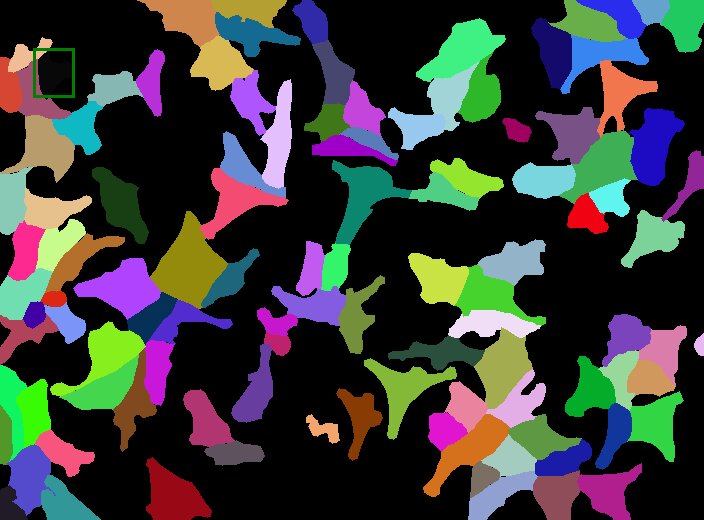}%
\includegraphics[width=0.165\linewidth,clip,trim=0 170 300 0]{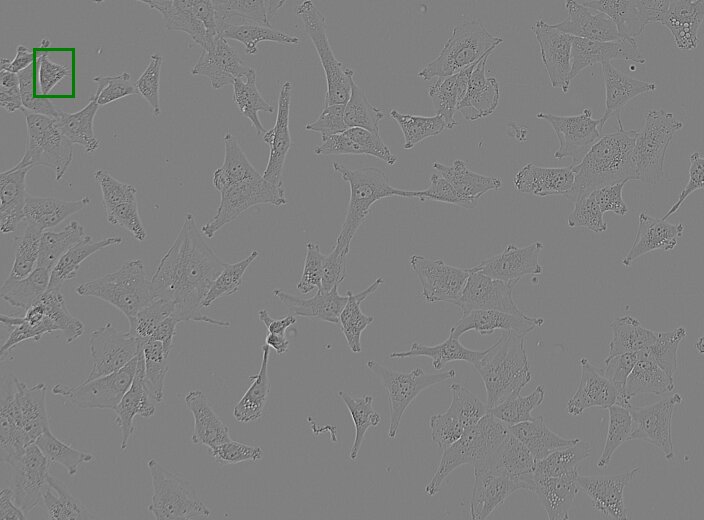}%
\hspace{\imgsep}%
\includegraphics[width=0.165\linewidth,clip,trim=0 20 200 70]{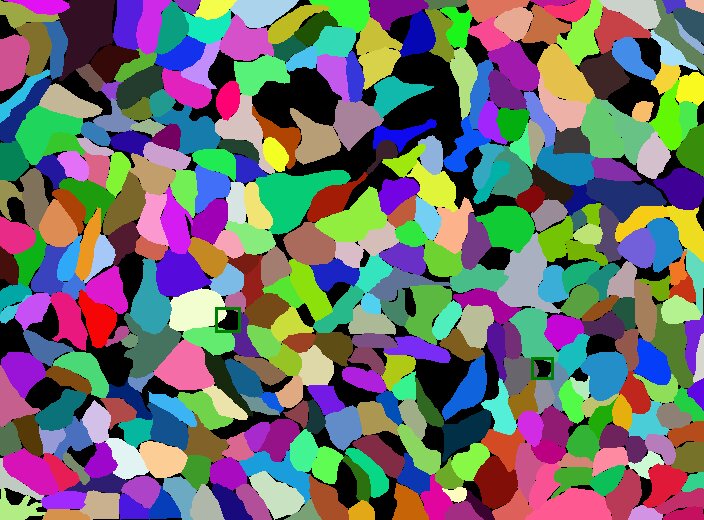}%
\includegraphics[width=0.165\linewidth,clip,trim=0 20 200 70]{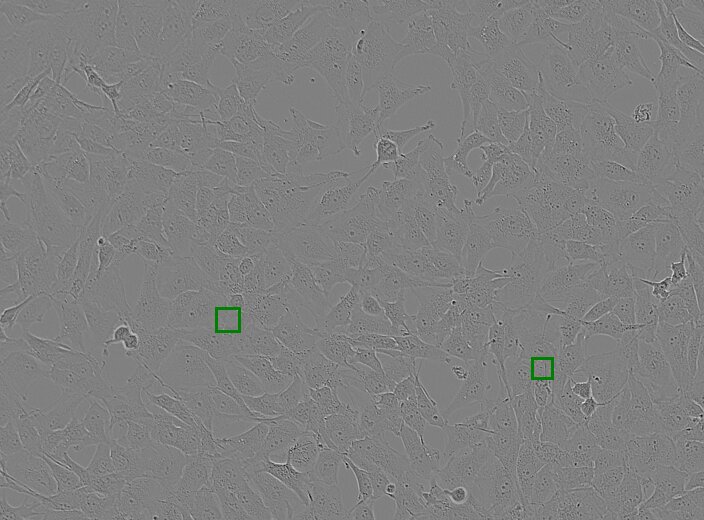}%
\includegraphics[width=0.165\linewidth,clip,trim=0 0 440 300]{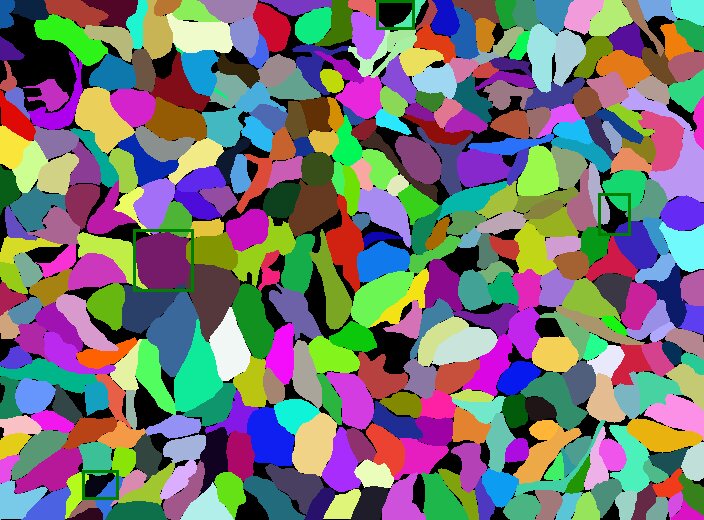}%
\includegraphics[width=0.165\linewidth,clip,trim=0 0 440 300]{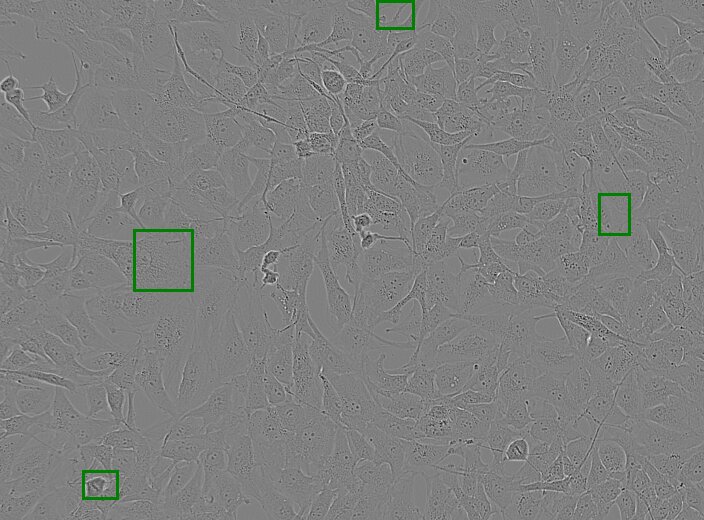}%
\includegraphics[width=0.165\linewidth,clip,trim=200 100 120 100]{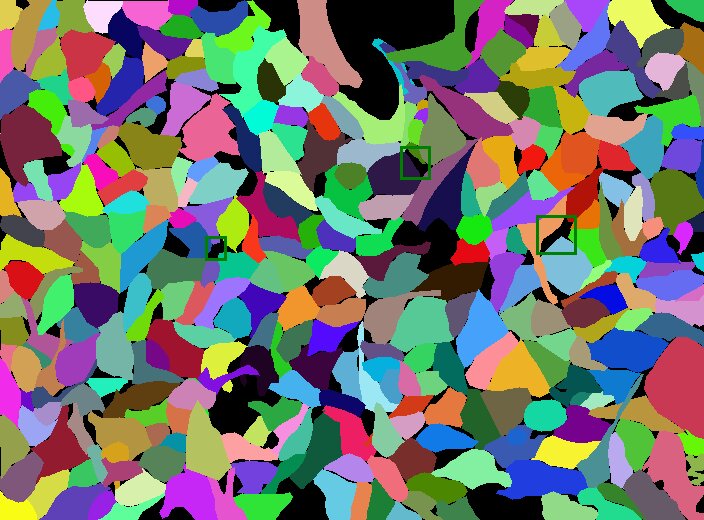}%
\includegraphics[width=0.165\linewidth,clip,trim=200 100 120 100]{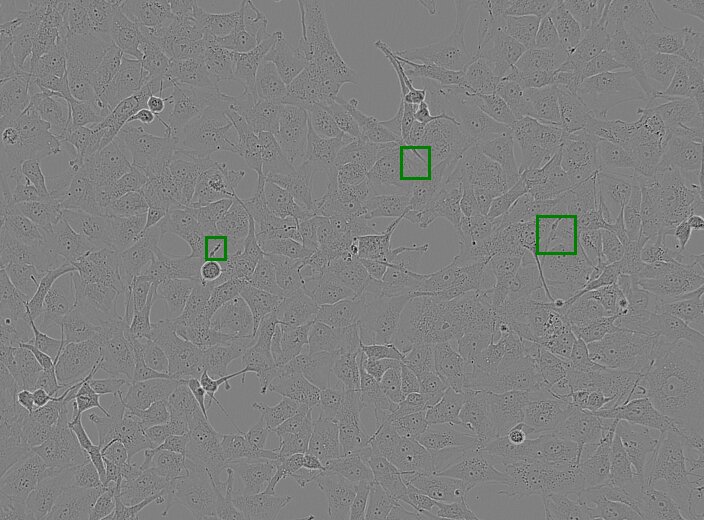}
\hspace{\imgsep}%
\includegraphics[width=0.165\linewidth,clip,trim=300 170 0 0]{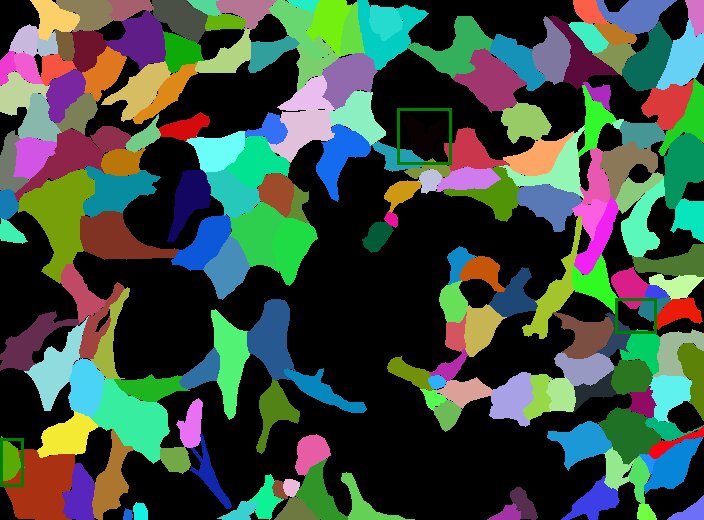}%
\includegraphics[width=0.165\linewidth,clip,trim=300 170 0 0]{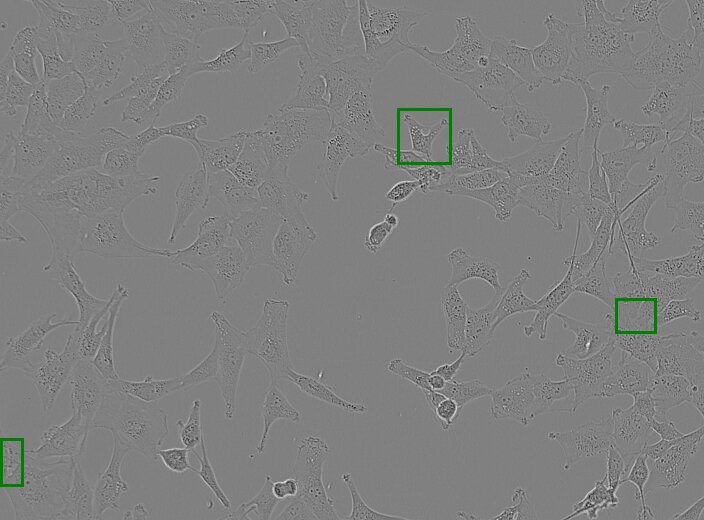}%
\includegraphics[width=0.165\linewidth,clip,trim=0 170 300 0]{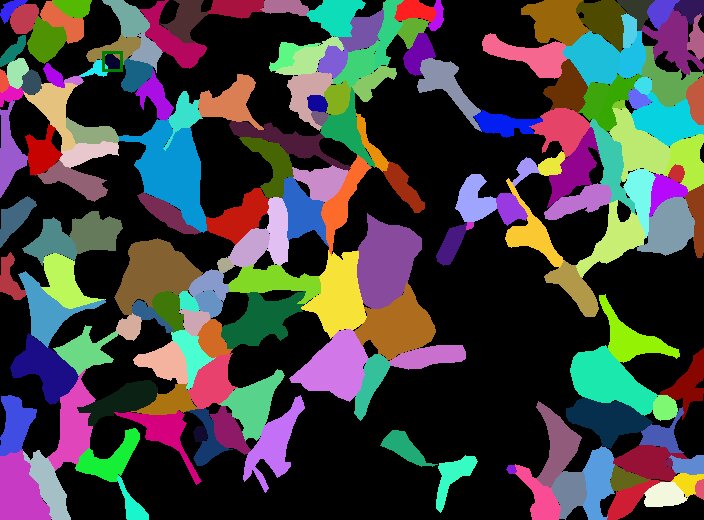}%
\includegraphics[width=0.165\linewidth,clip,trim=0 170 300 0]{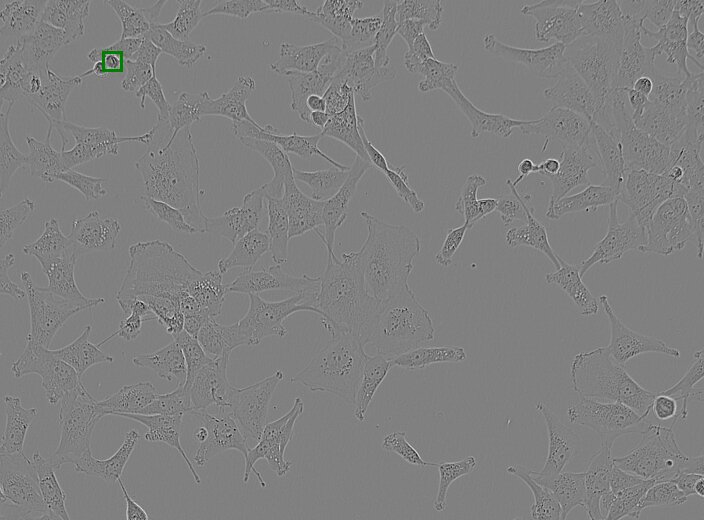}%
\includegraphics[width=0.165\linewidth,clip,trim=100 210 250 0]{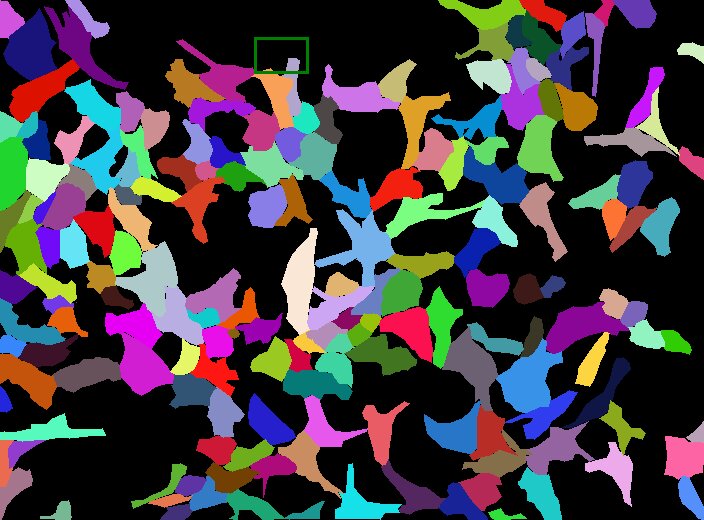}%
\includegraphics[width=0.165\linewidth,clip,trim=100 210 250 0]{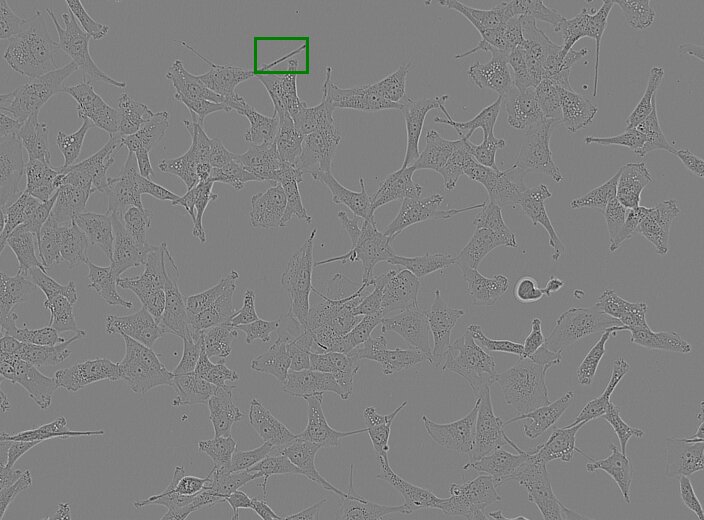}
\hspace{\imgsep}%
\includegraphics[width=0.165\linewidth,clip,trim=380 220 0 0]{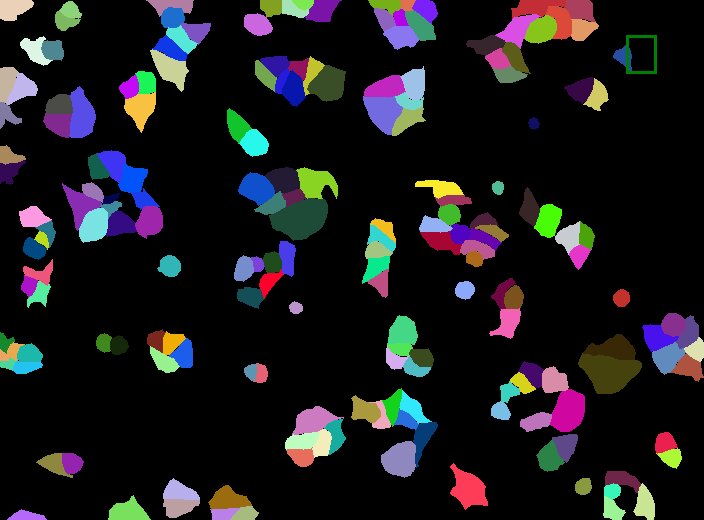}%
\includegraphics[width=0.165\linewidth,clip,trim=380 220 0 0]{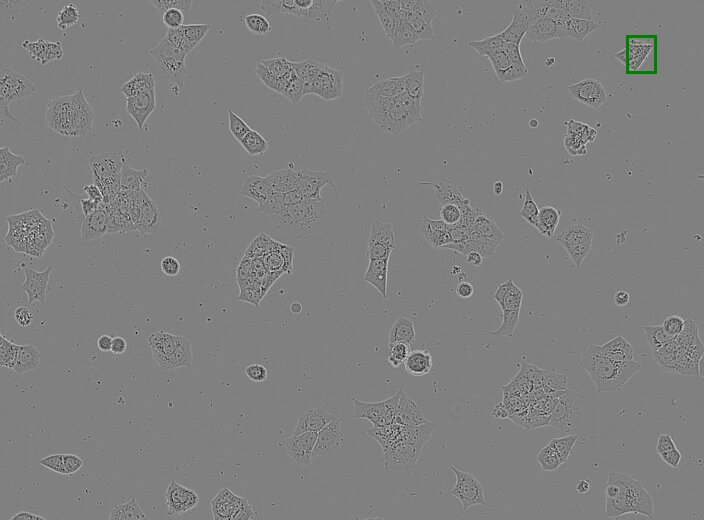}%
\includegraphics[width=0.165\linewidth,clip,trim=300 0 0 150]{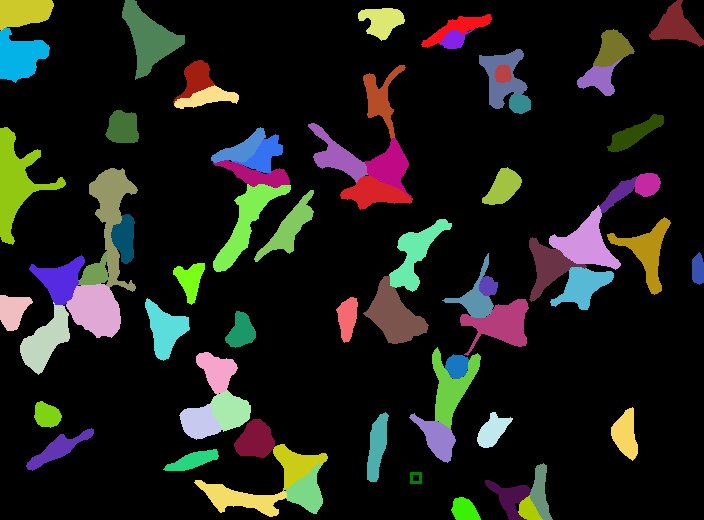}%
\includegraphics[width=0.165\linewidth,clip,trim=300 0 0 150]{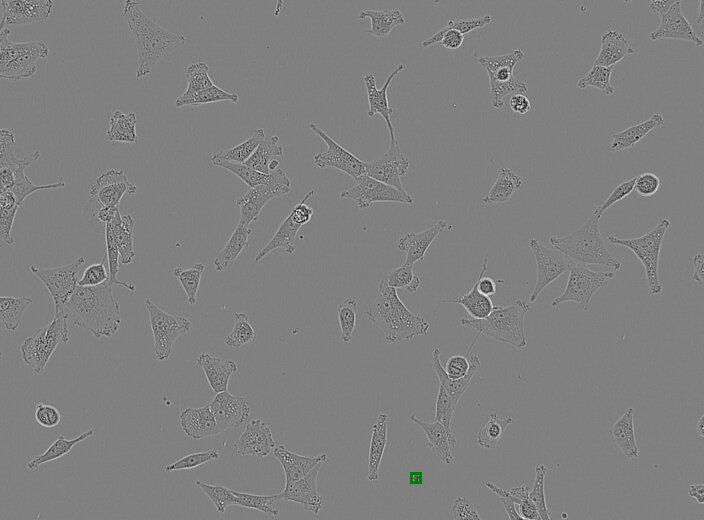}%
\includegraphics[width=0.165\linewidth,clip,trim=280 270 150 0]{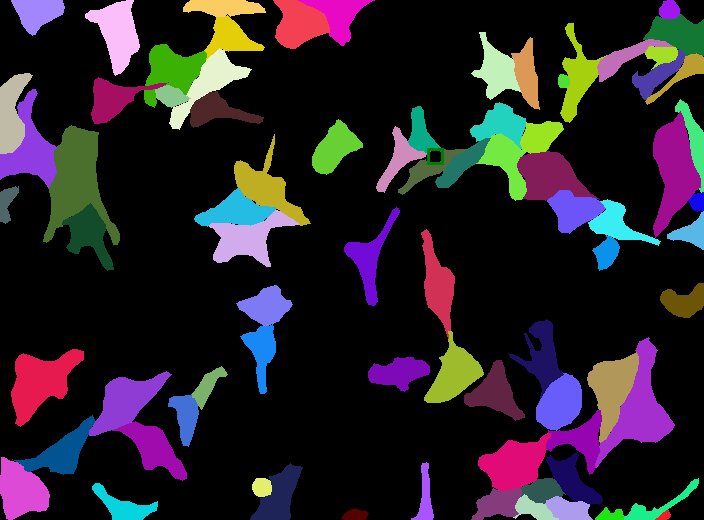}%
\includegraphics[width=0.165\linewidth,clip,trim=280 270 150 0]{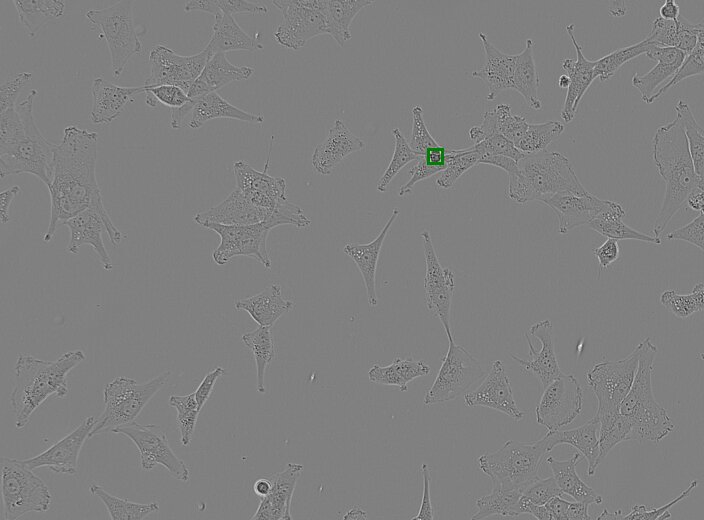}%

\caption{A collection of label errors present in LIVECell. The segmentation masks show the ground truth annotations and the green boxes are the predictions of our label error detection method.}
\label{fig:real_cell}
\end{figure}

\end{document}